\documentclass[draft,phd,11pt]{psuthesis}

\usepackage{amsmath}
\usepackage{amssymb}
\usepackage{amsthm}
\usepackage{exscale}
\usepackage[mathscr]{eucal}
\usepackage{bm}
\usepackage{eqlist} 
\usepackage[final]{graphicx}
\usepackage[dvipsnames]{color}
\DeclareGraphicsExtensions{.pdf, .jpg}

\usepackage{epsf,psfig}
\usepackage{subfigure}
\usepackage{epsfig}
\usepackage{latexsym}
\usepackage{algorithm,algorithmic}
\usepackage{cite,url}
\usepackage{multirow}

\newcommand{\vect}[1]{\pmb{#1}}
\newcommand{\mat}[1]{\pmb{#1}}

\newcommand{\norm}[1]{\left\|#1\right\|}
\newcommand{\R}{\mathbb{R}}

\def\ben{\begin{equation*}}
\def\een{\end{equation*}}
\def\be{\begin{equation}}
\def\ee{\end{equation}}
\def\beaa{\begin{eqnarray*}}
\def\eeaa{\end{eqnarray*}}
\def\bea{\begin{eqnarray}}
\def\eea{\end{eqnarray}}

\usepackage[Lenny]{fncychap}
\ChTitleVar{\Huge\sffamily\bfseries}

\title{Discriminative Models for Robust Image Classification}

\author{Umamahesh Srinivas}
\dept{Electrical Engineering}
\degreedate{August 2013}
\copyrightyear{2013}

\honorsdegreeinfo{for a baccalaureate degree \\ in Engineering Science \\ with honors in Engineering Science}

\documenttype{Dissertation}

\submittedto{The Graduate School}

\numberofreaders{4}

\honorsadviser{Honors P. Adviser}

\secondthesissupervisor{Second T. Supervisor}

\honorsdepthead{Department Q. Head}

\advisor[Dissertation Advisor, Chair of Committee]
        {Vishal Monga}
        {Monkowski Assistant Professor of Electrical Engineering}

\readerone[]
          {William E. Higgins}
          {Distinguished Professor of Electrical Engineering}

\readertwo[]
          {Ram M. Narayanan}
          {Professor of Electrical Engineering}

\readerthree[]
            {Robert T. Collins}
            {Associate Professor of Computer Science and Engineering}

\readerfour[Head of the Department of Electrical Engineering]
           {Kultegin Aydin}
           {Professor of Electrical Engineering}
\begin{document}
\frontmatter

%


\psutitlepage

\psucommitteepage

%
\pagestyle{plain}
\chapter*{Abstract}
    \begin{singlespace}
A variety of real-world tasks involve the classification of images into pre-determined categories. Designing image classification algorithms that exhibit robustness to acquisition noise and image distortions, particularly when the available training data are insufficient to learn accurate models, is a significant challenge. This dissertation explores the development of \emph{discriminative models} for \emph{robust image classification} that exploit underlying signal structure, via probabilistic graphical models and sparse signal representations.

Probabilistic graphical models are widely used in many applications to approximate high-dimensional data in a reduced complexity set-up. Learning graphical structures to approximate probability distributions is an area of active research. Recent work has focused on learning graphs in a discriminative manner with the goal of minimizing classification error. In the first part of the dissertation, we develop a discriminative learning framework that exploits the complementary yet correlated information offered by multiple representations (or projections) of a given signal/image. Specifically, we propose a discriminative tree-based scheme for feature fusion by explicitly learning the conditional correlations among such multiple projections in an iterative manner. Experiments reveal the robustness of the resulting graphical model classifier to training insufficiency.

The next part of this dissertation leverages the discriminative power of sparse signal representations. The value of parsimony in signal representation has been recognized for a long time, most recently in the emergence of compressive sensing. A recent significant contribution to image classification has incorporated the analytical underpinnings of compressive sensing for classification tasks via class-specific dictionaries. In continuation of our theme of exploiting information from multiple signal representations, we propose a discriminative sparsity model for image classification applicable to a general multi-sensor fusion scenario. As a specific instance, we develop a color image classification framework that combines the complementary merits of the the red, green and blue channels of color images. Here signal structure manifests itself in the form of block-sparse coefficient matrices, leading to the formulation and solution of new optimization problems.

As a logical consummation of these ideas, we explore the possibility of learning discriminative graphical models on sparse signal representations. Our efforts are inspired by exciting ongoing work towards uncovering fundamental relationships between graphical models and sparse signals. First, we show the effectiveness of the graph-based feature fusion framework wherein the trees are learned on multiple sparse representations obtained from a collection of training images. A caveat for the success of sparse classification methods is the requirement of abundant training information. On the other hand, many practical situations suffer from the limitation of limited available training. So next, we revisit the sparse representation-based classification problem from a Bayesian perspective. We show that using class-specific priors in conjunction with class-specific dictionaries leads to better discrimination. We employ spike-and-slab graphical priors to simultaneously capture the class-specific structure and sparsity inherent in the signal coefficients. We demonstrate that using graphical priors in a Bayesian set-up alleviates the burden on training set size for sparsity-based classification methods.

An important goal of this dissertation is to demonstrate the wide applicability of these algorithmic tools for practical applications. To that end, we consider important problems in the areas of:
\begin{enumerate}
  \item Remote sensing: automatic target recognition using synthetic aperture radar images, hyperspectral target detection and classification
	\item Biometrics for security: human face recognition
	\item Medical imaging: histopathological images acquired from mammalian tissues.
\end{enumerate}

    \end{singlespace}
\newpage

\thesistableofcontents

\thesislistoffigures

\thesislistoftables


%
\chapter{Acknowledgments}
\begin{singlespace}

It is with the deepest sense of gratitude that I acknowledge the role of my advisor, Professor Vishal Monga, in shaping this dissertation. His zeal for research has been a constant source of inspiration. I am indebted to him for his patient and relentless efforts towards nurturing me into what I am today. His influence has enriched every aspect - adherence to rigorous standards of quality, clarity in technical writing, effective presentation and pedagogy - of my development as a researcher. I will fondly remember his insightful observations about everything during our many conversations, and I look forward to many more years of fruitful association.

I would be remiss not to recall the contributions of the late Professor Nirmal Bose, who advised my MS thesis. I thank him for inculcating in me his philosophy of ``good theory for good practice.''

Words are inadequate to express my gratitude to my parents Srilatha and Srinivas, my grandparents Vishalakshi and Krishnamurthy, my sister Srividya, and extended family for their unconditional love and unfailing support; I owe them everything. This dissertation is my humble dedication to the memory of my late grandfather, whose ideals have profoundly influenced my upbringing and will continue to guide me. A special thank-you to Sandhya for making the last few months enjoyable and the next many years worth looking forward to.

I am thankful to the members of my committee, Professors Bob Collins, Bill Higgins, and Ram Narayanan, for their feedback to improve the quality of this dissertation.

I consider myself fortunate to have been mentored by many wonderful collaborators whose contributions have permeated my research: Professor Trac Tran of The Johns Hopkins University; Dr. Raghu Raj, U.S. Naval Research Laboratory; Dr. Nasser Nasrabadi, U.S. Army Research Laboratory; Dr. Jennifer Gille and Dr. Manu Parmar, Qualcomm MEMS Technologies; and Dr. Arthur Hattel and Dr. Bhushan Jayarao, Animal Diagnostics Laboratory, Penn State. I thank them for their words of wisdom and encouragement.

The Information Processing and Algorithms Laboratory (iPAL) has been my academic home during the last four years. I thank my wonderful lab mates Mu Li, Xuan Mo, Bosung Kang, and Hojjat Mousavi for the many stimulating discussions on a variety of topics. I have also thoroughly enjoyed and benefited from my collaboration with Professor Tran's students at JHU: Yi Chen, Yuanming Suo, and Minh Dao.

Over the past six years, many friends have helped make my stay in State College memorable. Special thanks to my roommate Harisha for 
being there always for me. To Abhijith, Ashwath, Bhaskar, Gautham, Mahesh, Manjula, Niveditha, Shubha, and Vikas, I will remain grateful for the wonderful company and friendship. I will cherish our umpteen discussions - invariably accompanied by great food! - about things casual and serious. To my friends from AID and the Raaga group, thank you all.
\end{singlespace}

\thesisdedication{SupplementaryMaterial/Dedication}{Dedication}

\thesismainmatter

\allowdisplaybreaks{
%
\chapter{Introduction}
\label{chapter:intro}
\section{Motivation}
\label{ch1:motivation}
\begin{figure}[h]
  \centering
	\includegraphics[scale=0.5]{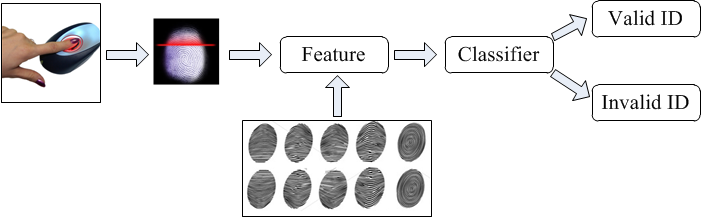}
	\caption[Schematic of fingerprint verification]{Schematic of fingerprint verification.}
	\label{fig:fingerprint}
\end{figure}
Let us consider a scenario where a select group of individuals has access to a secure facility, and their identity is authenticated using their fingerprints. Fig. \ref{fig:fingerprint} describes the steps involved in this process. A person places his/her finger on a scanning device, which compares it with images of fingerprints from a stored labeled database and decides to either allow or deny access. This is just one example of image classification, a commonly encountered task in many domains of image processing today. Typically, in what is referred to as a supervised setting, we have access to a collection of labeled training images belonging to two or more perceptual classes or categories. The challenge lies in utilizing the training data to learn an effective automatic scheme of assigning a new image to its correct category. Image classification is employed in real-world applications as diverse as disease identification in medical imaging, video tracking, optical character recognition, document classification, biometric applications for security (fingerprints or faces), automatic target recognition in remote sensing, and many more.

Image classification can be considered to be a two-stage process in general. First, features that encapsulate image information are extracted from training data. The process of feature extraction can be interpreted as a projection from the image space to a feature space. It is desirable for such features to exhibit two characteristics:
\begin{enumerate}
  \item Low dimensionality: High-dimensional data are commonly encountered in real-world image classification problems. In the interest of computational efficiency, feature extraction techniques condense image information into feature vectors of much smaller dimension by exploiting the redundancy in image pixel intensity information. For example, 2-D wavelet features offer a compact multi-resolution representation of images based on the observation that image content is mostly sparse in the high spatial frequency regions.
	\item Discriminability: The features are designed to capture information about a class of images sufficient to distinguish it from images of other classes. This requires a good understanding of the domain of application and the underlying imaging physics. Different sensing mechanisms and applications lend themselves to the design of different types of features. For example, the arches, loops and whorls that characterize fingerprints necessitate geometry-based image features that can capture the specific curve patterns; eigenfaces derived from a principal component analysis of the training image matrix have proved to be effective for face recognition; morphological image features have seen success in medical imaging problems.
\end{enumerate}
In the second stage of image classification, a decision engine (or equivalently, a classifier) is learned using training features from all classes. The feature vector corresponding to a test image, whose class association is unknown, is evaluated using the classifier rule and the image is assigned to a particular class. A desirable property of such classifiers is generalization, i.e. the ability to exhibit good classification accuracy over a large set of unknown test images.

From a detection-theoretic perspective, classification can be cast as a hypothesis testing problem. For simplicity of exposition, let us consider binary classification, wherein an image has to be assigned to one of two possible classes. Classical binary hypothesis testing is the cornerstone of a variety of signal classification problems. Based on the observation of a single random vector $\vect{x} \in \mathcal{X}^n$, we consider the two simple hypotheses
\bea
H_0 &:& \vect{x} \sim f(\vect{x}|H_0) \nonumber\\
H_1 &:& \vect{x} \sim f(\vect{x}|H_1).
\label{eq:binary_hypothesis}
\eea
Traditionally, $H_0$ and $H_1$ are referred to as the \emph{null} and \emph{alternative} hypotheses respectively. For continuous-valued variables, $\mathcal{X}\equiv \mathbb{R}$, while in the discrete-valued case, $\mathcal{X}$ is a countable collection of indexed values. In the Bayesian testing scenario, the optimal test compares the likelihood ratio $L(\vect{x})$ to a threshold $\tau$,
\be
L(\vect{x}) := \frac{f(\vect{x}|H_1)}{f(\vect{x}|H_0)} ~\overset{H_1}{\underset{H_0}{\gtrless} } \tau.
\label{eq:lrt}
\ee
Classical hypothesis testing mandates that the true conditional densities $f(\vect{x}|\cdot)$ are known exactly. In most real-world classification problems however, this assumption rarely holds. One reason for this is the prevalence of high-dimensional data as discussed earlier. The vectorized version of an image of size $200 \times 200$ lies in $\mathbb{R}^{40000}$. Even for the case of binary images, where each pixel assumes the value 0 or 1, there are over $10^{12000}$ possible different images. In practice, we usually have access only to a much smaller set of labeled training images $\{(\vect{x}_1,y_1),(\vect{x}_2,y_2),\ldots,(\vect{x}_T,y_T)\}$, where each $\vect{x}_i \in \mathcal{X}^n$ and each $y_i$ is a binary label; for example, $y_i \in \{-1,+1\}$. Class decisions are made using (often inaccurate) empirical estimates of the true densities learned from available training. This problem persists even when low-dimensional image features are considered instead of entire images. \emph{Training insufficiency} is thus an important concern in image classification.

Further, the acquired images are often corrupted by various types of noise native to the sensing mechanism. Examples include optical limitations in digital cameras and speckle in ultrasound sensors. Some unexpected sources of noise or image distortions are not always incorporated into the classification framework. The minutiae features in fingerprints can get distorted with aging or injury. As a different example, the exemplar images used for training in face recognition are usually captured under good illumination and are well-aligned. A test image however could include the face of the same individual wearing sunglasses or a hat, or the test image could be captured in poor lighting. These illustrations highlight the need for classification schemes which incorporate a notion of robustness to various types of uncertainty for effective application in practical problems.

In this dissertation, I present new theoretical ideas and experimental validation to support the following thesis:\\
\emph{Discriminative models that exploit signal structure can offer benefits of robustness to noise and training insufficiency in image classification problems.}

\section{Overview of Dissertation Contributions}
This dissertation explores the design of discriminative models for robust image classification. The different models proposed here are unified by the common goal of identifying and leveraging the \emph{discriminative structure} inherent in the images or feature representations thereof. By structure, we refer to the relationships (deterministic or stochastic) between the variables and, more importantly, the existence of discriminative information in such variables that is most relevant for classification tasks. Specifically, we develop two different families of discriminative models for image classification using ideas founded in probabilistic graphical models and the theory of sparse signal representation.

How does discriminative structure manifest itself in real-world image classification tasks? In spite of a proliferation of techniques for feature extraction and classifier design, consensus has evolved that no single feature set or decision engine is uniformly superior for all image classification tasks. In parallel, ongoing research in decision fusion and feature fusion has unearthed the potential of such schemes to infuse robustness into classification by mining the diversity of available information. Motivated by this, a recurring theme in this dissertation is to develop robust discriminative models by exploiting structure in the \emph{complementary yet correlated} information from different feature sets. 

\subsection{Discriminative Graphical Models}
In the first part of this dissertation, we pose the following question: Given a collection of multiple feature sets pertaining to the same observed images, (how) can we learn their class-conditional dependencies to improve classification performance over schemes that use a subset of these features? In response, we propose a framework for feature fusion using \emph{discriminative graphical models}. Graphical models offer a tractable means of learning models for high-dimensional data efficiently by capturing dependencies crucial to classification in a reduced complexity set-up. A graph $\mathcal{G} = (\mathcal{V},\mathcal{E})$ is defined by a set of nodes $\mathcal{V} = \{v_1,\ldots,v_n\}$ and a set of (undirected) edges $\mathcal{E} \subset \binom{\mathcal{V}}{2}$, i.e. the set of unordered pairs of nodes. A probabilistic graphical model is obtained by defining a random vector on $\mathcal{G}$ such that each node represents one (or more) random variables and the edges reveal conditional dependencies. This marriage of graph theory and probability theory offers an intuitive visualization of a probability distribution from which conditional dependence relations can be easily identified. While we focus only on undirected graphs in this dissertation, directed graphs also are an active area of research. The use of graphical models also enables us to draw upon the rich resource of efficient graph-theoretic algorithms to learn complex models and perform inference. Graphical models have thus found application in a variety of tasks, such as speech recognition, computer vision, sensor networks, biological modeling, artificial intelligence, and combinatorial optimization. A more elaborate treatment of the topic is available in \cite{lauritzen:book96,wainwright:book08}.

Our contribution builds on recent work in discriminative graph learning \cite{tan:tsp10}. First, for each type of features, we learn pairs of tree-structured graphs in a discriminative manner. Now we have a collection of unconnected trees for each hypothesis, equivalent to the scenario of na\"{\i}ve Bayes classification which assumes statistical independence of the feature sets. Next, we iteratively learn multiple trees on the larger graphs formed by concatenating all nodes from each hypothesis. Crucially, the new edges learned in each iteration capture conditional dependencies across feature sets. By learning simple tree graphs in each iteration and accumulating all edges in the final graphical structure, we learn a dense edge graphical structure which encodes discriminative information in a tractable manner. It must be noted that this framework makes minimal assumptions on the feature extraction process - specifically, that the feature sets must be correlated. An important consequence of mining the conditional dependencies using discriminative trees is the relative robustness of classification performance as the size of the available training set reduces. As we shall demonstrate through experimental validation, this addresses an important practical concern.

\subsection{Discriminative Sparse Representations}
It is well-known that a large class of signals, including audio and images, can be expressed naturally in a compact manner with respect to well-chosen basis representations. Among the most widely applicable of such basis representations are the Fourier and wavelet basis. This has inspired a proliferation of applications that involve sparse signal representations for acquisition \cite{candes:tit06}, compression \cite{taubman_jpeg:01}, and modeling \cite{lustig_mri:mrm07}. The central problem in compressive sensing (CS) is to recover a signal $\vect x \in \mathbb{R}^n$ given a vector of linear measurements $\vect y \in \mathbb{R}^m$ of the form $\vect y = \mat A\vect x$, where $m \ll n$. Assuming $\vect x$ is compressible, it can be recovered from this underdetermined system of equations by solving the following problem \cite{candes:tit06}:
\be
\quad \min_{\vect x}\|\vect x\|_0 ~\mbox{subject to}~ \vect y = \mat A\vect x,
\label{eq:l0}
\ee
where $\|\vect x\|_0$ is the $l_0$-``norm'' that counts the number of non-zero entries in $\vect x$. Under certain conditions, the $l_0$-norm can be relaxed to the $l_1$-norm, leading to convex optimization formulations.

Although the concept of sparsity was introduced to solve inverse reconstructive problems, where it acts as a strong prior to the abbreviated ill-posed nature of the problems, recent work \cite{wright:tpami09,wagner:tpami12} has demonstrated the effectiveness of sparse representation in classification applications too. The crucial observation is that a test image can be reasonably approximated as a linear combination of training images belonging to the same class, with (ideally) no contributions from training images of other classes. Therefore, with $\mat A := [\mat A_1 ~ \ldots ~ \mat A_K]$ and $\mat A_i$ representing the matrix of vectorized training images from the $i$-th class, the corresponding coefficient vector $\vect x := [\vect x_1^T ~ \ldots ~ \vect x_K^T]^T$ is sparse and naturally encodes discriminative information. In other words, the semantic information of the signal of interest is often captured in the sparse representation. Albeit simplistic in formulation, this linear sparse representation model is rooted in well-conditioned optimization theory. The sparse representations exhibit robustness to a variety of real-world image distortions, leading to their widespread use in applications such as face recognition, remote sensing and medical image classification for disease diagnosis.

In pursuit of our goal of robust classification, a natural progression of thought is to inquire if this sparse representation-based classification (SRC) framework can be extended to the scenario of multiple correlated observations of a given test image. Efforts have already been directed towards collaborative (or alternately discriminative/group/simultaneous) sparse models for sensing and recovery as well as classification. In practice, the measurements could come from homogeneous sensors - for example, multiple camera views or remote sensing arrays used in hyperspectral or radar imaging. An emerging area of research interest is multi-sensor fusion, where the data/features are accumulated using heterogeneous sensing modalities. To illustrate, multi-modal biometrics for security applications could combine fingerprint verification as well as face recognition. While the observations are still correlated since they pertain to the same image, new notions of signal structure and correlation emerge.

Accordingly, the second part of this dissertation concerns itself with innovative ways of mining structural dependencies among discriminative sparse signal representations for classification. First, we develop a multi-task multi-variate model for sparse representation-based classification that is applicable for a variety of multi-modal fusion applications. As a specific instantiation of the framework, we consider the problem of categorizing histopathological (medical tissue) images as either healthy or diseased (inflammatory). The tissue staining process unique to histopathology leads to digitized images that encode class information in the red and blue color channels. So we develop a simultaneous sparsity model for color images which exploits the color channel correlations. The sparse coefficient matrices have a block-diagonal structure due to meaningful constraints imposed by imaging physics. We propose a variation of a well-known greedy algorithm to recover this new sparse structure.

A caveat for the success of sparse classification methods is the requirement of abundant training information; the linear combination model will not hold well otherwise. On the other hand, many practical situations suffer from the limitation of limited training. So, we revisit the SRC framework from a Bayesian standpoint. It is well known that \emph{a priori} knowledge about the structure of signals often leads to significant performance improvements in many signal analysis and processing applications. Such information typically manifests itself in the form of priors, constraints or regularizers in analytical formulations of the problems. In fact, the $l_1$-norm variant of \eqref{eq:l0} is equivalent to enforcing a sparsity-inducing Laplacian prior on the coefficients $\vect x$. Taking this idea further, we investigate new optimization problems resulting from the enforcement of other sparsity-inducing distributions such as the spike-and-slab prior. Significantly, we look for discriminative \emph{graphical} priors that can simultaneously encode signal \emph{structure} and \emph{sparsity}. Using graphical priors in a Bayesian set-up alleviates the burden on training set size for sparsity-based classification methods. Our efforts in this direction are inspired by ongoing work towards uncovering fundamental relationships between graphical models and sparse representations \cite{cevher:spm10}.

\section{Organization}
A snapshot of the main contributions of this dissertation is presented next. Publications related to the contribution in each chapter are also listed where applicable.

In \textbf{Chapter \ref{chapter:gm}}, the primary contribution is the proposal of probabilistic graphical models as a tool for low-level feature fusion and classification. The algorithm can be applied broadly to any fusion-based classification problem and is described in all generality. We also show a specific application to the problem of automatic target recognition (ATR) using synthetic aperture radar (SAR) images. Model-based algorithmic approaches to target classification \cite{Sullivan2001,DeVore2002} attempt to learn the underlying class-specific statistics for each class of target vehicles. This is clearly not an easy problem since real-world distortions lead to significant deviations from the assumed ideal model. Also the effects of limited training are particularly pronounced when working with high-dimensional data typical of ATR problems. We leverage a recent advance in discriminative graphical model learning \cite{tan:tsp10} to overcome these issues. The novelty of our contribution is in building a \emph{feature fusion} framework for ATR, which explicitly learns class-conditional statistical dependencies between distinct feature sets. To the best of our knowledge, this is the first such application of probabilistic graphical models in ATR.

As for target image representations or feature sets, we employ well-known wavelet LL, LH, and HL sub-bands (H = high, L = low). Usually the LL sub-band wavelet coefficients are used as features, since they capture most of the low-frequency content in the images. Although largely ignored in ATR problems, the LH and HL coefficients carry high frequency discriminative information by capturing scene edges in different orientations. We utilize this discriminative aspect of the LH and HL sub-bands together with the approximate (coarse) information from the LL sub-band as the sources of complementary yet correlated information about the target scene.

A significant experimental contribution of this work is the comparison of classification performance as a function of the size of the training set. Traditionally classification performance is reported in terms of confusion matrices and receiver operating characteristic (ROC) curves. We believe that a more relevant and necessary comparison is with the number of training samples available per class, which is often a serious practical concern in ATR systems. In comparison with state-of-the-art alternatives, we show that our graphical model framework exhibits a more graceful decay in performance under reduced training, thereby highlighting its superior robustness.
	
This material was presented at the 2011 IEEE International Conference on Image Processing \cite{srinivas:icip11} and will appear shortly in the IEEE Transactions on Aerospace and Electronic Systems \cite{srinivas:taes13}.

\textbf{Chapter \ref{chapter:sparsity_gm}} serves as an introduction to sparse signal representations and their role in discriminative tasks. In joint sparsity models for classification, one assumption is that all test vectors obey the exact same linear representation model, leading to coefficient vectors with identical sparse structure but possibly different weights. Motivation for such an assumption is often derived from the underlying imaging physics. Since sparse representations are known to be inherently discriminative, these sparsity model-based approaches perform class assignment using the class-specific \emph{reconstruction} error, although the original task is one of classification. We investigate the performance benefits of using truly discriminative classifiers on sparse features in this chapter. Specifically, in continuation of the graph-based framework from Chapter 2, we present two instantiations of learning such graphical models directly on \emph{sparse} image features. These experimental demonstrations offer validation of the effectiveness of training graphs on sparse features. This work was done in collaboration with Prof. Trac Tran at the Johns Hopkins University, Baltimore, MD.

The first application is to hyperspectral target image detection and classification. Spatio-spectral information is fused for robust classification by exploiting the correlations among sparse representations learned from local pixel neighborhoods. This material was presented at the 2012 IEEE International Symposium on Geoscience and Remote Sensing \cite{srinivas:igarss12} and was published in the IEEE Geoscience and Remote Sensing Letters \cite{srinivas:grsl13} in May 2013.
	
The second application is the well-known problem of face recognition. Recognizing the potential of local image features to encode discriminative information more robustly than global image features, we extract sparse features separately from the eyes, nose, and mouth of human subjects in training images and combine this information via discriminative trees. This material was presented at the 2011 IEEE Asilomar Conference on Signals, Systems and Computers \cite{srinivas_facerec:11}.
	
On a related note, we employed the graph-based framework for transient acoustic signal classification in collaboration with the U.S. Army Research Laboratory, Adelphi, MD. This work is however not included in this dissertation. The material was presented at the 2013 IEEE International Conference on Acoustics, Speech, and Signal Processing \cite{srinivas:icassp13a} and submitted to the IEEE Transactions on Cybernetics \cite{srinivas:tc13} in April 2013.

\textbf{Chapter \ref{chapter:struct_sparsity}} discusses innovative ways of learning discriminative models on sparse signal representations. First, we develop a simultaneous sparsity model for color medical image classification. The motivation for this work originated from discussions with veterinary pathologists at the Animal Diagnostic Laboratory (ADL), Pennsylvania State University. Digitized tissue images from mammalian tissue - renal, splenic and pulmonary - were provided along with the results of classification by human observers as a baseline. The multi-channel nature of digital histopathological images presents an opportunity to exploit the correlated color channel information for better image modeling. We develop a new simultaneous Sparsity model for multi-channel Histopathological Image Representation and Classification (SHIRC). Essentially, we represent a histopathological image as a sparse linear combination of training examples under suitable channel-wise constraints. Classification is performed by solving a newly formulated simultaneous sparsity-based optimization problem. A practical challenge is the correspondence of image objects (cellular and nuclear structures) at different spatial locations in the image. We propose a robust locally adaptive variant of SHIRC (LA-SHIRC) to tackle this issue. This material was presented at the 2013 IEEE International Symposium on Biomedical Imaging \cite{srinivas:isbi13} and has been submitted to the IEEE Transactions on Image Processing \cite{srinivas:tip13a} in April 2013.

Next, we present the results of our recent and ongoing explorations into the design of structured graphical priors for sparsity-based classification. Model-based CS exploits the structure inherent in sparse signals for the design of better signal recovery algorithms. Analogously, our contribution is a logical extension of the sparse representation-based classification idea to discriminative models for structured sparsity. We incorporate class-specific dictionaries in conjunction with discriminative \emph{class-specific priors}. Specifically, we use the spike-and-slab prior that has widely been acknowledged to be the gold standard for sparse signal modeling. Theoretical analysis reveals similarities to well-known regularization-based formulations such as the lasso and the elastic net. Crucially, our Bayesian approach alleviates the requirement of abundant training necessary for the success of sparsity-based schemes. We also show that the framework organically evolves in complexity for multi-task classification while maintaining parallels with collaborative hierarchical schemes. The material was presented at the 2013 IEEE International Conference on Acoustic, Speech, and Signal Processing \cite{srinivas:icassp13b}, will be presented at the 2013 IEEE International Conference on Image Processing \cite{srinivas:icip13}, and has been submitted to the IEEE Transactions on Image Processing \cite{srinivas:tip13b} in July 2013.

In \textbf{Chapter \ref{chapter:conclusion}}, the main contributions of this dissertation are summarized. We also outline interesting directions for future research that can combine the robustness benefits of graphical models and sparse representations in a synergistic manner for image classification applications.

\chapter{Feature Fusion for Classification via Discriminative Graphical Models}
\label{chapter:gm}


\section{Introduction}
\label{ch2:intro}

The primary contribution of this chapter is the proposal of probabilistic graphical models as a tool for low-level feature fusion in image classification problems. We are interested in scenarios where we have access to multiple sets of features corresponding to the same observed images. We argue that {\em explicitly} capturing statistical dependencies between \emph{distinct} \emph{low-level} image feature sets can improve classification performance compared to existing fusion techniques. Accordingly, we develop a two-stage framework to \emph{directly} model dependencies between different feature representations of a given collection of images. The first stage involves the generation of multiple image representations (or feature sets) that carry complementary benefits for image classification. In the second stage, we perform inference (class assignment) that can exploit {\em class-conditional correlations} among the aforementioned feature sets. This is a reasonably hard task as images (or representative features thereof) are typically high-dimensional and the number of training images corresponding to a target class is limited. We address this problem by using discriminative graphical models. Our graphical model-based classifiers are designed by first learning discriminative tree graphs on each distinct feature set. These initial disjoint trees are then {\em thickened}, i.e.\ augmented with more edges to capture feature correlations, via boosting on disjoint graphs.

The proposed framework is quite general in its applicability to a variety of real-world fusion problems. In this chapter, we demonstrate the experimental benefits of our proposed approach for the problem of automatic target recognition using synthetic aperture radar images.

\section{Background}

\subsection{Probabilistic Graphical Models}
Probabilistic graphical models present a tractable means of learning models for high-dimensional data efficiently by capturing dependencies crucial to the application task in a reduced complexity set-up. A graph $\mathcal{G} = (\mathcal{V},\mathcal{E})$ is defined by a set of nodes $\mathcal{V} = \{v_1,\ldots,v_n\}$ and a set of (undirected) edges $\mathcal{E} \subset \binom{\mathcal{V}}{2}$, i.e. the set of unordered pairs of nodes. Graphs vary in structural complexity from sparse tree graphs to fully-connected dense graphs. A probabilistic graphical model is obtained by defining a random vector on $\mathcal{G}$ such that each node represents one (or more) random variables and the edges reveal conditional dependencies. The graph structure thus defines a particular way of factorizing the joint probability distribution. Graphical models offer an intuitive visualization of a probability distribution from which conditional dependence relations can be easily identified. The use of graphical models also enables us to draw upon the rich resource of efficient graph-theoretic algorithms to learn complex models and perform inference. Graphical models have thus found application in a variety of modeling and inference tasks such as speech recognition, computer vision, sensor networks, biological modeling, and artificial intelligence. The interested reader is referred to \cite{lauritzen:book96,wainwright:book08} for a more elaborate treatment of the topic.

In the ensuing discussion, we assume a binary classification problem for simplicity of exposition. The proposed algorithm naturally extends to the multi-class classification scenario by designing discriminative graphs in a one-versus-all manner (discussed in Section \ref{sec:multiClass}). The history of graph-based learning can be traced to Chow and Liu's \cite{chow:tit68} seminal idea of learning the optimal \emph{tree approximation} $\hat{p}$ of a multivariate distribution $p$ using first- and second-order statistics:
\be
\hat{p} = \arg \min_{p_t \in \mathcal{T}_{{p}}} D(p||p_t),
\label{eq:chow-liu}
\ee
where $D(p||p_t) = \mathrm{E}_{p}[\log (p/p_t)]$ denotes the Kullback-Leibler (KL) divergence and $\mathcal{T}_{{p}}$ is the set of all tree-structured approximations to $p$. This optimization problem is shown to be equivalent to a maximum-weight spanning tree (MWST) problem, which can be solved using Kruskal's algorithm \cite{kruskal56} for example. The mutual information between a pair of random variables is chosen to be the weight assigned to the edge connecting those random variables in the graph. This idea exemplifies \emph{generative learning}, wherein a graph is learned to approximate a given distribution by minimizing a measure of \emph{approximation error}. This approach has been extended to a classification framework by independently learning two graphs to respectively approximate the two class empirical estimates, and then performing inference. A decidedly better approach is to jointly learn a pair of graphs from the pair of empirical estimates by minimizing \emph{classification error} in a \emph{discriminative learning} paradigm. An example of such an approach is \cite{Friedman97}.

Recently, Tan {\em et al.} \cite{tan:tsp10} proposed a graph-based discriminative learning framework, based on maximizing an approximation to the $J$-divergence. Given two probability distributions $p$ and $q$, the $J$-divergence is defined as: $J(p,q) = D(p||q) + D(q||p)$. This is a symmetric extension of the KL-divergence. The tree-approximate $J$-divergence is then defined \cite{tan:tsp10} as 
\be
\hat{J}(\hat{p},\hat{q};p,q) = \int (p(x)-q(x))\log\left[\frac{\hat{p}(x)}{\hat{q}(x)}\right] dx,
\label{eq:J-approx}
\ee
and it measures the ``separation'' between tree-structured approximations $\hat{p}$ ($\in \mathcal{T}_{{p}}$) and $\hat{q}$ ($\in \mathcal{T}_{{q}}$). Using the key observation that maximizing the $J$-divergence minimizes the upper bound on probability of classification error, the discriminative tree learning problem is stated (in terms of empirical estimates $\tilde{p}$ and $\tilde{q}$) as
\be
(\hat{p},\hat{q}) = \arg \max_{\hat{p} \in \mathcal{T}_{\widetilde{p}}, \hat{q} \in \mathcal{T}_{\widetilde{q}}} \hat{J}(\hat{p},\hat{q};\tilde{p},\tilde{q}).
\label{eq:disc-learn}
\ee
It is shown in \cite{tan:tsp10} that this optimization further decouples into two MWST problems (see Fig. \ref{fig:schematicDiscTree}):
\bea
\hat{p} & = & \arg \min_{\hat{p} \in \mathcal{T}_{\widetilde{p}}} D(\tilde{p}||\hat{p}) - D(\tilde{q}||\hat{p}) \label{eq:p-opt}\\
\hat{q} & = & \arg \min_{\hat{q} \in \mathcal{T}_{\widetilde{q}}} D(\tilde{q}||\hat{q}) - D(\tilde{p}||\hat{q}) \label{eq:q-opt}.
\eea

\begin{figure}[t]
  \centering
  \includegraphics[scale=0.45]{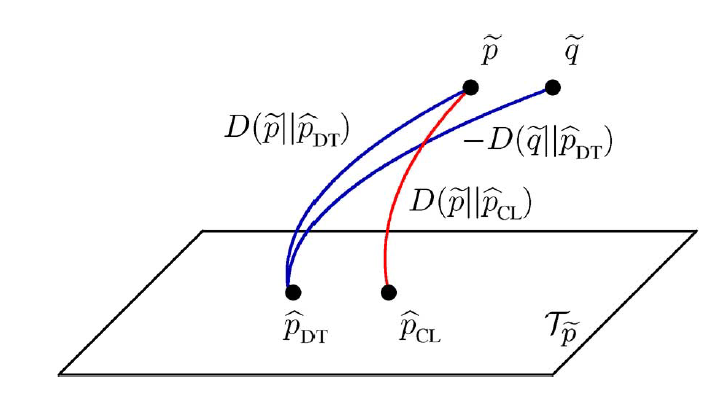}
  \caption[Illustration of \eqref{eq:p-opt}-\eqref{eq:q-opt}]{Illustration of \eqref{eq:p-opt}-\eqref{eq:q-opt} (figure reproduced from \cite{tan:tsp10}). $\mathcal{T}_{\widetilde{p}}$ is the subset of tree distributions marginally consistent with $\widetilde{p}$. The generatively-learned distribution (via Chow-Liu) $\widehat{p}_{\mbox{\tiny{CL}}}$, is the projection of $\widetilde{p}$ onto $\mathcal{T}_{\widehat{p}}$, according to \eqref{eq:chow-liu}. The discriminatively-learned distribution $\hat{p}_{\mbox{\tiny{DT}}}$, is the solution of \eqref{eq:p-opt} which is ``further'' from $\widetilde{q}$ in the KL-divergence sense.}
  \label{fig:schematicDiscTree}
\end{figure}

\noindent \textbf{Learning thicker graphical models:} The optimal tree graphs in the sense of minimizing classification error are obtained as solutions to \eqref{eq:p-opt}-\eqref{eq:q-opt}. Such tree graphs have limited ability to model general distributions; learning arbitrarily complex graphs with more involved edge structure is known to be NP-hard \cite{cooper:1990}, \cite{Karger01}. In practice, one way of addressing this inherent tradeoff between complexity and performance is to boost simpler graphs \cite{tan:tsp10,boosting:Drucker96,Downs04} into richer structures. In this chapter, we employ the boosting-based algorithm from \cite{tan:tsp10}, wherein different pairs of discriminative graphs are learned - by solving \ \eqref{eq:p-opt}-\eqref{eq:q-opt} repeatedly - over the same sets of nodes (but weighted differently) in different iterations via boosting. This process results in a ``thicker'' graph formed by augmenting the original tree with the newly-learned edges from each iteration.

\subsection{Boosting}
Boosting \cite{freund:jsai99}, which traces its roots to the probably approximately correct (PAC) learning model, iteratively improves the performance of weak learners which are marginally better than random guessing into a classification algorithm having arbitrarily accurate performance. A distribution of weights is maintained over the training set. In each iteration, a weak learner $h_t$ minimizes the weighted training error to determine class confidence values. Weights on incorrectly classified samples are increased so that the weak learners are penalized for the harder examples. The final boosted classifier makes a class decision based on a weighted average of the individual confidence values. In the Real-AdaBoost version, each $h_t$ determines class confidence values instead of discrete class decisions; it is described in Algorithm \ref{alg:adaboost}. $S$ is the set of training features and $N$ is the number of available training samples.
\begin{algorithm}[t]
\caption{AdaBoost learning algorithm}
\label{alg:adaboost}
\begin{algorithmic}[1]
\STATE Input data $(x_i,y_i),~i = 1,2,\ldots,N$, where $x_i \in S,~y_i \in \{-1,+1\}$
\STATE Initialize $D_1(i) = \frac{1}{N}, i = 1,2,\ldots,N$
\STATE For $t = 1,2,\ldots,T$:
        \begin{itemize}
        \item Train weak learner using distribution $D_t$
        \item Determine weak hypothesis $h_t: S \mapsto \mathbb{R}$ with error $\epsilon_t$
        \item Choose $\beta_t = \frac{1}{2}\ln\left(\frac{1-\epsilon_t}{\epsilon_t}\right)$
        \item $D_{t+1}(i) = \frac{1}{Z_t}\left\{D_t(i)\exp(-\beta_t y_i h_t(x_i))\right\}$, where $Z_t$ is a normalization factor
        \end{itemize}
\STATE Output decision $H(x) = \mbox{sign}\left[\sum_{t=1}^{T}\beta_th_t(x)\right]$.
\end{algorithmic}
\end{algorithm}

Boosting has been used in a variety of practical applications as a means of enhancing the performance of weak classifiers. In \cite{boosting:Sun07} for example, radial basis function (RBF) nets are used as the weak learners for boosting. Boosting has also been deployed on classifiers derived from decision trees\cite{boosting:Drucker96}. In the sequel, we use boosting as a tool to combine initially \emph{disjoint} discriminative tree graphs, learned from \emph{distinct} feature representations of a given image, into a thicker graphical model which captures correlation among the different target image representations.

\section[Discriminative Graphical Models for Robust Image Classification]{Discriminative Graphical Models for Robust Image \\Classification}
\label{ch2:discgraphs}

\begin{figure}[t]
  \begin{center}
  \includegraphics[scale=0.85]{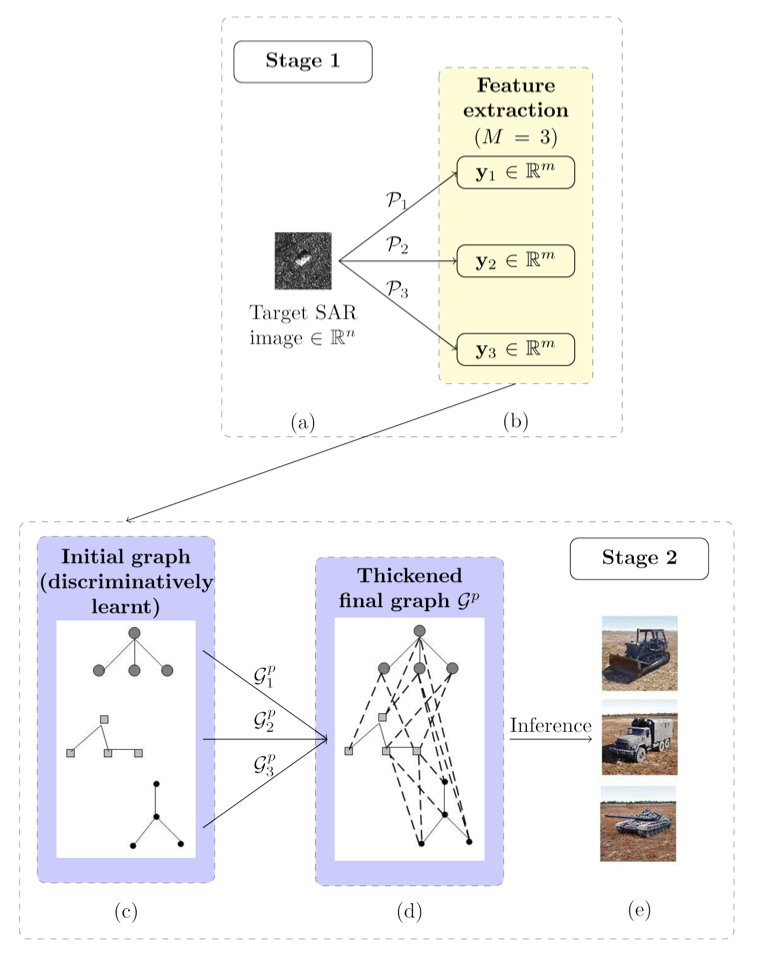}
  \caption[Two-stage framework for designing {\em discriminative} graphical models]{Proposed two-stage framework for designing {\em discriminative} graphical models $\mathcal{G}^p$ and $\mathcal{G}^q$ illustrated for the case when $M = 3$. (a) Sample target image, (b) Feature extraction via a projection $\mathcal{P}$, (c) Initial sparse graph, (d) Final thickened graph: newly learned edges represented by dashed lines, (e) Graph-based inference. In (c)-(e), the corresponding graphs for distribution $q$ are not shown but are learned analogously.}
  \label{fig:igt_main}
  \end{center}
\end{figure}

The schematic of our framework is shown in Fig.\ \ref{fig:igt_main}. The idea is first illustrated for binary classification. Extension to the more general multi-class scenario is discussed in \ref{sec:multiClass}. The graphical model-based algorithm is summarized in Algorithm \ref{alg:igt}, and it consists of an {\em offline} stage to learn the discriminative graphs (Steps 1-4) followed by an {\em online} stage (Steps 5-6) where a new test image is classified. The offline stage involves extraction of features from training images from which approximate probability distribution functions (pdfs) for each class are learned after the graph thickening procedure.

\subsection{Feature Extraction}
\label{sec:featureExt}
The (vectorized) images are assumed to belong to $\mathbb{R}^n$. The extraction of different sets of features from training images is performed in Stage 1. Each such image representation may be viewed as a dimensionality reduction via projection $\mathcal{P}_i: \mathbb{R}^n \mapsto \mathbb{R}^{m_i}$ to some lower dimensional space $\mathbb{R}^{m_i}, m_i < n$. In our framework, we consider $M$ distinct projections $\mathcal{P}_i, i = 1,2,\ldots,M$. For every input $n$-dimensional image, $M$ different features $\vect{y}_i \in \mathbb{R}^{m_i}, i = 1,2,\ldots, M$, are obtained. Fig.\ \ref{fig:igt_main}(b) depicts this process for the particular case $M = 3$. For notational simplicity, in the ensuing description we assume $m_1 = m_2 = \ldots = m_M = m$. The framework only requires that the different projections lead to low-level features that have complementary yet correlated information. For the application to target recognition described in the latter half of this chapter, wavelet sub-bands obtained after a multi-level wavelet decomposition are used as features.

\subsection{Initial Disjoint Tree Graphs}
\label{sec:proposedFramework}

\begin{algorithm}[t]
\caption{Discriminative graphical models (Steps 1-4 offline)}
\label{alg:igt}
\begin{algorithmic}[1]
\STATE \textbf{Feature extraction (training):} Obtain vectors $\vect{y}_i,i=1,\ldots,M$ in ${\mathbb R}^{m}$ using projections ${\mathcal P}_{i}, i=1,\ldots,M$ on target image in ${\mathbb R}^{n}$
\STATE \textbf{Initial disjoint graphs:}\\
       For $i = 1,\ldots,M$:\\
       Discriminatively learn $m$-node tree graphs $\mathcal{G}_i^p$ and $\mathcal{G}_i^q$ from the vectors $\{\vect{y}_i\}$
\STATE Concatenate nodes of the graphs $\mathcal{G}_i^p, i = 1,\ldots,M$ to generate initial graph $\mathcal{G}^{p,0}$; likewise for $\mathcal{G}^{q,0}$
\STATE \textbf{Boosting on disjoint graphs}: Iteratively thicken $\mathcal{G}^{p,0}$ and $\mathcal{G}^{q,0}$ via boosting to obtain final graphs $\mathcal{G}^p$ and $\mathcal{G}^q$
\hrule
\COMMENT{\textbf{Online process}}
\STATE \textbf{Feature extraction (test):} Obtain vectors $\vect{y}_i,i=1,\ldots,M$ in ${\mathbb R}^{m}$ from test target image
\STATE \textbf{Inference}: Classify based on output of the resulting strong classifier using \eqref{eq:decision_rule}.
\end{algorithmic}
\end{algorithm}

Figs.\ \ref{fig:igt_main}(c)-(e) represent Stage 2 of the framework. Consistent with notation in \cite{tan:tsp10}, we denote the two different class distributions by $p$ and $q$ respectively. Let $\{\vect{y}_{i}^{p}\}_{tr}$ represent the set of training feature vectors corresponding to projection $\mathcal{P}_i$ and class $p$; similarly define $\{\vect{y}_{i}^{q}\}_{tr}$. Further, let the class distributions corresponding to $p$ and $q$ for the $i$-th set of features be denoted by $f^{p}(\vect{y}_i)$ and $f^{q}(\vect{y}_i)$ respectively.

A pair of $m$-node discriminative tree graphs $\mathcal{G}_i^p$ and $\mathcal{G}_i^q$ is learned for each feature projection $\mathcal{P}_i, i = 1,2,\ldots,M$, by solving \eqref{eq:p-opt}-\eqref{eq:q-opt}. Fig. \ref{fig:igt_main}(c) shows a toy example of three 4-node tree graphs $\mathcal{G}_1^p, \mathcal{G}_2^p$ and $\mathcal{G}_3^p$. (The corresponding graphs $\mathcal{G}_i^q$ are not shown in the figure.) By concatenating the nodes of the graphs $\mathcal{G}_i^p, i = 1,\ldots,M,$ we have one initial sparse graph structure $\mathcal{G}^{p,0}$ with $Mm$ nodes (Fig. \ref{fig:igt_main}(c)). Similarly, we obtain the initial graph $\mathcal{G}^{q,0}$ by concatenating the nodes of the graphs $\mathcal{G}_i^q, i = 1,\ldots,M$. The joint probability distribution corresponding to $\mathcal{G}^{p,0}$ is the product of the individual probability distributions corresponding to $\mathcal{G}_i^p,i=1,\ldots,M$; likewise for $\mathcal{G}^{q,0}$. Inference based on the graphs $\mathcal{G}^{p,0}$ and $\mathcal{G}^{q,0}$ can thus be interpreted as feature fusion under the na\"{\i}ve Bayes assumption, i.e. statistical independence of the individual target image representations. We have now learned (graphical) pdfs $\hat{f}^p(\vect{y}_i)$ and $\hat{f}^{q}(\vect{y}_i)$ ($i = 1,\ldots,M$).

Our final goal is to learn probabilistic graphs for classification purposes which do not just model the individual feature projections but also capture their mutual class-conditional correlations. This is tantamount to discovering new edges that connect features from the aforementioned three individual tree graphs on the distinct feature sets. Thicker graphs or newer edges can be learned with the {\em same} training sets by boosting several simple graphs \cite{tan:tsp10,freund:jsai99,Downs04,boosting:Drucker96}. In particular, we employ boosting of multiple discriminative tree graphs as described in \cite{tan:tsp10}. Crucially, as the initial graph for boosting, we choose the {\em forest of disjoint graphs} over the individual feature sets. Such initialization has two benefits:
\begin{enumerate}
  \item When the number of training samples is limited, each individual graph in the forest is still well-learned over lower-dimensional feature sets and the forest graph offers a good initialization for the final discriminative graph to be learned iteratively.
  \item This naturally offers a relatively general way of low-level feature fusion, where the initial disjoint graphs could be determined from domain knowledge or learned using any effective graph learning technique (not necessarily tree-structured).
\end{enumerate}

\subsection{Feature Fusion via Boosting on Disjoint Graphs}

We next iteratively thicken the graphs $\mathcal{G}^{p,0}$ and $\mathcal{G}^{q,0}$ using the boosting-based method in \cite{tan:tsp10}. In each iteration, the classifier $h_t$ is a likelihood test using the tree-based approximation of the conditional pdfs. We begin with the set of training features $\{\vect{y}_{i}^{p}\}_{tr}$ and $\{\vect{y}_{i}^{q}\}_{tr}, i = 1,\ldots,M$, and obtain the empirical error $\epsilon_t$ for $t = 1$ as the fraction of training samples that are misclassified. We then compute the parameter $\beta_1$, which varies directly as $\epsilon_1$. Finally, we identify the misclassified samples and re-sample the training set in a way that the re-sampled (or re-weighted) training set now contains a higher number of such samples. This is captured by the distribution $D_{t+1}$ in Algorithm \ref{alg:adaboost}. After each such iteration (indexed by $t$ in Algorithm \ref{alg:adaboost}), a new pair of discriminatively-learned tree graphs $\mathcal{G}^{p,t}$ and $\mathcal{G}^{q,t}$ is obtained by modifying the weights on the existing training sets. The final graph $\mathcal{G}^p$ after $T$ iterations is obtained by augmenting the initial graph $\mathcal{G}^{p,0}$ with all the newly learned edges in the $T$ iterations (Fig. \ref{fig:igt_main}(d)). Graph $\mathcal{G}^q$ is obtained in a similar manner. A simple stopping criterion is devised to decide the number of iterations based on the value of the approximate $J$-divergence at successive iterations.

This process of learning new edges is tantamount to discovering new conditional correlations between distinct sets of image features, as illustrated by the dashed edges in Fig. \ref{fig:igt_main}(d). The thickened graphs $\hat{f}^p(\vect{y})$ and $\hat{f}^q(\vect{y})$ are therefore estimates of the true (but unknown) class-conditional pdfs over the concatenated feature vector $\vect{y}$. If $\mathcal{E}^{p,t}$ represents the set of edges learned for distribution $p$ in iteration $t$ and $\mathcal{E}^p$ represents the edge set of $\mathcal{G}^p$, then
\be
\mathcal{E}^p = \bigcup_{t=0}^{T} \mathcal{E}^{p,t}; \qquad \mathcal{E}^q = \bigcup_{t=0}^{T} \mathcal{E}^{q,t}.
\label{eq:edge-set}
\ee

\noindent \textbf{Inference:} The graph learning procedure described so far is performed offline. The actual classification of a new test image is performed in an online process, via the likelihood ratio test using the estimates of the class-conditional pdfs obtained above, and for a suitably chosen threshold $\tau$,
\be
\log\left(\frac{\hat{f}^{p}(\vect{y})}{\hat{f}^{q}(\vect{y})}\right) \underset{q}{\overset{p}{\gtrless}} \tau.
\label{eq:decision_rule}
\ee

\noindent \textbf{A note on inference complexity:} The initial graph has $M(m-1)$ edges, and each iteration of boosting adds at most $mM-1$ new edges. Making inferences from the learned graph after $T$ boosting iterations requires the multiplication of about $2mMT$ conditional probability densities corresponding to the edges in the two graphs. This is comparable to the cost of making inferences using classifiers such as support vector machines (SVMs). In comparison, inference performed from likelihood ratios by assuming a Gaussian $mM\times mM$ covariance matrix requires $O(m^3M^3)$ computations due to matrix inversion.
%

\subsection{Multi-class Image Classification}
\label{sec:multiClass}
The proposed framework can be extended to a multi-class scenario in a one-versus-all manner as follows. Let ${C}_k, k = 1,2,\ldots,K$, denote the $k$-th class of images, and let $\widetilde{C}_k$ denote the class of images complementary to class ${C}_k$, i.e., $\widetilde{C}_k = \bigcup_{l=1,\ldots,K, l \neq k} {C}_l$. The $k$-th binary classification problem is then concerned with classifying a query image (or corresponding feature) into ${C}_k$ or $\widetilde{C}_k$ ($k = 1,\ldots,K$).

For each such binary problem, we learn graphical estimates of the p.d.fs $\hat{f}_{k}^{p}(\vect{y})$ and $\hat{f}_{k}^{q}(\vect{y})$ as described previously. This process is done in parallel and offline. The image feature vector corresponding to a new test image is assigned to the class $k^{*}$ according to the following decision rule:
\be
k^{*} = \arg\max_{k \in \{1,\ldots,K\}}\log\left(\frac{\hat{f}_{k}^{p}(\vect{y})}{\hat{f}_{k}^{q}(\vect{y})}\right).
\label{eq:decision_rule_multi}
\ee

\noindent \textbf{Inclusion of new image class:} Of practical interest is the flexibility of the algorithm to incorporate training data from a new target class. The inclusion of the $(K+1)$-th class will require just one more pair of graphs to be learned - corresponding to the problem of classifying a query into either $C_{K+1}$ or $\widetilde{C}_{K+1}$. Crucially, these graphs are learned in the \emph{offline} training phase itself, thereby incurring minimal additional computation during the test phase. It must be noted that the features corresponding to the $(K+1)$-th class are not used for training the complementary classes $\widetilde{C}_k$ in the $C_{k}$-versus-$\widetilde{C}_{k}$ problems for $k = 1,2,\ldots,K$. Incorporating these features into the training process of all the $(K+1)$ binary problems can lead to better discrimination but it will require a re-learning (offline) of all $K$ pairs of graphs.

\section{Application: Automatic Target Recognition}

\subsection{Introduction}

The classification of real-world empirical targets using sensed imagery into different perceptual classes is one of the most challenging algorithmic components of radar systems. This problem, popularly known as automatic target recognition (ATR), exploits imagery from diverse sensing sources such as synthetic aperture radar (SAR), inverse SAR (ISAR), and forward-looking infra-red (FLIR) for automatic identification of targets. A review of ATR can be found in \cite{Bhanu93}.

SAR imaging offers the advantages of day-night operation, reduced sensitivity to weather conditions, penetration capability through obstacles, etc. Some of the earlier approaches to SAR ATR can be found in \cite{Zhao01,ross:MSTARprep,bhatnagar:SVDATR,Casasent05,Tison07}. A discussion of SAR ATR theory and algorithms is provided in \cite{Mossing98}. The Moving and Stationary Target Acquisition and Recognition (MSTAR) data set \cite{mstar} is widely used as a benchmark for SAR ATR experimental validation and comparison. Robustness to real-world distortions is a highly desirable characteristic of ATR systems, since targets are often classified in the presence of clutter, occlusion and shadow effects, different capture orientations, confuser vehicles, and in some cases, different serial number variants of the same target vehicle. Typically the performance of ATR algorithms is tested under a variety of operating conditions as discussed in \cite{Mossing98}.

\subsection{ATR Algorithms: Prior Art}

Over two decades' worth of investigations have provided a rich family of algorithmic tools for target classification. Early research in ATR mainly concerned itself with the task of uncovering novel feature sets. Choices for features have been inspired by image processing techniques in other application domains. Popular spatial features are computer vision-based geometric descriptors such as robust edges and corners  \cite{Olson97}, and template classes \cite{bhatnagar:SVDATR,ross:MSTARprep,Mahalanobis98}. Selective retention of transform domain coefficients based on wavelets \cite{Sandirasegaram:report} and Karhunen-Loeve Transform \cite{scroederDSP2002}, and estimation-theoretic templates \cite{Grenander98} have also been proposed. Likewise a wealth of decision engines has been proposed for target classification. Approaches range from model-based classifiers such as linear, quadratic and kernel discriminant analysis \cite{yu:kernelDiscriminant}, neural networks \cite{daniell1992}, to machine learning-based classifiers such as SVM \cite{Zhao01} and its variations \cite{Casasent05} and boosting-based classifiers \cite{boosting:Sun07}.

Here we use three distinct target image representations derived from wavelet basis functions, which have been used widely in ATR applications \cite{Simard99,Sandirasegaram:report}. In particular, the LL, LH and HL sub-bands obtained after a multi-level wavelet decomposition using 2-D reverse biorthogonal wavelets \cite{Sandirasegaram:report} comprise the three sets of complementary features $\{\vect{y}_i\},i=1,2,3$. The spectra of natural images are known to have a power law fall-off, with most of the energy concentrated at low frequencies. The LL sub-band encapsulates this low frequency information. On the other hand, the HL and LH sub-bands encode discriminatory high frequency information. Our assumption about the complementary nature of target image representations is still valid for this choice of basis functions.

The design of composite classifiers for ATR is an area of active research interest. Paul {\em et al.} \cite{Paul03} combine outputs from eigen-template based matched filters and hidden Markov models (HMM)-based clustering using a product-of-classification-probabilities rule. More recently, Gomes {\em et al.} \cite{Gomes08} proposed simple voting combinations of individual classifier decisions. Invariably, the aforementioned methods use educated heuristics in combining decisions from multiple decision engines while advocating the choice of a {\em fixed} set of features. In \cite{Kittler98}, a comprehensive review of popular classification techniques is provided in addition to useful theoretical justifications for the improved performance of such ensemble classifiers. In \cite{boosting:Sun07}, the best set of features is adaptively learned from a collection of two different types of features. We have also proposed a two-stage meta-classification framework \cite{Srinivas11}, wherein the vector of `soft' outputs from multiple classifiers is interpreted as a meta-feature vector and fed to a second classification stage to obtain the final class decision.

Related research in multi-sensor ATR \cite{rizvi:ATRfusion,papson:ImageFusion,kwon:hyperspectralATR} has investigated the idea of combining different ``looks'' of the same scene for improved detection and classification. The availability of accurately geo-located, multi-sensor data has created new opportunities for the exploration of multi-sensor target detection and classification algorithms. Perlovsky {\em et al.} \cite{Perlovsky95} exploit multi-sensor ATR information to distinguish between friend and foe, using a neural network-based classifier. In \cite{nasrabadi:ATR}, a simple non-linear kernel-based fusion technique is applied to SAR and hyper-spectral data obtained from the same spatial footprint for target detection using anomaly detectors. Papson {\em et al.} \cite{papson:ImageFusion} focus on image fusion of SAR and ISAR imagery for enhanced characterization of targets.

While such intuitively motivated fusion methods have yielded benefits in ATR, a fundamental understanding of the relationships between heterogeneous and complementary sources of data has not been satisfactorily achieved yet. Our approach is an attempt towards gaining more insight into these inter-feature relationships with the aim of improving classification performance.

\section{Experiments and Results}

\begin{table}[t]
  \caption[Training images from the MSTAR data set used for experiments]{Training images used for experiments. All images are taken from the MSTAR data set.}
  \begin{center}
  \begin{tabular}{|c|c|c|c|}
    \hline
    \hline
    Target & Vehicles & Number of images  & Depression angle \\
     \hline
     \hline
    BMP-2 & 9563, 9566, c21 & 299 & 17$^\circ$\\
    \hline
    BTR-70 & c71 & 697 & 17$^\circ$ \\
    \hline
    T-72 & 132, 812, s7 & 298 & 17$^\circ$ \\
    \hline
    BTR-60 & k10yt7532 & 256 &17$^\circ$\\
    \hline
    2S1 & b01 & 233 &17$^\circ$ \\
    \hline
    BRDM-2 & E-71 & 299 &17$^\circ$ \\
    \hline
    D7 & 92v13015 & 299 &17$^\circ$ \\
    \hline
    T62 & A51 & 691 & 17$^\circ$ \\
    \hline
    ZIL131 & E12 & 299 & 17$^\circ$ \\
    \hline
    ZSU234 & d08 & 299 & 17$^\circ$ \\
    \hline
  \end{tabular}
  \end{center}
  \label{tab:mstar_trng}
\end{table}

\begin{table}[t]
  \caption[Test images from the MSTAR data set used for experiments]{Test images used for experiments. All images are taken from the MSTAR data set.}
  \begin{center}
  \begin{tabular}{|c|c|c|c|}
    \hline
    \hline
    Target & Vehicles & Number of images  & Depression angle \\
     \hline
     \hline
    BMP-2 &  9563, 9566, c21 & 587 & 15$^\circ$ \\
    \hline
    BTR-70 & c71 & 196 & 15$^\circ$ \\
    \hline
    T-72 & 132, 812, s7 & 582 & 15$^\circ$ \\
    \hline
    BTR-60 & k10yt7532 & 195 & 15$^\circ$ \\
    \hline
    2S1 & b01& 274 & 15$^\circ$ \\
    \hline
    BRDM-2 & E-71 & 263 &  15$^\circ$ \\
    \hline
    D7 & 92v13015 & 274&15$^\circ$ \\
    \hline
    T62 & A51 & 582& 15$^\circ$ \\
    \hline
    ZIL131 & E12& 274 &15$^\circ$ \\
    \hline
    ZSU234 & d08 & 274 & 15$^\circ$ \\
    \hline
  \end{tabular}
  \end{center}
  \label{tab:mstar_test}
\end{table}

We test the proposed framework on the benchmark MSTAR data set under a variety of operating conditions. It is well-known that pose estimation can significantly improve classification performance. Accordingly, we use the pose estimation technique from \cite{boosting:Sun07} in our framework as a pre-processing step. The overall performance of our proposed approach is better than that of well-known competing techniques in the literature \cite{Zhao01,boosting:Sun07,Sullivan2001}. We also compare our approach (in the absence of pose estimation) with the template-based correlation filter method in \cite{Singh02} which is designed to be approximately invariant to pose.  Another important merit of our algorithm - robustness to limited training - is revealed by observing classification performance as a function of training set size, a comparison not usually performed for ATR problems although it has high practical significance.

\begin{table}[t]
  \caption[Test images used for the version variant testing set EOC-1]{Test images used for the version variant testing set EOC-1 in Section \ref{sec:rec_acc}.}
  \begin{center}
  \begin{tabular}{|c|c|c|c|}
    \hline
     Target & Serial number & Depression angle &  Number of images\\
    \hline
    T-72 & s7 & 15$^\circ$, 17$^\circ$& 419\\
    \hline
    T-72 & A32 & 15$^\circ$, 17$^\circ$ & 572\\
    \hline
    T-72 & A62 & 15$^\circ$, 17$^\circ$ & 573\\
    \hline
    T-72 & A63 & 15$^\circ$, 17$^\circ$  & 573\\
    \hline
    T-72 & A64 & 15$^\circ$, 17$^\circ$  & 573\\
    \hline
  \end{tabular}
  \end{center}
  \label{tab:ver_var}
\end{table}

\subsection{Experimental Set-up}
\label{ch2:expt_setup}

Our experiments are performed on magnitude SAR images obtained from the benchmark MSTAR program \cite{mstar}, which has released a large set of SAR images in the public domain. These consist of one-foot resolution X-band SAR images of ten different vehicle classes as well as separate clutter and confuser images. Target images are captured under varying operating conditions including different depression angles, aspect angles, serial numbers, and articulations. This collection of images, referred to as the MSTAR data set, is suitable for testing the performance of ATR algorithms since the number of images is statistically significant, ground truth data for acquisition conditions are available and a variety of targets have been imaged. Standard experimental conditions and performance benchmarks to compare classification experiments using the MSTAR database have been provided in \cite{ross:MSTARprep}.

The ten target classes from the database used for our experiments are: T-72, BMP-2, BTR-70, BTR-60, 2S1, BRDM-2, D7, T62, ZIL131, and ZSU234. Tables \ref{tab:mstar_trng} and \ref{tab:mstar_test} list the various target classes with vehicle variant descriptions, number of images per class available for training and testing, as well as the depression angle. Each target chip - the processed image input to the classification algorithm - is normalized to a $64 \times 64$ region of interest.

Images for training corresponding to all the ten vehicle classes are acquired at $17^\circ$ depression angle. We consider three different test scenarios in our experiments. Under standard operating conditions (SOC) \cite{ross:MSTARprep}, we test the algorithms with images from all ten classes as listed in Table \ref{tab:mstar_test}. These images are of vehicles with the same serial numbers as those in the training set, captured at depression angles of $15^\circ$. The other two test scenarios pertain to extended operating conditions (EOC). Specifically, we first consider a four-class problem (denoted by EOC-1) with training images chosen from BMP-2, BTR-70, BRDM-2, and T-72 as listed in Table \ref{tab:mstar_trng}, while the test set comprises five version variants of T-72 as described in Table \ref{tab:ver_var}. This is consistent with the version variant testing scenario in \cite{Sullivan2001}. We also compare algorithm performance for another test set (EOC-2) comprising four vehicles (2S1, BRDM2, T72, ZSU234) with the same serial numbers as the training group but acquired at $30^\circ$ depression angle. This EOC is consistent with the Test Group 3 in \cite{Singh02}. It is well-known that test images acquired with depression angle different from the training set are harder to classify \cite{Mossing98}.

\noindent \textbf{Pose estimation:} The target images are acquired with pose angles randomly varying between $0^\circ$ and $360^\circ$. Eliminating variations in pose can lead to significant improvement in overall classification performance. Many pose estimation approaches have been proposed for the ATR problem (\cite{pose:Zhao98} for example). A few other approaches \cite{Sullivan2001,boosting:Sun07} have incorporated a pose estimator within the target recognition framework. On the other hand, template-based approaches like \cite{Mahalanobis98,Singh02} are designed to be invariant to pose variations. Here, we use the pose estimation technique proposed in \cite{boosting:Sun07}. The pre-processed chip is first filtered using a Sobel edge detector to identify target edges, followed by an exhaustive target-pose search over different pose angles. The details of the pose estimation process are available in \cite{boosting:Sun07}.

We compare our proposed Iterative Graph Thickening (IGT) approach with four widely cited methods in ATR literature:
\begin{enumerate}
  \item {\em EMACH}: the extended maximum average correlation height filter in \cite{Singh02}
  \item {\em SVM}: support vector machine classifier in \cite{Zhao01}
  \item {\em CondGauss}: conditional Gaussian model in \cite{Sullivan2001}
  \item {\em AdaBoost}: feature fusion via boosting on RBF net classifiers \cite{boosting:Sun07}.
\end{enumerate}

In the subsequent sections, we provide a variety of experimental results to demonstrate the merits of our IGT approach compared to existing methods. First, in Section \ref{sec:rec_acc}, we present confusion matrices, shown in Tables \ref{tab:corrfilt}-\ref{tab:igtpose_eoc2}, for the SOC and EOC scenarios. The confusion matrix is commonly used in ATR literature to represent classification performance. Each element of the confusion matrix gives the probability of classification into one of the target classes. Each row corresponds to the true class of the target image, and each column corresponds to the class chosen by the classifier. In Section \ref{sec:roc_atr}, we test the outlier rejection performance of our proposed approach via ROC plots. Finally, we evaluate the performance of the five approaches as a function of training set size in Section \ref{sec:trng_size_atr}.

\subsection{Recognition Accuracy}
\label{sec:rec_acc}

\subsubsection{Standard Operating Condition (SOC)} 
Tables \ref{tab:corrfilt}-\ref{tab:igtnopose} show results for the SOC scenario. As discussed earlier, estimating the pose of the target image can lead to improvements in classification performance. The \emph{SVM}, \emph{CondGauss}, and \emph{AdaBoost} approaches chosen for comparison in our experiments incorporate pose estimation as a pre-processing step before classification. However, the EMACH filter \cite{Singh02} is designed to be inherently invariant to pose. Accordingly, we consider two specific experimental cases: 

\noindent \textbf{With pose estimation:}
\label{sec:pose_est}
In this scenario, pose estimation is performed in all approaches other than EMACH. For our IGT framework, we use the same pose estimator from \cite{boosting:Sun07}. The confusion matrices are presented in Tables \ref{tab:corrfilt}-\ref{tab:igtpose}. The proposed approach results in better overall classification performance in comparison to the existing approaches. The classification rates in Tables \ref{tab:corrfilt}-\ref{tab:igtpose} are consistent with values reported in literature \cite{Zhao01,Singh02,boosting:Sun07,Sullivan2001}.\\

\noindent \textbf{No pose estimation}
\label{sec:no_pose_est}
For fairness of comparison with EMACH, we now compare performance in the absence of explicit pose estimation. The confusion matrices are presented in Tables \ref{tab:svmnopose}-\ref{tab:igtnopose}. Comparing these results with the confusion matrix in Table \ref{tab:corrfilt}, we observe that the EMACH filter, unsurprisingly, gives the best overall performance among the existing approaches, thereby establishing its robustness to pose invariance. Significantly, the classification accuracy of the IGT framework (without pose estimation) is comparable to EMACH.\\

\subsubsection{Extended Operating Conditions (EOC)} 
We now compare algorithm performance using more difficult test scenarios. Here, we do not provide separate results with and without pose estimation, and each existing approach being compared is chosen with its best settings (i.e. methods other than EMACH incorporate pose estimation). It must be mentioned however that the overall trends in the absence of pose estimation are similar to those observed for the SOC.

For the EOC-1 test set, the confusion matrices are presented in Tables \ref{tab:corrfilt_eoc1}-\ref{tab:igtpose_eoc1}. The corresponding confusion matrices for the EOC-2 test set are shown in Tables \ref{tab:corrfilt_eoc2}-\ref{tab:igtpose_eoc2}. In each test scenario, we see that IGT consistently outperforms the existing techniques. The average classification accuracies of the five methods for the different experimental scenarios are compared in Table \ref{tab:avg_rates}.

\subsection{ROC Curves: Outlier Rejection Performance}
\label{sec:roc_atr}
Perhaps the first work to address the problem of clutter and confuser rejection in SAR ATR is \cite{casasent06}. In this work, the outlier rejection capability of the EMACH filter \cite{Singh02} is demonstrated for a subset of the MSTAR data with three target classes - BMP2, BTR70, and T72. The D7 and 2S1 classes are treated as confusers. Extensions of this work include experiments on all ten MSTAR classes with Gaussian kernel SVM as classifier \cite{yuan06}, and similar comparisons using the Minace filter \cite{patnaik07}. In each case, no clutter or confuser images are used in the training phase. Our proposed framework can also be used to detect outliers in the data set using the decision rule in \eqref{eq:decision_rule_multi}. The likelihood ratios from each of the $K$ individual problems are compared against an experimentally determined threshold $\tau_{\mbox{\tiny{out}}}$. If all the $K$ likelihood values are lower than this threshold, the corresponding target image is identified as an outlier (confuser vehicle or clutter).

We use the SOC test set and include new confuser images - provided in the MSTAR database - in the test set. We consider a binary classification problem with the two classes being target and confuser. We also test the ability of the competing approaches to reject clutter by considering another experiment where we use clutter images from MSTAR instead of confuser vehicles. This experimental scenario is consistent with the ROC curves provided in \cite{Singh02}. We compute the probability of detection ($P_d$) and probability of false alarm ($P_{fa}$) using the threshold described in Section \ref{ch2:discgraphs}. Fig. \ref{fig:roc_atr}(a) shows the ROCs for the five methods for the target vs. confuser problem. The corresponding ROCs for target vs. clutter are shown in Fig. \ref{fig:roc_atr}(b). In both cases, the improved performance of IGT is readily apparent. A visual comparison also shows the improvements in IGT over the results in \cite{yuan06},\cite{patnaik07}.

\subsection{Classification Performance as Function of Training Size}
\label{sec:trng_size_atr}

Unlike some other real-world classification problems, ATR suffers from the drawback that the training images corresponding to one or more target classes may be limited. To illustrate, the typical dimension of a SAR image in the MSTAR data set is $128 \times 128 = 16384$. Even after cropping and normalization to $64 \times 64$ the data size is $4096$ coefficients. In comparison, the number of training SAR images per class is in the 50-250 range (see Table \ref{tab:mstar_test}). The Hughes phenomenon \cite{Hughes_1968} highlights the difficulty of learning models for high-dimensional data under limited training. So in order to test the robustness of various ATR algorithms in the literature to low training, we revisit the SOC and EOC experiments from Section \ref{sec:rec_acc}, and plot the overall classification accuracy as a function of training set size. The corresponding plots are shown in Figs.\ \ref{fig:trng_size_soc} and \ref{fig:trng_size_eoc}.

Figs.\ \ref{fig:trng_size_soc}(a)-\ref{fig:trng_size_soc}(b) show the variation in classification error as a function of training sample size for the SOC, both with and without pose estimation. For each value of training size, the experiment was performed using ten different random subsets of training vectors and the average error probability values are reported. The IGT algorithm consistently offers the lowest error rates in all regimes of training size. For large training (an unrealistic scenario for the ATR problem) as expected all techniques begin to converge. We also observe that our proposed algorithm exhibits a more graceful degradation with a decrease in training size vs. competing approaches. Similar trends can be inferred from the plots for the EOCs, shown in Figs. \ref{fig:trng_size_eoc}(a)-\ref{fig:trng_size_eoc}(b). Recognition performance as a function of training size in SAR ATR is a very significant practical issue and in this aspect, the use of probabilistic graphical models as decision engines offers appreciable benefits over existing alternatives based on SVMs \cite{Zhao01}, neural networks \cite{Sandirasegaram:report} etc. The superior performance of discriminative graphs in the low-training regime is attributed to the ability of the graphical structure to capture dominant conditional dependencies between features which are crucial to the classification task \cite{bishop:2006,wainwright:book08,tan:tsp10}.

\subsection{Summary of Results}
\label{sec:summary_results}
Here, we summarize the key messages from the various experiments described in Section \ref{sec:rec_acc} to \ref{sec:trng_size_atr}. First, we test the proposed IGT approach against competing approaches on the MSTAR SOC test set. We provide two types of results, with and without pose estimation. When pose estimation is explicitly included as a pre-processing step, the \emph{AdaBoost}, \emph{CondGauss}, and \emph{SVM} methods perform better overall compared to the EMACH filter, although the EMACH filter gives better results for specific vehicle classes. Overall, the performance of the IGT method is better than all the competing approaches. Since the EMACH filter is designed to be invariant to pose, we also provide confusion matrices for the scenario where pose is not estimated prior to classification using the competing approaches. Here, we observe that the EMACH filter and IGT perform best overall, while the other approaches suffer a significant degradation in performance. These two experiments demonstrate that IGT exhibits an inherent invariance to pose, to some extent. This can be explained due to the choice of wavelet features which are known to be less sensitive to image rotations, as well as the fact that the graphical models learn better from the available training image data compared to the other approaches. Similar trends hold for the harder EOC test scenarios too. It must be noted that the classification rates reported in the confusion matrices for the various approaches are consistent with values reported in literature \cite{Zhao01,Singh02,boosting:Sun07,Sullivan2001}.

Next, we introduce a comparison of classification performance as a function of training set size. While all methods perform appreciably when provided with large amount of training (representative of asymptotic scenario), the proposed IGT method exhibits a much more graceful decay in performance as the number of training samples per class is reduced. As Figs. \ref{fig:trng_size_soc}(a) and \ref{fig:trng_size_soc}(b) reveal, this is true for both the cases of with and without pose estimation. This is the first such explicit comparison in SAR ATR to the best of our knowledge. Finally, we test the outlier rejection performance of the approaches by plotting the ROCs. Here too, the overall improved performance of IGT is apparent. 

\section{Conclusion}
\label{ch2:concl}
The value of complementary feature representations and decision engines is well appreciated by the ATR community as confirmed by a variety of fusion approaches, which use high-level features, or equivalently, the outputs of individual classifiers. We leverage a recent advance in discriminative graph learning to explicitly capture dependencies between different competing sets of low-level features for the SAR target classification problem. Our algorithm learns tree-structured graphs on the LL, LH and HL wavelet sub-band coefficients extracted from SAR images and thickens them via boosting. The proposed framework readily generalizes to any suitable choice of feature sets that offer complementary benefits. Our algorithm is particularly effective in the challenging regime of low training and high dimensional data, a serious practical concern for ATR systems. Experiments on SAR images from the benchmark MSTAR data set demonstrate the success of our approach over well-known target classification algorithms.

Looking ahead, multi-sensor ATR \cite{Perlovsky95} offers fertile ground for the application of our feature fusion framework. Of related significance is the problem of feature selection \cite{guyon03,liu05}. Several hypotheses could be developed and the relative statistical significance of inter-relationships between learned target features can be determined via feature selection techniques.

\begin{table}[t]
\caption[Confusion matrix for SOC: EMACH correlation filter]{Confusion matrix for SOC: EMACH correlation filter\cite{Singh02}.}
\begin{center}
\scriptsize{
\begin{tabular}{|c|c|c|c|c|c|c|c|c|c|c|}
  \hline
  Class & BMP-2 & BTR-70 & T-72 & BTR-60 & 2S1 & BRDM-2 & D7 & T62 & ZIL131 & ZSU234 \\
  \hline
  BMP-2 & {\bf{0.90}} & 0.02 & 0.04 & 0.01 & 0.01 & 0 & 0 & 0.01 & 0 & 0.01 \\
   \hline
  BTR-70 & 0.02 & {\bf{0.93}} & 0.01 & 0 & 0.01 & 0.01 & 0 & 0.02 & 0 & 0 \\
  \hline
  T-72 & 0.02 & 0 & {\bf{0.96}}& 0 & 0.01 & 0&0 &0 & 0.01 & 0 \\
  \hline
  BTR-60 &0 & 0.01 &0 &{\bf{0.95}} &0.01 & 0 &0.03 & 0 &0 &0 \\
  \hline
  2S1 & 0.05 & 0.06 & 0.04 & 0.02 &{\bf{0.74}} & 0.03 & 0.01 & 0.02 & 0.01 & 0.02 \\
  \hline
  BRDM-2 & 0.03 & 0.06 & 0.03 &0 & 0.01 &{\bf{0.84}} & 0.02 &0  & 0 & 0.01\\
  \hline
  D7 & 0.02& 0.03 & 0.02& 0.01&0 &0 &{\bf{0.85}} & 0.03& 0.02& 0.02\\
  \hline
  T62 & 0.01&0.01 &0.01 & 0.01& 0.04&0 & 0&{\bf{0.86}} & 0.04& 0.02 \\
  \hline
  ZIL131 &0.02  & 0 & 0.01 & 0.02 & 0 &0 &0 & 0.04&{\bf{0.88}} & 0.03\\
  \hline
  ZSU234 & 0.01 & 0 & 0.04 & 0.02 & 0 & 0& 0& 0.01&0 & {\bf{0.92}}\\
  \hline
\end{tabular}
}
\end{center}
\label{tab:corrfilt}
\end{table}

\begin{table}[t]
\caption[Confusion matrix for SOC: SVM classifier with pose estimation]{Confusion matrix for SOC: SVM classifier \cite{Zhao01} with pose estimation.}
\begin{center}
\scriptsize{
\begin{tabular}{|c|c|c|c|c|c|c|c|c|c|c|}
  \hline
  Class & BMP-2 & BTR-70 & T-72 & BTR-60 & 2S1 & BRDM-2 & D7 & T62 & ZIL131 & ZSU234 \\
  \hline
  BMP-2 & {\bf{0.90}} & 0.02  & 0.03 & 0.01 & 0.01 &0.02 & 0& 0& 0.01&0 \\
   \hline
  BTR-70 & 0.03& {\bf{0.90}} & 0.03& 0&0 & 0.02& 0&0 & 0& 0.02\\
  \hline
  T-72 &0.02 &0.01 & {\bf{0.93}}& 0.03& 0& 0& 0& 0&0.01 &0 \\
  \hline
  BTR-60 & 0.02& 0.02&0.01 &{\bf{0.92}} & 0&0 &0.03 & 0&0 & 0\\
  \hline
  2S1 & 0.05& 0.03& 0.02& 0&{\bf{0.81}} & 0.03& 0.02&0.03 &0 &0.01 \\
  \hline
  BRDM-2 & 0.06& 0.08& 0.02& 0.01& 0&{\bf{0.79}} &0 &0.03 &0 &0.01 \\
  \hline
  D7 & 0& 0& 0& 0& 0.01&0 &{\bf{0.98}} &0 &0 &0.01 \\
  \hline
  T62 &0.01 &0 &0 &0.01 & 0& 0& 0&{\bf{0.91}} & 0.04& 0.03\\
  \hline
  ZIL131 & 0.02 & 0.01 & 0 & 0 & 0 & 0& 0& 0&{\bf{0.95}} & 0.02\\
  \hline
  ZSU234 & 0 & 0.01 & 0 & 0.03 & 0 & 0& 0.01&0 & 0.03& {\bf{0.92}}\\
  \hline
\end{tabular}
}
\end{center}
\label{tab:svmpose}
\end{table}

\begin{table}[t]
\caption[Confusion matrix for SOC: Conditional Gaussian model with pose estimation]{Confusion matrix for SOC: Conditional Gaussian model \cite{Sullivan2001} with pose estimation.}
\begin{center}
\scriptsize{
\begin{tabular}{|c|c|c|c|c|c|c|c|c|c|c|}
  \hline
  Class & BMP-2 & BTR-70 & T-72 & BTR-60 & 2S1 & BRDM-2 & D7 & T62 & ZIL131 & ZSU234 \\
  \hline
  BMP-2 & {\bf{0.93}} & 0.01  & 0.02 & 0 & 0.01 &0.02 & 0& 0& 0.01&0 \\
   \hline
  BTR-70 & 0.03& {\bf{0.91}} & 0.02& 0&0 & 0.02& 0&0 & 0& 0.02\\
  \hline
  T-72 &0.02 &0.01 & {\bf{0.95}}& 0.01& 0& 0& 0& 0&0.01 &0 \\
  \hline
  BTR-60 & 0.02& 0&0.01 &{\bf{0.95}} & 0&0 &0.02 & 0&0 & 0\\
  \hline
  2S1 & 0.03& 0.04& 0.01& 0&{\bf{0.87}} & 0.02& 0&0.02 &0 &0.01 \\
  \hline
  BRDM-2 & 0.04& 0.03&0.01 &0 &0 &{\bf{0.89}} &0.03 & 0& 0& 0\\
  \hline
  D7 & 0& 0& 0& 0& 0.01&0 &{\bf{0.98}} &0 &0 &0.01 \\
  \hline
  T62 &0.01 &0 &0 &0 & 0& 0& 0&{\bf{0.92}} & 0.05& 0.02\\
  \hline
  ZIL131 & 0.02 & 0.02 & 0 & 0 & 0 & 0& 0& 0&{\bf{0.95}} & 0.01\\
  \hline
  ZSU234 & 0.01 & 0.02 & 0 & 0.01 & 0 & 0.01& 0& 0&0.02 & {\bf{0.93}}\\
  \hline
\end{tabular}
}
\end{center}
\label{tab:condgausspose}
\end{table}

\begin{table}[t]
\caption[Confusion matrix for SOC: AdaBoost with pose estimation]{Confusion matrix for SOC: AdaBoost\cite{boosting:Sun07} with pose estimation.}
\begin{center}
\scriptsize{
\begin{tabular}{|c|c|c|c|c|c|c|c|c|c|c|}
  \hline
  Class & BMP-2 & BTR-70 & T-72 & BTR-60 & 2S1 & BRDM-2 & D7 & T62 & ZIL131 & ZSU234 \\
  \hline
  BMP-2 & {\bf{0.92}} & 0.02  & 0.02 & 0 & 0.01 &0.02 & 0& 0& 0.01&0 \\
   \hline
  BTR-70 & 0.03& {\bf{0.93}} & 0& 0&0 & 0.02& 0&0 & 0& 0.02\\
  \hline
  T-72 &0.02 &0.01 & {\bf{0.96}}& 0.01& 0& 0& 0& 0&0 &0 \\
  \hline
  BTR-60 & 0.02& 0&0.02 &{\bf{0.93}} & 0&0 &0.03 & 0&0 & 0\\
  \hline
  2S1 & 0.03& 0.04& 0.01& 0&{\bf{0.87}} & 0.02& 0&0.02 &0 &0.01 \\
  \hline
  BRDM-2 & 0.05& 0.03&0.02 &0 &0 &{\bf{0.85}} &0.05 & 0& 0& 0\\
  \hline
  D7 & 0& 0& 0& 0& 0.01&0 &{\bf{0.98}} &0 &0 &0.01 \\
  \hline
  T62 &0.01 &0 &0 &0 & 0& 0& 0&{\bf{0.93}} & 0.03& 0.03\\
  \hline
  ZIL131 & 0.02 & 0.02 & 0 & 0 & 0 & 0& 0& 0&{\bf{0.94}} & 0.02\\
  \hline
  ZSU234 & 0.01 & 0 & 0 & 0.01 & 0 & 0& 0& 0&0.02 & {\bf{0.96}}\\
  \hline
\end{tabular}
}
\end{center}
\label{tab:adaboostpose}
\end{table}

\begin{table}[t]
\caption[Confusion matrix for SOC: Iterative Graph Thickening (IGT) with pose estimation]{Confusion matrix for SOC: Iterative Graph Thickening (IGT) with pose estimation.}
\begin{center}
\scriptsize{
\begin{tabular}{|c|c|c|c|c|c|c|c|c|c|c|}
  \hline
  Class & BMP-2 & BTR-70 & T-72 & BTR-60 & 2S1 & BRDM-2 & D7 & T62 & ZIL131 & ZSU234 \\
  \hline
  BMP-2 & {\bf{0.95}} & 0.01  & 0.01 & 0 & 0.01 &0.01 & 0& 0& 0.01&0 \\
   \hline
  BTR-70 & 0.02& {\bf{0.94}} & 0& 0&0 & 0.02& 0&0 & 0& 0.02\\
  \hline
  T-72 &0.02 &0.01 & {\bf{0.96}}& 0.01& 0& 0& 0& 0&0 &0 \\
  \hline
  BTR-60 & 0.01& 0&0.01 &{\bf{0.97}} & 0&0 &0.01 & 0&0 & 0\\
  \hline
  2S1 & 0.03& 0.04& 0.01& 0&{\bf{0.89}} & 0& 0&0.01 &0 &0.02 \\
  \hline
  BRDM-2 & 0.02& 0.01&0.04 &0 &0 &{\bf{0.90}} &0.02 & 0& 0& 0.01\\
  \hline
  D7 & 0& 0& 0& 0& 0.01&0 &{\bf{0.99}} &0 &0 &0 \\
  \hline
  T62 &0.01 &0 &0 &0 & 0& 0& 0&{\bf{0.95}} & 0.03& 0.01\\
  \hline
  ZIL131 & 0.02 & 0.02 & 0.01 & 0 & 0 & 0& 0& 0&{\bf{0.95}} & 0\\
  \hline
  ZSU234 & 0.01 & 0 & 0 & 0.01 & 0 & 0& 0& 0&0.02 & {\bf{0.96}}\\
  \hline
\end{tabular}
}
\end{center}
\label{tab:igtpose}
\end{table}


\begin{table}[t]
\caption[Confusion matrix for SOC: SVM classifier without pose estimation]{Confusion matrix for SOC: SVM classifier \cite{Zhao01} without pose estimation.}
\begin{center}
\scriptsize{
\begin{tabular}{|c|c|c|c|c|c|c|c|c|c|c|}
  \hline
  Class & BMP-2 & BTR-70 & T-72 & BTR-60 & 2S1 & BRDM-2 & D7 & T62 & ZIL131 & ZSU234 \\
  \hline
  BMP-2 & {\bf{0.84}} & 0.05 & 0.04 & 0.01 &0  & 0.02& 0& 0.03& 0 & 0.01\\
   \hline
  BTR-70 &0.02 & {\bf{0.83}} & 0& 0.05& 0.03& 0.02& 0.01& 0.01&0 &0.03 \\
  \hline
  T-72 & 0.03& 0.05& {\bf{0.87}}& 0& 0& 0.03& 0.01& 0&0 &0.01 \\
  \hline
  BTR-60 & 0.04& 0.04& 0&{\bf{0.86}} &0.03 &0 & 0.03& 0&0 &0 \\
  \hline
  2S1 & 0.06& 0.04& 0.03& 0.04&{\bf{0.75}} & 0.05& 0.01& 0.01&0.01 & 0\\
  \hline
  BRDM-2 &0.08 & 0.02&0.03 &0.02 & 0&{\bf{0.74}} &0 &0.09 & 0.01& 0.01\\
  \hline
  D7 & 0.03&0.02 & 0& 0& 0.02&0 &{\bf{0.91}} & 0.02& 0&0 \\
  \hline
  T62 &0.04 &0.03 &0.05 &0.01 &0 &0 &0 &{\bf{0.85}} &0.02 & 0\\
  \hline
  ZIL131 & 0.02 & 0.03 & 0.03 & 0 & 0 &0 &0 &0 &{\bf{0.89}} & 0.03\\
  \hline
  ZSU234 & 0.01 & 0 &  0& 0.02 & 0 &0 &0.08 &0.04 &0 & {\bf{0.85}}\\
  \hline
\end{tabular}
}
\end{center}
\label{tab:svmnopose}
\end{table}

\begin{table}[t]
\caption[Confusion matrix for SOC: Conditional Gaussian model without pose estimation]{Confusion matrix for SOC: Conditional Gaussian model \cite{Sullivan2001} without pose estimation.}
\begin{center}
\scriptsize{
\begin{tabular}{|c|c|c|c|c|c|c|c|c|c|c|}
  \hline
  Class & BMP-2 & BTR-70 & T-72 & BTR-60 & 2S1 & BRDM-2 & D7 & T62 & ZIL131 & ZSU234 \\
  \hline
  BMP-2 & {\bf{0.89}} & 0.03 & 0.02 & 0.01 &0  & 0.01& 0& 0.03& 0 & 0.01\\
   \hline
  BTR-70 &0.02 & {\bf{0.88}} & 0& 0.05& 0& 0& 0.03& 0.01&0 &0.01 \\
  \hline
  T-72 & 0.03& 0& {\bf{0.92}}& 0& 0& 0&0.03 &0 &0.02 &0 \\
  \hline
  BTR-60 & 0.02& 0&0.03 &{\bf{0.90}} & 0& 0&0.05 &0 &0 &0 \\
  \hline
  2S1 & 0.03& 0.03& 0.04& 0.01&{\bf{0.82}} &0 & 0&0.03 &0.03 &0.01 \\
  \hline
  BRDM-2 & 0.08&0.02 &0.05 & 0.01&0 &{\bf{0.75}} & 0.01&0.05 &0 &0.03 \\
  \hline
  D7 & 0.03& 0.02& 0& 0& 0&0 &{\bf{0.89}} & 0.04& 0& 0.02\\
  \hline
  T62 &0.03 & 0.03& 0.01& 0& 0& 0.05& 0&{\bf{0.84}} & 0.03& 0.01\\
  \hline
  ZIL131 & 0.03 & 0 & 0.03 & 0 & 0 & 0.05&0.01 &0.01 &{\bf{0.85}} & 0.02\\
  \hline
  ZSU234 & 0.01 & 0 & 0.02 & 0 & 0.03 &0 & 0&0.04 & 0.03& {\bf{0.87}}\\
  \hline
\end{tabular}
}
\end{center}
\label{tab:condgaussnopose}
\end{table}

\begin{table}[t]
\caption[Confusion matrix for SOC: AdaBoost without pose estimation]{Confusion matrix for SOC: AdaBoost\cite{boosting:Sun07} without pose estimation.}
\begin{center}
\scriptsize{
\begin{tabular}{|c|c|c|c|c|c|c|c|c|c|c|}
  \hline
  Class & BMP-2 & BTR-70 & T-72 & BTR-60 & 2S1 & BRDM-2 & D7 & T62 & ZIL131 & ZSU234 \\
  \hline
  BMP-2 & {\bf{0.88}} & 0.03 & 0.02 & 0.01 &0  & 0.01& 0& 0.04& 0 & 0.01\\
   \hline
  BTR-70 & 0.02 & {\bf{0.90}} &0.04 &0 & 0&0.03 &0 &0 & 0& 0.01\\
  \hline
  T-72 & 0.02&0.02 & {\bf{0.91}}&0.02 &0 & 0& 0& 0.01& 0.01& 0.01\\
  \hline
  BTR-60 & 0.04& 0.03&0.01 &{\bf{0.89}} & 0&0 &0 &0.01 &0 &0.02 \\
  \hline
  2S1 & 0.04&0.02 &0.04 & 0.02&{\bf{0.84}} & 0.01& 0&0.03 &0 &0 \\
  \hline
  BRDM-2 & 0.05&0.02 &0 &0.03 & 0&{\bf{0.81}} & 0.03& 0& 0.05& 0.01\\
  \hline
  D7 & 0.03& 0.02& 0& 0& 0&0 &{\bf{0.91}} & 0.04& 0& 0\\
  \hline
  T62 & 0.02&0.03 &0 &0.01 &0.02 &0 &0 &{\bf{0.86}} &0.05 &0.01 \\
  \hline
  ZIL131 & 0.03 & 0 & 0.01 & 0.02 & 0 &0.07 &0 &0 &{\bf{0.86}} & 0.01\\
  \hline
  ZSU234 & 0.05 & 0.02 & 0.01 & 0 & 0 & 0.01&0.03 & 0.01& 0.01& {\bf{0.86}}\\
  \hline
\end{tabular}
}
\end{center}
\label{tab:adaboostnopose}
\end{table}

\begin{table}[t]
  \caption[Confusion matrix for SOC: Iterative Graph Thickening (IGT) without pose estimation]{Confusion matrix for SOC: Iterative Graph Thickening (IGT) without pose estimation.}
  \begin{center}
  \scriptsize{
  \begin{tabular}{|c|c|c|c|c|c|c|c|c|c|c|}
    \hline
    Class & BMP-2 & BTR-70 & T-72 & BTR-60 & 2S1 & BRDM-2 & D7 & T62 & ZIL131 & ZSU234 \\
    \hline
    BMP-2 & {\bf{0.89}} & 0.03 & 0.02 & 0.01 &0  & 0.01& 0& 0.01& 0.01 & 0.01\\
     \hline
    BTR-70 & 0.01 & {\bf{0.91}} &0.04 &0 & 0&0.03 &0 &0 & 0& 0.01\\
    \hline
    T-72 & 0.02&0.01 & {\bf{0.93}}&0.02 &0 & 0& 0& 0.01& 0.01& 0\\
    \hline
    BTR-60 & 0.03& 0.02&0.01 &{\bf{0.92}} & 0&0 &0 &0.01 &0 &0.01 \\
    \hline
    2S1 & 0.04&0.02 &0.04 & 0.02&{\bf{0.77}} & 0.05& 0&0.03 &0 &0.03 \\
    \hline
    BRDM-2 & 0.01&0.03 &0 &0.03 & 0&{\bf{0.88}} & 0.02& 0& 0.02& 0.01\\
    \hline
    D7 & 0.01& 0.02& 0& 0.01& 0&0 &{\bf{0.89}} & 0.02& 0.05& 0\\
    \hline
    T62 &0 &0 & 0.04& 0.05& 0& 0& 0&{\bf{0.88}} & 0.01& 0.02\\
    \hline
    ZIL131 & 0.04 & 0 & 0.03 & 0 & 0 & 0.02& 0&0.02 &{\bf{0.88}} & 0.01\\
    \hline
    ZSU234 & 0.02 & 0 & 0.03 & 0 & 0 & 0.03& 0& 0.01&0 & {\bf{0.91}}\\
    \hline
  \end{tabular}
  }
  \end{center}
  \label{tab:igtnopose}
\end{table}
\begin{table}[]
\caption[Confusion matrix for EOC-1: EMACH]{Confusion matrix for EOC-1: EMACH\cite{Singh02}.}
\begin{center}
\begin{tabular}{|c|c|c|c|c|}
  \hline
  Class & BMP-2 & BTR-70 & BRDM-2 & T-72\\
  \hline
  T-72 s7 & 0.04 & 0.08 & 0.06 & {\bf{0.82}} \\
   \hline
  T-72 s7 & 0.09& 0.05 & 0.05&{\bf{0.81}} \\
  \hline
  T-72 62 & 0.08 & 0.06 & 0.03 & {\bf{0.83}}\\
  \hline
  T-72 63 & 0.13 & 0.06 & 0.11 & {\bf{0.70}}\\
  \hline
  T-72 64 & 0.16 & 0.04 & 0.12 & {\bf{0.68}}\\
  \hline
\end{tabular}
\end{center}
\label{tab:corrfilt_eoc1}
\vspace{9mm}
\end{table}

\begin{table}[]
\caption[Confusion matrix for EOC-1: SVM classifier]{Confusion matrix for EOC-1: SVM classifier \cite{Zhao01}.}
\begin{center}
\begin{tabular}{|c|c|c|c|c|}
  \hline
  Class & BMP-2 & BTR-70 & BRDM-2 & T-72\\
  \hline
  T-72 s7 & 0.05 & 0.04 & 0.04 & {\bf{0.87}}  \\
   \hline
  T-72 s7 & 0.07& 0.05  & 0.02 & {\bf{0.86}} \\
  \hline
  T-72 62 & 0.05& 0.05& 0.06& {\bf{0.84}}  \\
  \hline
  T-72 63 & 0.15 & 0.06 & 0.03& {\bf{0.76}}\\
  \hline
  T-72 64 & 0.09& 0.05&0.13 & {\bf{0.73}}\\
  \hline
\end{tabular}
\end{center}
\label{tab:svm_eoc1}
\vspace{9mm}
\end{table}

\begin{table}[t]
\caption[Confusion matrix for EOC-1: Conditional Gaussian model]{Confusion matrix for EOC-1: Conditional Gaussian model \cite{Sullivan2001}.}
\begin{center}
\begin{tabular}{|c|c|c|c|c|}
  \hline
  Class & BMP-2 & BTR-70 & BRDM-2 & T-72\\
  \hline
  T-72 s7 &0.09 &  0.02& 0.02 & {\bf{0.87}} \\
   \hline
  T-72 s7 & 0.05& 0.05 &0.06 &{\bf{0.84}} \\
  \hline
  T-72 62 & 0.07& 0.06& 0.06& {\bf{0.81}}\\
  \hline
  T-72 63 &0.08 &0.05 &0.12 & {\bf{0.75}}\\
  \hline
  T-72 64 & 0.10& 0.08& 0.09& {\bf{0.73}}\\
  \hline
\end{tabular}
\end{center}
\label{tab:condgauss_eoc1}
\vspace{9mm}
\end{table}

\begin{table}[t]
\caption[Confusion matrix for EOC-1: AdaBoost]{Confusion matrix for EOC-1: AdaBoost\cite{boosting:Sun07}.}
\begin{center}
\begin{tabular}{|c|c|c|c|c|}
  \hline
  Class & BMP-2 & BTR-70 & BRDM-2 & T-72\\
  \hline
  T72 s7 & 0.04 & 0.06 & 0.02 & {\bf{0.88}}\\
   \hline
  T-72 s7 & 0.05& 0.08& 0.03& {\bf{0.84}} \\
  \hline
  T-72 62 & 0.06&0.05 & 0.04& {\bf{0.85}}\\
  \hline
  T-72 63 & 0.11& 0.09& 0.04& {\bf{0.76}}\\
  \hline
  T-72 64 & 0.10& 0.05& 0.09& {\bf{0.76}}\\
  \hline
\end{tabular}
\end{center}
\label{tab:adaboost_eoc1}
\vspace{9mm}
\end{table}

\begin{table}[t]
\caption[Confusion matrix for EOC-1: IGT]{Confusion matrix for EOC-1: IGT.}
\begin{center}
\begin{tabular}{|c|c|c|c|c|}
  \hline
  Class & BMP-2 & BTR-70 & BRDM-2 & T-72\\
  \hline
  T-72 s7 & 0.05 & 0.04 & 0.03 & {\bf{0.88}} \\
   \hline
  T-72 s7 & 0.06 &0.03  & 0.02& {\bf{0.89}}\\
  \hline
  T-72 62 & 0.05& 0.04& 0.04& {\bf{0.87}}\\
  \hline
  T-72 63 & 0.07& 0.05& 0.07& {\bf{0.81}}\\
  \hline
  T-72 64 & 0.11& 0.04& 0.06& {\bf{0.79}}\\
  \hline
\end{tabular}
\end{center}
\label{tab:igtpose_eoc1}
\vspace{9mm}
\end{table}

\begin{table}[t]
\caption[Confusion matrix for EOC-2: EMACH]{Confusion matrix for EOC-2: EMACH\cite{Singh02}.}
\begin{center}
\begin{tabular}{|c|c|c|c|c|}
  \hline
  Class & 2S1 & BRDM-2 & T-72 & ZSU234\\
  \hline
  2S1 & {\bf{0.67}} & 0.15 & 0.12 & 0.06 \\
   \hline
  BRDM-2 & 0.17& {\bf{0.57}} & 0.19& 0.07\\
  \hline
  T-72 &0.07 & 0.09& {\bf{0.66}}& 0.18\\
  \hline
  ZSU234 & 0.10& 0.07& 0.02& {\bf{0.81}}\\
  \hline
\end{tabular}
\end{center}
\label{tab:corrfilt_eoc2}
\end{table}

\begin{table}[t]
\caption[Confusion matrix for EOC-2: SVM classifier]{Confusion matrix for EOC-2: SVM classifier \cite{Zhao01}.}
\begin{center}
\begin{tabular}{|c|c|c|c|c|}
  \hline
  Class & 2S1 & BRDM-2 & T-72 & ZSU234\\
  \hline
  2S1 & {\bf{0.74}} & 0.08 & 0.09 &0.09  \\
   \hline
  BRDM-2 & 0.12& {\bf{0.66}} & 0.09& 0.13\\
  \hline
  T-72 & 0.17& 0.06& {\bf{0.73}}& 0.04\\
  \hline
  ZSU234 & 0.07& 0.05&0.03 & {\bf{0.85}}\\
  \hline
\end{tabular}
\end{center}
\label{tab:svm_eoc2}
\end{table}

\begin{table}[t]
\caption[Confusion matrix for EOC-2: Conditional Gaussian model]{Confusion matrix for EOC-2: Conditional Gaussian model \cite{Sullivan2001}.}
\begin{center}
\begin{tabular}{|c|c|c|c|c|}
  \hline
  Class & 2S1 & BRDM-2 & T-72 & ZSU234\\
  \hline
  2S1 & {\bf{0.75}} & 0.09 &0.08  & 0.08 \\
   \hline
  BRDM-2 &0.14 & {\bf{0.69}} &0.11 & 0.06\\
  \hline
  T-72 & 0.16& 0.05& {\bf{0.72}}& 0.07\\
  \hline
  ZSU234 & 0.05&0.05 & 0.04& {\bf{0.86}}\\
  \hline
\end{tabular}
\end{center}
\label{tab:condgauss_eoc2}
\end{table}

\begin{table}[t]
\caption[Confusion matrix for EOC-2: AdaBoost]{Confusion matrix for EOC-2: AdaBoost\cite{boosting:Sun07}.}
\begin{center}
\begin{tabular}{|c|c|c|c|c|}
  \hline
  Class & 2S1 & BRDM-2 & T-72 & ZSU234\\
  \hline
  2S1 & {\bf{0.77}} & 0.05 & 0.11 & 0.07 \\
   \hline
  BRDM-2 &0.15 & {\bf{0.73}} & 0.05& 0.07\\
  \hline
  T-72 & 0.11& 0.09& {\bf{0.75}}& 0.05\\
  \hline
  ZSU234 & 0.04& 0.06& 0.02& {\bf{0.88}}\\
  \hline
\end{tabular}
\end{center}
\label{tab:adaboost_eoc2}
\end{table}

\begin{table}[t]
\caption[Confusion matrix for EOC-2: IGT]{Confusion matrix for EOC-2: IGT.}
\begin{center}
\begin{tabular}{|c|c|c|c|c|}
  \hline
  Class & 2S1 & BRDM-2 & T-72 & ZSU234\\
  \hline
  2S1 & {\bf{0.78}} & 0.06 & 0.09 & 0.07 \\
   \hline
  BRDM-2 & 0.15& {\bf{0.76}} & 0.06& 0.03\\
  \hline
  T-72 & 0.10& 0.09& {\bf{0.78}}& 0.03\\
  \hline
  ZSU234 & 0.05& 0.05& 0.02& {\bf{0.88}}\\
  \hline
\end{tabular}
\end{center}
\label{tab:igtpose_eoc2}
\end{table}

\begin{table}[t]
\caption[Average classification accuracy]{Average classification accuracy.}
\begin{center}
\begin{tabular}{|c|c|c|c|c|}
  \hline
  Class & SOC (pose) & SOC (no pose) & EOC-1 & EOC-2\\
  \hline
  EMACH & 0.88 & - & 0.77 & 0.68 \\
   \hline
  SVM & 0.90 & 0.84 & 0.81 & 0.75\\
  \hline
  CondGauss & 0.92 & 0.86 & 0.80 & 0.76\\
  \hline
  AdaBoost & 0.92 & 0.87 & 0.82 & 0.78\\
  \hline
	IGT & 0.95 & 0.89 & 0.85 & 0.80\\
	\hline
\end{tabular}
\end{center}
\label{tab:avg_rates}
\end{table}

\begin{figure}[t]
  \centering
  \subfigure[Target vs. confuser.]{\includegraphics[scale=0.65]{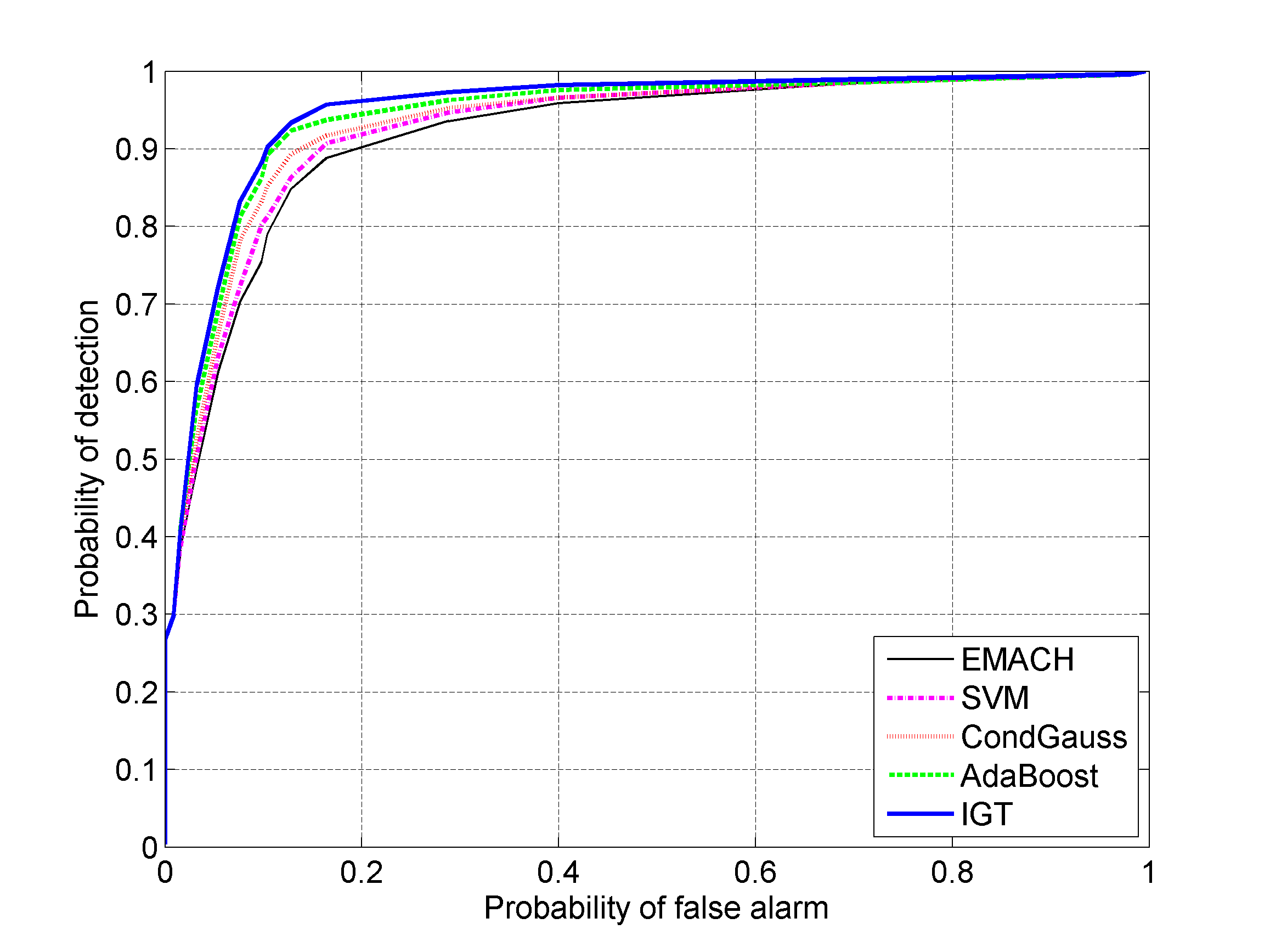}}
  \subfigure[Target vs. clutter.]{\includegraphics[scale=0.65]{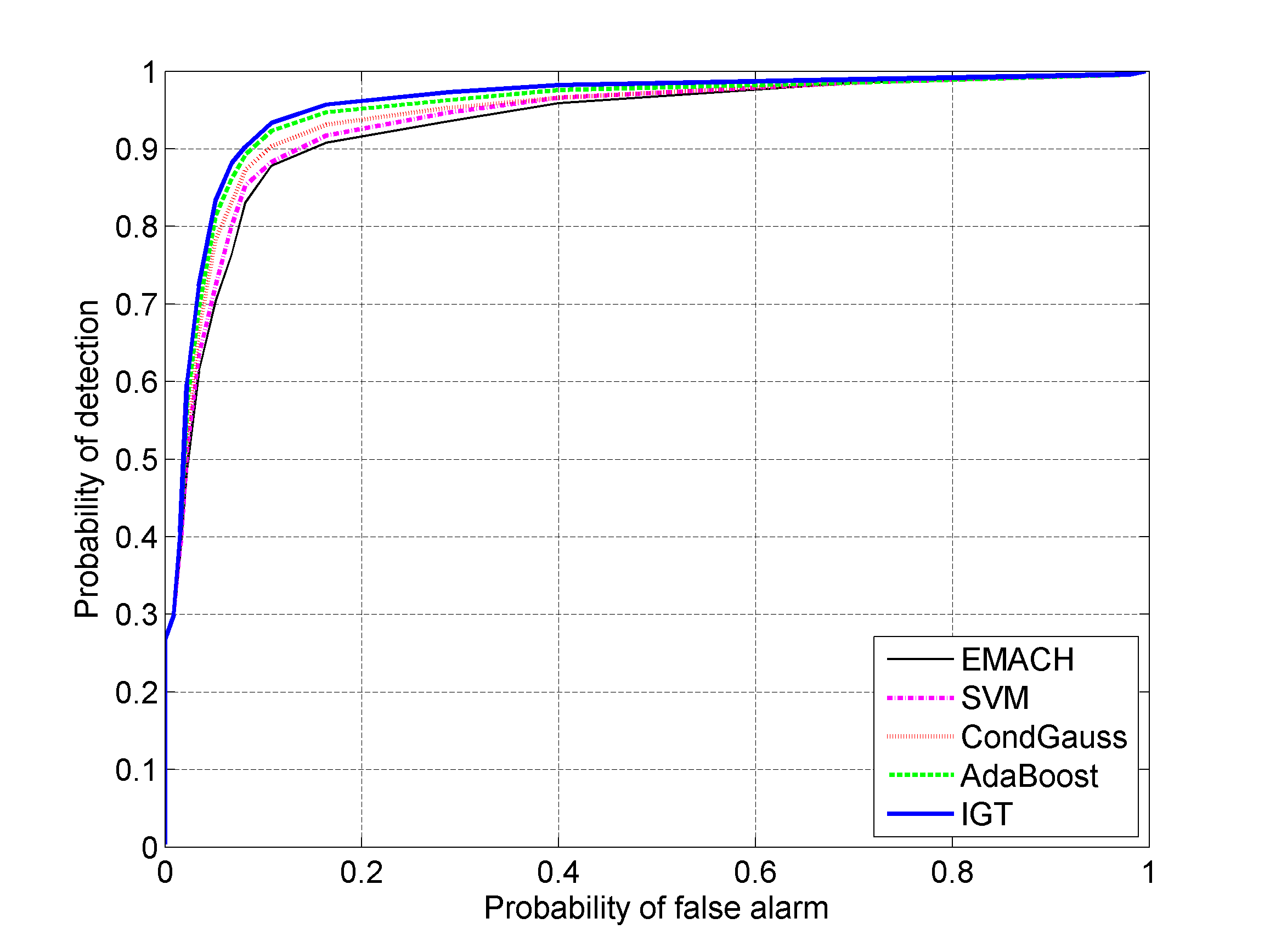}}
  \caption[Receiver operating characteristic curves]{Receiver operating characteristic curves. The proposed IGT method is compared with the EMACH filter\cite{Singh02}, SVM\cite{Zhao01}, conditionally Gaussian model\cite{Sullivan2001}, and the AdaBoost method\cite{boosting:Sun07}.}
  \label{fig:roc_atr}
\end{figure}

\begin{figure}[t]
  \begin{center}
  \subfigure[SOC with pose estimation.]{\includegraphics[scale=0.65]{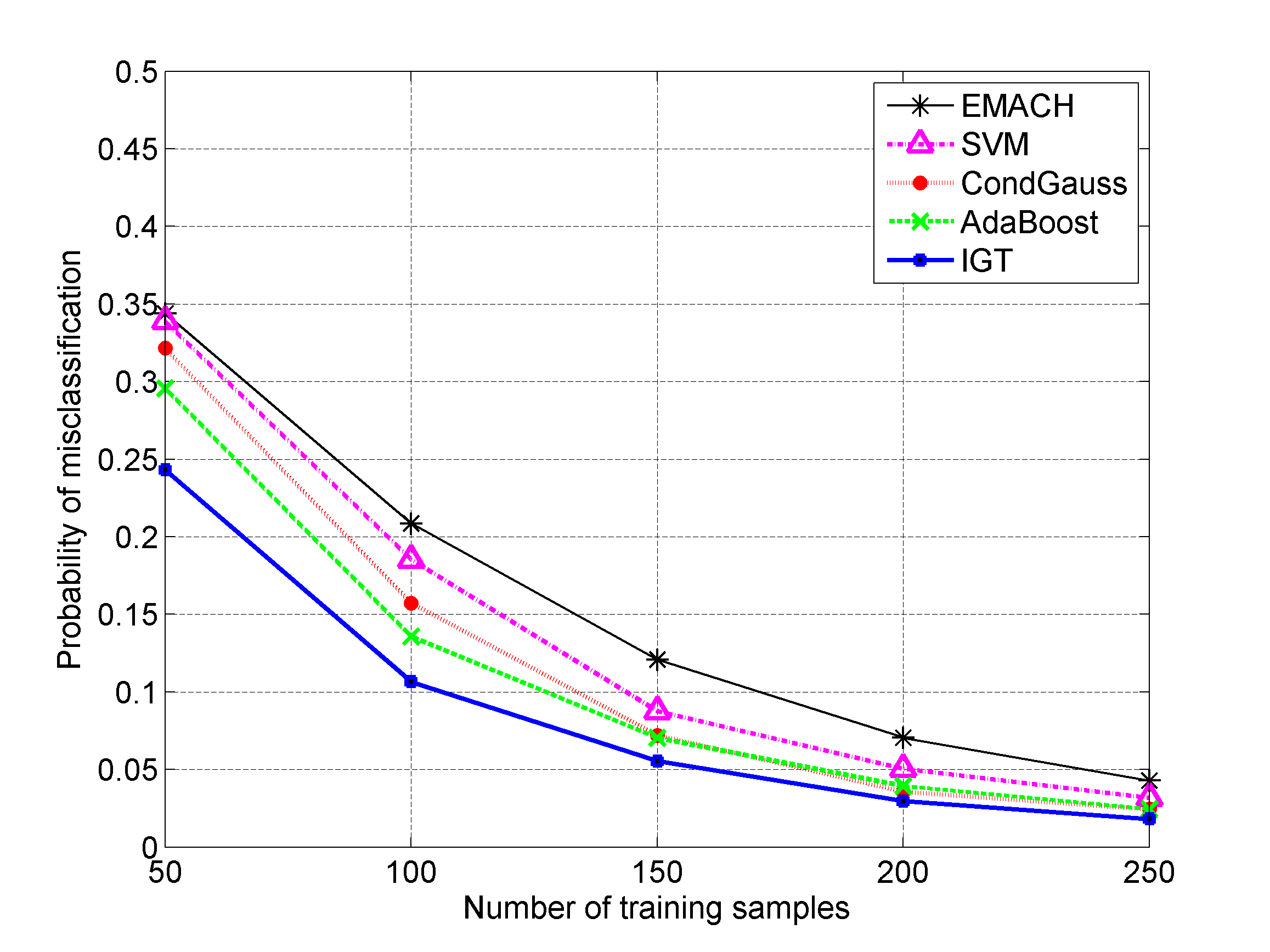}}
  \subfigure[SOC with no pose estimation.]{\includegraphics[scale=0.65]{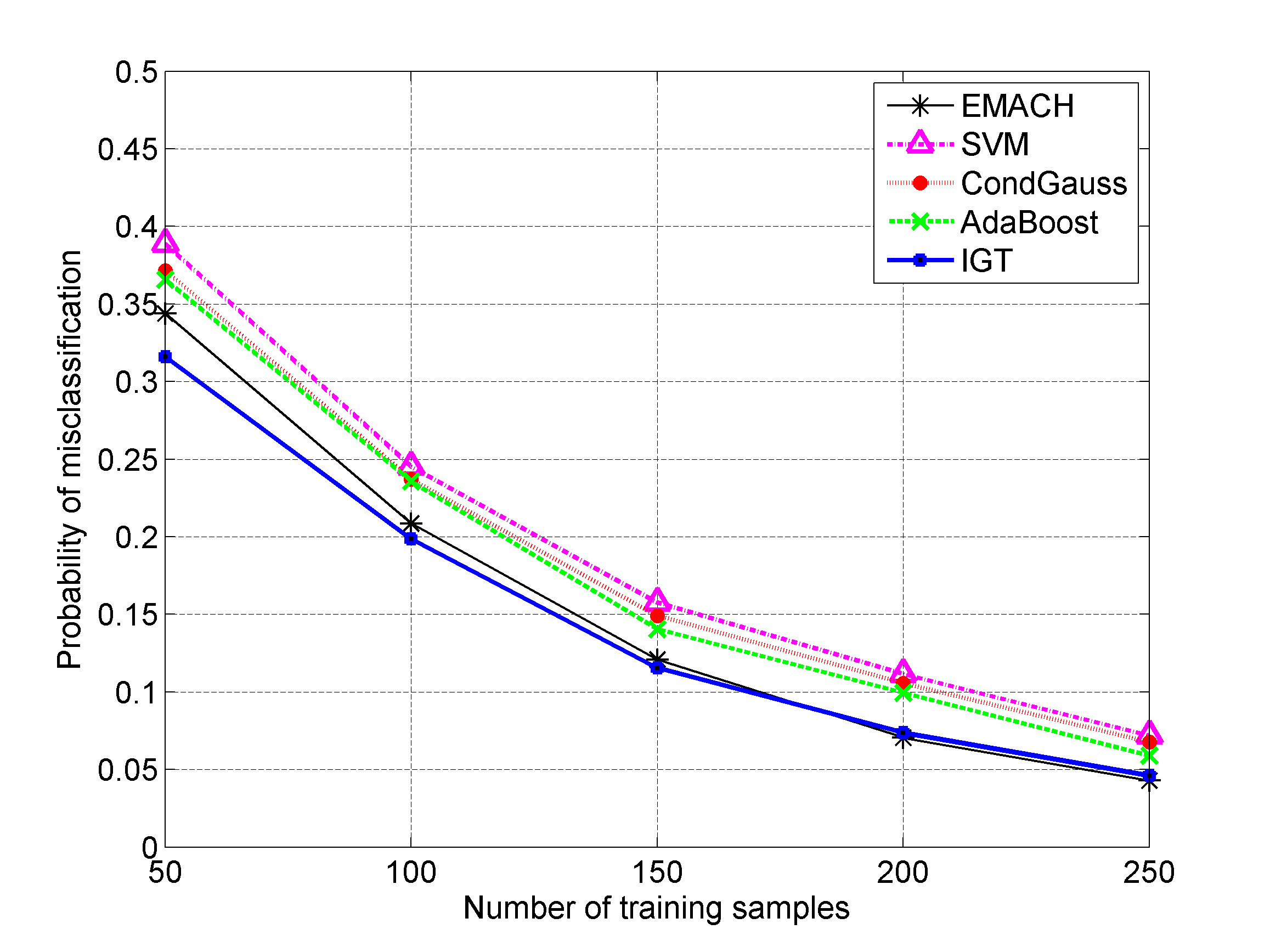}}
  \caption[Classification error vs. training sample size: SOC]{Classification error vs. training sample size. The proposed IGT method is compared with the EMACH filter\cite{Singh02}, SVM\cite{Zhao01}, conditionally Gaussian model\cite{Sullivan2001}, and the AdaBoost method\cite{boosting:Sun07}.}
  \label{fig:trng_size_soc}
  \end{center}
\end{figure}

\begin{figure}[t]
  \begin{center}
  \subfigure[EOC-1.]{\includegraphics[scale=0.65]{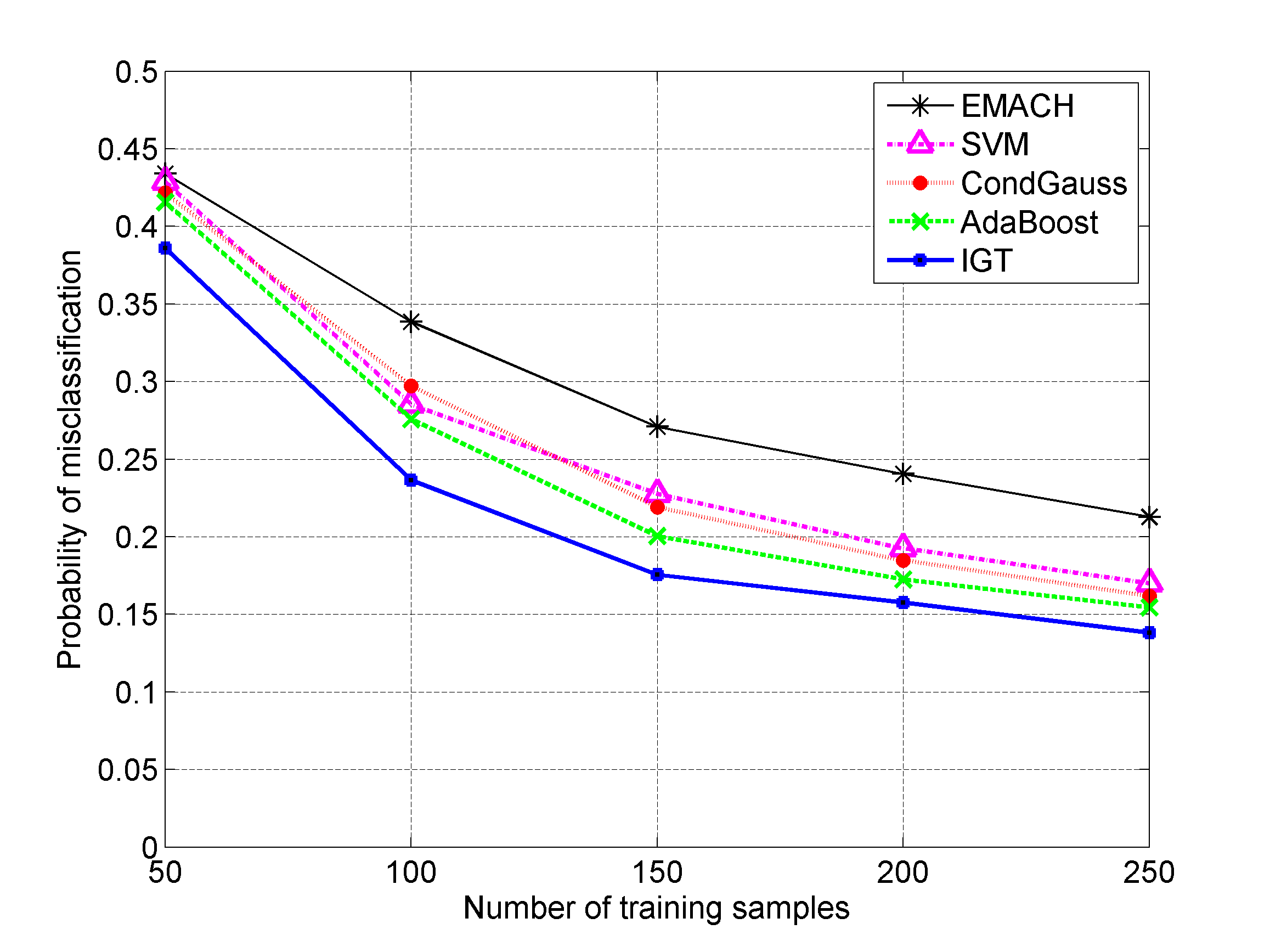}}
  \subfigure[EOC-2.]{\includegraphics[scale=0.65]{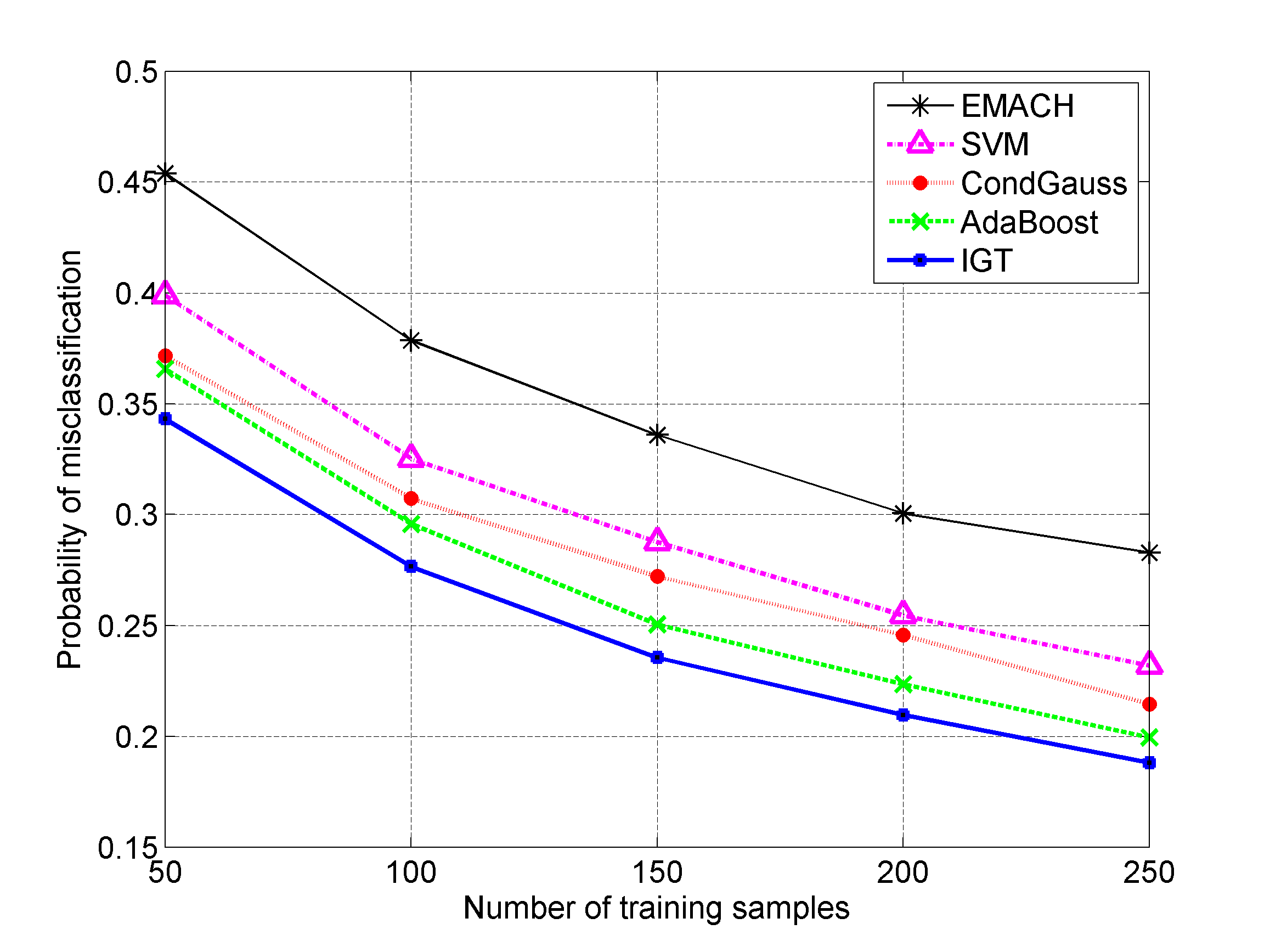}}
  \caption[Classification error vs. training sample size: EOC]{Classification error vs. training sample size. The proposed IGT method is compared with the EMACH filter\cite{Singh02}, SVM\cite{Zhao01}, conditionally Gaussian model\cite{Sullivan2001}, and the AdaBoost method\cite{boosting:Sun07}.}
  \label{fig:trng_size_eoc}
  \end{center}
\end{figure}

\chapter{Application: Learning Graphical Models on Sparse Features}
\label{chapter:sparsity_gm}

\section{Introduction}
The goals of this chapter are two-fold. First, the role of sparsity in signal processing applications is reviewed. Starting with a brief description of the compressive sensing (CS) problem, we present a recent seminal contribution that exploits the underlying analytical framework of CS for classification tasks via class-specific dictionaries. Such sparse representations have been shown to be discriminative and robust to a variety of real-world imaging distortions. We also briefly review an extension of this framework that can handle the scenario of multiple measurements through simultaneous/joint sparsity models. This discussion naturally serves as the background for the contributions in Chapter \ref{chapter:struct_sparsity}, which deal with discriminative structured models on sparse coefficients.

The second goal of this chapter is to offer preliminary validation of the links between graphical models and sparse features \cite{cevher:spm10}. Specifically, we use our graphical model framework from Chapter \ref{chapter:gm} to learn discriminative trees on collections of sparse features that have been jointly extracted from images in a specific manner. We demonstrate this for two different practical applications: (i) hyperspectral target detection and classification, and (ii) robust face recognition. In hyperspectral imaging, we exploit the fact that hyperspectral scenes are spatially homogeneous to enforce identical sparse representation structure on a local neighborhood of pixels. In face recognition, we exploit the correlations among sparse features extracted from different parts of the face that carry discriminative information - the eyes, nose, and mouth. In each application, we see that our graphical approach offers robustness in the scenario of limited training.

\section{Sparsity in Signal Processing}
\label{sec:sparsity_sig_proc}

The value of parsimony in signal representation has been recognized for a long time now. It is well-known that a large class of signals, including audio and images, can be expressed naturally in a compact manner with respect to well-chosen basis representations. Among the most widely applicable of such basis representations are the Fourier and wavelet basis. Sparse signal decomposition and representation have emerged as the cornerstone of high-performing signal processing algorithms. A sparse representation can not only provide better signal compression for bandwidth efficiency, but also lead to faster processing algorithms. Sparse signal representation allows us to capture the simple structure often hidden in visual data, and thus minimizes the undesirable effects of noise in practical settings. This has inspired a proliferation of applications that involve sparse signal representations for acquisition \cite{candes:tit06}, compression \cite{taubman_jpeg:01}, and modeling \cite{lustig_mri:mrm07}.

The pre-eminence of vision among all our sensory systems has led to our enduring fascination with the workings of the human visual system\cite{marr:book82}. The seminal work of Olshausen and Field \cite{olshausen:nature96} established that the receptive fields of simple cells in mammalian primary visual cortex (V1) can be characterized as being selective to a variety of specific stimuli such as color, texture, orientation, and scale. Interpreted differently, the firing of neurons with respect to a particular input scene is typically highly \emph{sparse} if we assume that the neurons form an overcomplete dictionary of base signals \cite{olshausen:nature96,olshausen:vr97,serre:phd06}. 

The Sparseland model \cite{aharon06:ksvd} has emerged as a powerful method to describe signals based on the sparsity and redundancy of their representations. This linear representation model is simplistic albeit quite powerful in its ability to capture redundancy given a good basis for representation. The quest for economical signal representations in practice has fueled the development of very efficient compression algorithms for digital image and video signals\cite{taubman_jpeg:01,watkinson_mpeg:01}. Compressive sensing \cite{candes:tit06} has formalized this notion of sparse representations by seeking the vector with minimum number of non-zero entries that minimizes reconstruction error (w.r.t. a sparsifying basis representation $\mat A$).

\subsection{Compressive Sensing}
Analytically, the challenge in CS is to recover a signal $\vect x \in \mathbb{R}^n$ given a vector of linear measurements $\vect y \in \mathbb{R}^m$ of the form $\vect y = \mat A\vect x$, where $m \ll n$. Assuming $\vect x$ is compressible, it can be recovered from this underdetermined system of equations by solving the following problem \cite{candes:tit06}:
\be
(\mbox{P}_0) \quad \min_{\vect x}\|\vect x\|_0 ~\mbox{subject to}~ \vect y = \mat A\vect x,
\label{eq:l0}
\ee
where $\|\vect x\|_0$ is the $l_0$-``norm'' that counts the number of non-zero entries in $\vect x$. This is an NP-hard problem and it has been shown \cite{donoho:cpam06} that if $\vect x$ is sufficiently sparse, it can be exactly recovered by solving the convex program
\be
(\mbox{P}_1) \quad \min_{\vect x}\|\vect x\|_1 ~\mbox{subject to}~ \vect y = \mat A\vect x.
\label{eq:l1}
\ee
The $l_1$-norm problem has the additional interpretation of enforcing an i.i.d. Laplacian \emph{prior} on $\vect x$. In fact, this is a specific example of the broader Bayesian perspective that prior information can improve signal comprehension. In practice, the presence of noise in real signals is accounted for by relaxing the equality constraint $\vect y = \mat A\vect x$ to the inequality constraint $\|\vect y - \mat A\vect x\|_2 < \epsilon$ for some fixed noise level $\epsilon$,
\be
(\mbox{P}_2) \quad \min_{\vect x}\|\vect x\|_1 ~\mbox{subject to}~ \|\vect y - \mat A\vect x\|_2 < \epsilon.
\label{eq:l1_noise}
\ee
Approaches to solve the problem (P$_2$) are well-known, e.g. lasso in statistics \cite{tibshirani_lasso:jrss96}.

With the goal of achieving better signal \emph{recovery}, research in CS has primarily focused on two aspects: (i) the design of optimal ``sparsifying'' projections $\mat A$ \cite{rubinstein:pieee10}, and (ii) efficient algorithms to solve (P$_0$) \cite{bruckstein:siam09}.

\subsection{Sparse Representation-based Classification}
\label{secch3:src}
%
A seminal contribution to the development of algorithms for signal classification and decision-making is a recent sparse representation-based classification (SRC) framework \cite{wright:tpami09}. In this work, Wright \emph{et al.} explicitly mined the \emph{discriminative capability} of sparse representations for image classification by combining two ideas: (i) the analytical framework underlying CS, and (ii) the development of models for human vision based on overcomplete dictionaries \cite{olshausen:vr97}. Given a sufficiently diverse collection of training images from each class, a linear representation model is assumed, whereby an image from a specific class can be expressed approximately as a linear combination of training images from the same class. So, if a basis matrix or dictionary is built from training images of all classes, any test image has a sparse representation with respect to such a dictionary. Here, sparsity manifests itself due to the class-specific design of dictionaries as well as the assumption of the linear representation model. This model alleviates the challenge of designing sophisticated task-specific features. 

Suppose that there are $K$ different image classes, labeled ${C}_1,\ldots,{C}_K$. Let there be $N_i$ training samples (each in $\R^n$) corresponding to class ${C}_i, i = 1,\ldots,K$. It is understood that each sample is the vectorized version of the corresponding grayscale (or single channel) image. The training samples corresponding to class ${C}_i$ can be collected in a matrix $\mat D_i \in \R^{n\times N_i}$, and the collection of all training samples is expressed using the matrix
\be
\mat D = [\mat D_1 ~ \mat D_2 ~\ldots ~ \mat D_K],
\label{eq:basis_rep}
\ee
where $\mat D \in \R^{n \times T}$, with $T = \sum_{k=1}^{K}N_k$. A new test sample $\vect{y} \in \R^{n}$ can be expressed as a sparse linear combination of the training samples,
\be
\vect y ~\simeq~ \mat D_1 \vect \alpha_1 + \ldots + \mat D_K \vect \alpha_K = \mat D \bm{\alpha},
\label{eq:sparse_rep}
\ee
where $\vect{\alpha}$ is ideally expected to be a sparse vector (i.e., only a few entries in $\vect{\alpha}$ are nonzero). The classifier seeks the sparsest representation by solving the following problem:
\be
(\mbox{P}_3) \quad \hat{\vect{\alpha}} = \arg\min \norm{\vect{\alpha}}_0 \quad\text{subject to}\quad \|\vect{y} - \mat D\vect{\alpha}\|_2 \leq \epsilon,
\label{eq:global_l0}
\ee
where $\norm{\cdot}_0$ denotes the number of nonzero entries in the vector and $\varepsilon$ is a suitably chosen reconstruction error tolerance. Essentially, this is a modification of (P$_2$) with $\mat D := [\mat D_1, ~\mat D_2 ~\ldots, ~\mat D_K]$ and the $l_0$-norm. The problem in \eqref{eq:global_l0} can be solved by greedy pursuit algorithms~\cite{tropp:tit05,dai:tit09}. Once the sparse vector is recovered, the identity of $\vect{y}$ is given by the minimal class-specific reconstruction residual,
\be
\text{Class}(\vect{y}) = \arg\min_i\norm{\vect{y}  - \mat D \delta_i(\hat{\vect{\alpha}})},
\label{eq:global_identity}
\ee
where $\delta_i(\vect{\alpha})$ is a vector whose only nonzero entries are the same as those in $\vect{\alpha}$ which are associated with class ${C}_i$. 

Rooted in optimization theory, the robustness of the sparse feature vector to real-world image distortions like noise and occlusion has led to its widespread application in practical classification tasks. Modifications to \eqref{eq:global_l0} include relaxing the non-convex $l_0$-term to the $l_1$-norm \cite{donoho:cpam06} and introducing regularization terms to capture physically meaningful constraints \cite{yu:isbi11}. This sparsity-based algorithm has been shown \cite{wright:tpami09,wagner:tpami12} to yield markedly improved performance over traditional efforts in the practical problem of face recognition under various distortion scenarios, including illumination, disguise, occlusion, and random pixel corruption. Sparse representation-based classification has also been applied successfully to other discriminative applications such as subspace clustering~\cite{Elhamifar_2009}, iris recognition~\cite{pillai:pami11}, and classification of hyperspectral imagery~\cite{Chen_2011c}.

Modifications to the original SRC problem have considered regularizers that exploit the correlated behavior of groups of coefficients corresponding to the same training sub-dictionary. The simplest extension minimizes the sum of $l_2$-norms of the sub-vectors $\vect x_i$, giving rise to a $l_1-l_2$ group sparse regularizer \cite{yu:isbi11}. This is similar to the idea of the group Lasso \cite{yuan:jrss07}. This method does not explicitly enforce sparsity within each group, an issue which has subsequently been addressed using the hierarchical Lasso in \cite{sprechmann:tsp11}. Other types of group regularizers have been proposed in \cite{majumdar:icassp09}.

In many scenarios, we have access to multiple sets of measurements that capture information about the same image. The SRC model is extended to incorporate this additional information by enforcing a common support set of training images for the $T$ correlated test images $\vect y_1,\ldots, \vect y_T$:
\be
    \label{eqn::joint_sparsity_model}
    \begin{split}
      \mat Y &= \begin{bmatrix} \vect y_1 & \vect y_2 & \cdots & \vect y_T \end{bmatrix}
    = \begin{bmatrix} \mat D\vect \alpha_1 & \mat D\vect \alpha_2 & \cdots & \mat D\vect \alpha_T\end{bmatrix}\\
    &= \mat D \underbrace{\begin{bmatrix} \vect \alpha_1 & \vect \alpha_2 & \cdots & \vect \alpha_T\end{bmatrix}}_{\mat S}
    = \mat D \mat S.
    \end{split}
\ee
The vectors $\vect \alpha_i, i = 1,\ldots,T$, all have non-zero entries at the \emph{same} locations, albeit with different weights, leading to the recovery of a sparse matrix $\mat S$ with only a few nonzero rows,
\begin{equation}
\label{eqn::joint_sparse_recovery}
\hat{\mat S} = \arg\min \norm{\mat Y - \mat D \mat S}_F \quad\text{subject to}\quad \norm{\mat S}_{\text{row},0} \leq K_0,
\end{equation}
where $\norm{\mat S}_{\text{row},0}$ denotes the number of non-zero rows of $\mat S$ and $\norm{\cdot}_F$ is the Frobenius norm. The greedy Simultaneous Orthogonal Matching Pursuit (SOMP) \cite{tropp:sp06a} algorithm and convex relaxations of the row-sparsity norm \cite{tropp:sp06b} have been proposed to solve the non-convex problem \eqref{eqn::joint_sparse_recovery}.


\section{Hyperspectral Imaging}

\subsection{Introduction}
Hyperspectral imaging (HSI) sensors acquire digital images in hundreds of continuous narrow spectral bands spanning the visible to infrared spectrum \cite{Borengasser_2008}. A pixel in hyperspectral images is typically a high-dimensional vector of intensities as a function of wavelength. The high spectral resolution of the HSI pixels facilitates superior discrimination of object types.

An important research problem in HSI \cite{Borengasser_2008} is hyperspectral target detection, which can be viewed as a binary classification problem where hyperspectral pixels are labeled as either target or background based on their spectral characteristics. Many statistical hypothesis testing techniques \cite{manolakis_2002} have been proposed for hyperspectral target detection, including the spectral matched filter (SMF), matched subspace detector (MSD) and adaptive subspace detector (ASD). Advances in machine learning theory have contributed to the popularity of SVMs \cite{vapnik:book95} as a powerful tool to classify hyperspectral data \cite{Melgani_2004}. Variants such as SVM with composite kernels, which incorporates spatial information directly in the kernels\cite{camps:svm-ck_06}, have led to improved performance. HSI classification is the generalization of the binary problem to multiple classes.

Recent work has highlighted the relevance of incorporating contextual information during HSI classification to improve performance \cite{Bovolo_2006,Tarabalka_2009,Li_2011,camps:svm-ck_06}, particularly because HSI pixels in a local neighborhood generally correspond to the same material and have similar spectral characteristics. Many approaches have exploited this aspect, for example by including post-processing of individually-labeled samples \cite{Bovolo_2006, Tarabalka_2009} and Markov random fields in Bayesian approaches~\cite{Li_2011}. The composite kernel approach \cite{camps:svm-ck_06} combines the spectral and spatial information from each HSI pixel via kernel composition.

A significant recent advance exploits sparsity for HSI classification \cite{Chen_2011c} based on the observation that spectral signatures of the same material lie in a subspace of reduced dimensionality compared to the number of spectral bands. An unknown pixel is then expressed as a sparse linear combination of a few training samples from a given dictionary and the underlying sparse representation vector encodes the class information. Further, to exploit spatial correlation, a joint sparsity model is employed in \cite{Chen_2011c}, wherein neighboring pixels are assumed to be represented by linear combinations of a few \emph{common} training samples to enforce smoothness across these pixels.

\subsection{Motivation and Contribution}
The technique in \cite{Chen_2011c} performs classification by using (spectral) reconstruction error computed over the pixel neighborhood. The sparse representations corresponding to different pixels in a local neighborhood are statistically correlated, and this correlation is captured intuitively by the joint sparsity model. A challenging open problem, therefore, is to mine the \emph{class-conditional correlations} among these distinct feature representations in a discriminative manner for detection and classification.
\begin{figure}[t]
  \centering
  \includegraphics[scale=0.33]{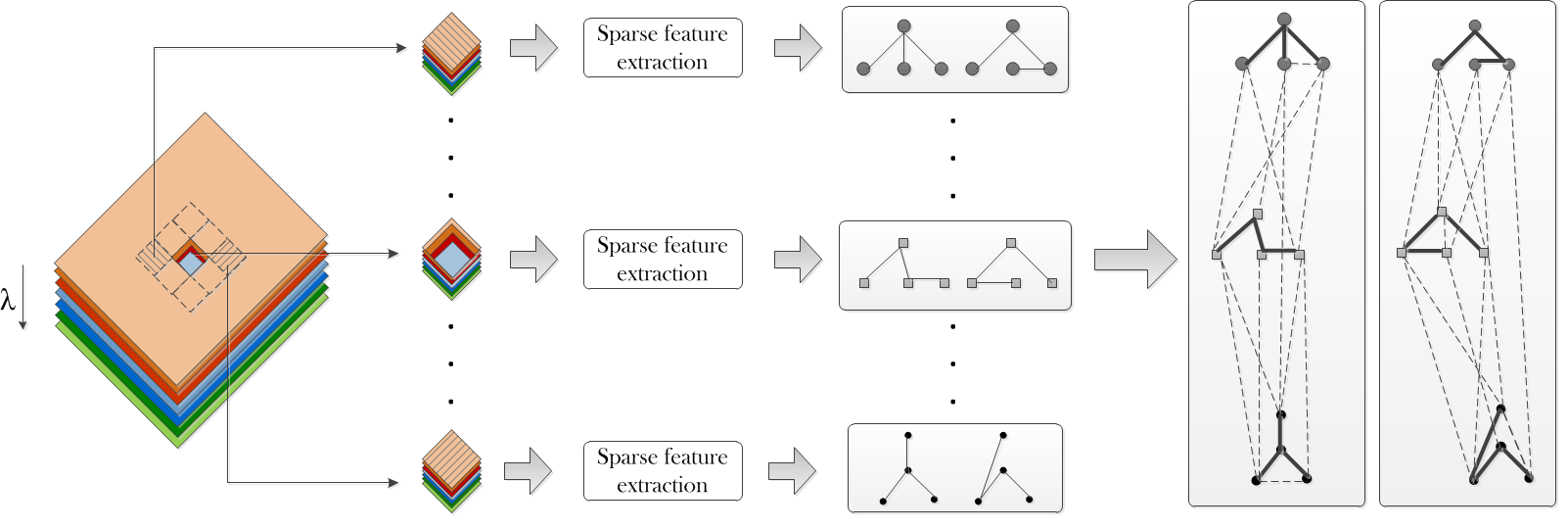}
  \caption[Hyperspectral image classification using discriminative graphical models]{Hyperspectral image classification using discriminative graphical models on sparse feature representations obtained from local pixel neighborhoods.}
  \label{fig:hsi}
\end{figure}

Recent work \cite{cevher:spm10} in model-based compressed sensing has shown the benefits of using probabilistic graphical models as priors on sparse coefficients for signal (e.g.\ image) reconstruction problems. Inspired by this, we use probabilistic graphical models to enforce {\em class-specific} structure on sparse coefficients, wherein our designed graphs represent class conditional densities. Fig. \ref{fig:hsi} shows an illustration of the overall framework. 

First, multiple sparse representations (corresponding to each pixel in a spatial neighborhood) are extracted using the joint sparsity model \cite{Chen_2011c}. We claim that these sparse representations offer complementary yet correlated information that is useful for classification. Our graphical model-based framework introduced in Chapter \ref{chapter:gm} then exploits these class conditional correlations into building a powerful classifier. Specifically, a pair of discriminative tree graphs \cite{tan:tsp10} is first learned for each distinct set of features, i.e. the sparse representation vectors of each pixel in the local spatial neighborhood of a central pixel. These initially disjoint graphs are then thickened (by introducing new edges) into a richer graphical structure via boosting \cite{tan:tsp10,freund:jsai99,Downs04}. The training phase of our graphical model learning uses sparse coefficients from all HSI classes. Consequently we learn a \emph{discriminative} graph-based classifier that captures inter-class information, which is ignored by the \emph{reconstruction} residual-based scheme in \cite{Chen_2011c}.

\section{Experimental Results and Discussion}
\label{sec:intro}

\begin{table}[t]
\caption[Target detection: Confusion matrix for the FR-I hyperspectral image. Four different methods are compared. ($N_t = 18$ and $N_b$ = 216.)]{Target detection: Confusion matrix for the FR-I hyperspectral image. Four different methods are compared. ($N_t = 18$ and $N_b$ = 216.)}
\centering
\begin{tabular}{|c|c|c||c|}
  \hline
  Class & Target & Background & Method\\
  \hline
  Target & 0.6512 & 0.3488 & MSD\\
  & 0.9493 & 0.0507 & SVM-CK\\
  & 0.9556 & 0.0444 & SOMP\\
  & {\bf 0.9612} & {\bf 0.0388} & {\bf LSGM}\\
  \hline
  Background & 0.0239 & 0.9761 & MSD\\
  & 0.0090 & 0.9910 & SVM-CK\\
  & 0.0097 & 0.9903 & SOMP\\
  & {\bf 0.0086} & {\bf 0.9914} & {\bf LSGM}\\
  \hline
\end{tabular}
\label{tab:conf_mat_fr1}
\end{table}

\begin{figure}[t]
  \centering
  \includegraphics[scale=0.7]{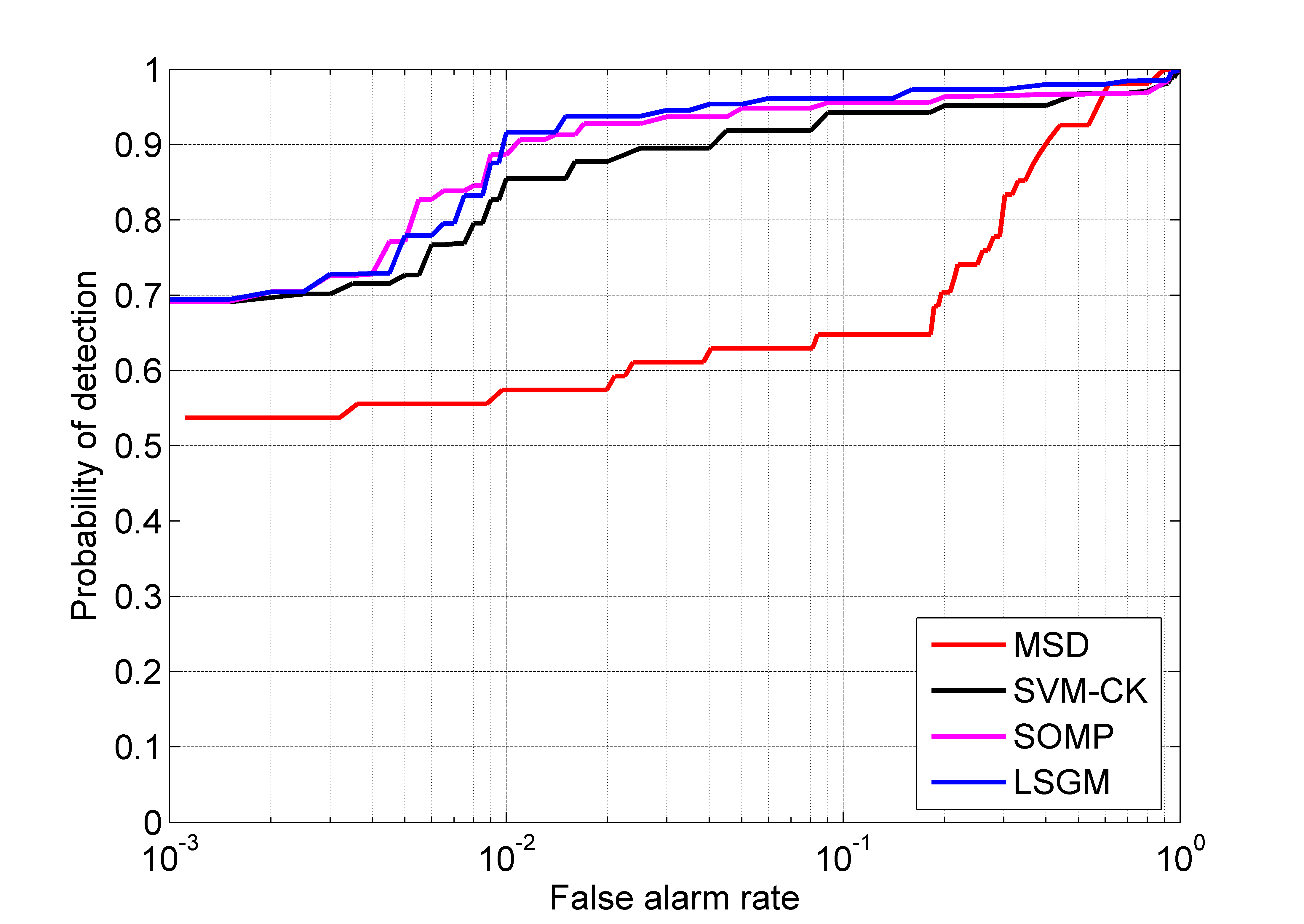}
  \caption[Target detection: ROC for FR-I]{Target detection. ROC for FR-I comparing the four approaches: (i) MSD \cite{Scharf_1994}, (ii) SVM-CK \cite{camps:svm-ck_06}, (iii) SOMP \cite{Chen_2011b}, and (iv) the proposed algorithm (LSGM).}
  \label{fig:roc_hsi}
\end{figure}

In this section, we present separate sets of experimental results for the detection and classification problems. Our proposed algorithm is termed as Local-Sparse-GM (LSGM).

\subsection{Hyperspectral Target Detection}

\begin{table}[t]
\centering
\caption[Classification rates for the AVIRIS Indian Pines test set. LSGM $z$-score = $-2.13$]{Classification rates for the AVIRIS Indian Pines test set. LSGM $z$-score = $-2.13$.}
\begin{tabular}{|l|c|c|c|c|c|c|}
  \hline
  Class type & Training & Test & SVM & SVM-CK & SOMP & LSGM \\
  \hline
  Alfalfa & 6 & 48 &83.33 & 95.83 & 87.50 & 89.58 \\
  Corn-notill & 144 & 1290 & 88.06 & 96.82 & 94.80 & 95.50 \\
  Corn-min & 84 & 750 & 72.40 & 91.20 & 94.53 & 94.80 \\
  Corn & 24 & 210 & 60.48 & 87.62 & 93.33 & 94.76 \\
  Grass/pasture & 50 & 447 & 92.39 & 93.74 & 89.71 & 90.82 \\
  Grass/trees & 75 & 672 & 96.72 & 97.62 & 98.51 & 99.55 \\
  Pasture-mowed & 3 & 23 & 47.82 & 73.91 & 91.30 & 91.30 \\
  Hay-windrowed & 49 & 440 & 98.41 & 98.86 & 99.32 & 99.55 \\
  Oats & 2 & 18 & 50.00 & 55.56 & 0 & 44.44 \\
  Soybeans-notill & 97 & 871 & 72.91 & 94.26 & 89.44 & 90.93 \\
  Soybeans-min & 247 & 2221 & 85.14 & 94.73 & 97.03 & 97.39 \\
  Soybeans-clean & 62 & 552 & 86.23 & 93.84 & 88.94 & 92.39 \\
  Wheat & 22 & 190 & 99.47 & 99.47 & 100 & 100 \\
  Woods & 130 & 1164 & 93.73 & 99.05 & 99.57 & 99.65 \\
  Building-trees & 38 & 342 & 63.45 & 88.01 & 98.83 & 99.71 \\
  Stone-steel & 10 & 85 & 87.05 & 100 & 97.65 & 98.82 \\
  \hline
  Overall & 1043 & 9323 & 85.11 & 95.15 & 95.31 & 96.18 \\
  \hline
\end{tabular}
\label{tab:hsi_aviris_gm}
\end{table}

\begin{figure}
\centering
\subfigure[]{\includegraphics[scale=0.35]{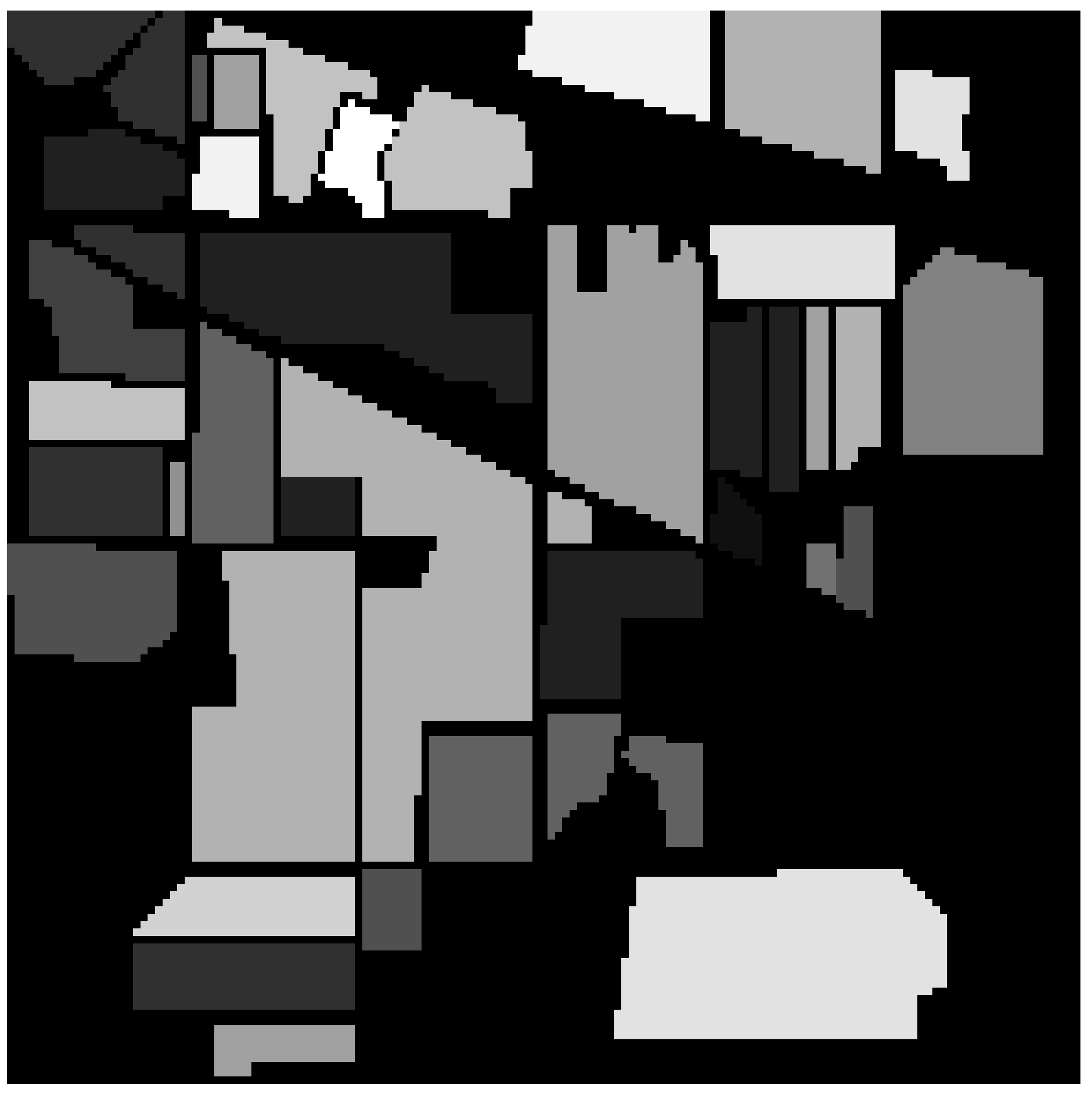}} \hspace{2mm}
\subfigure[]{\includegraphics[scale=0.35]{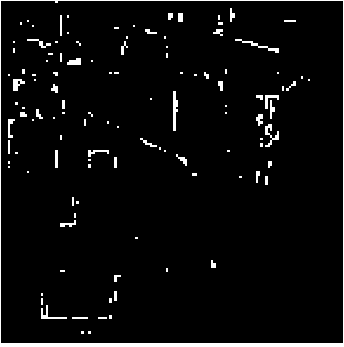}} \hspace{2mm}
\subfigure[]{\includegraphics[scale=0.35]{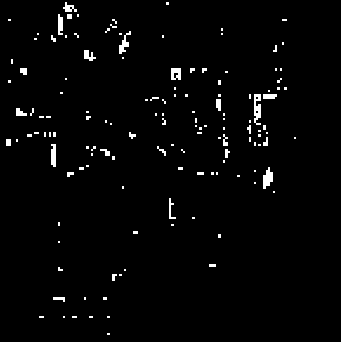}} \hspace{2mm}
\subfigure[]{\includegraphics[scale=0.35]{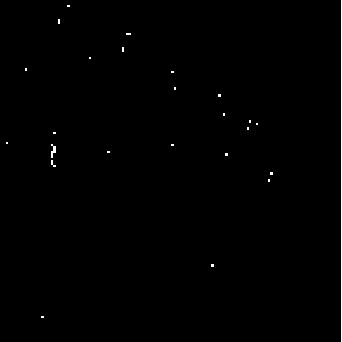}}
\caption[Difference maps for AVIRIS Indian Pine data set]{Difference maps for AVIRIS Indian Pine data set, for the ground truth map in (a). (b) SVM-CK \cite{camps:svm-ck_06}. (c) SOMP \cite{Chen_2011c}. (d) Proposed LSGM approach.}
\label{fig:class_map_gm}
\end{figure}

\begin{table}[t]
\centering
\caption[Classification rates for the University of Pavia test set. LSGM $z$-score = $-2.01$]{Classification rates for the University of Pavia test set. LSGM $z$-score = $-2.01$.}
\begin{tabular}{|l|c|c|c|c|c|c|}
  \hline
  Class type & Training & Test & SVM & SVM-CK & SOMP & LSGM \\
  \hline
  Asphalt & 548 & 6304 & 84.01 & 80.20 & 59.49 & 66.55\\
  Meadows & 540 & 18146 & 67.50 & 84.99 & 78.31 & 86.10\\
  Gravel & 392 & 1815 & 67.49 & 82.37 & 84.13 & 86.72\\
  Trees & 524 & 2912 & 97.32 & 96.33 & 96.30 & 96.94\\
  Metal sheets & 265 & 1113 & 99.28 & 99.82 & 87.78 & 98.83 \\
  Bare soil & 532 & 4572 & 92.65 & 93.35 & 77.45 & 94.62\\
  Bitumen & 375 & 981 & 89.70 & 90.21 & 98.67 & 99.18\\
  Bricks & 514 & 3364 & 92.24 & 92.95 & 89.00 & 94.44\\
  Shadows & 231 & 795 & 96.73 & 95.85 & 91.70 &  96.10\\
  \hline
  Overall & 3921 & 40002 & 79.24 & 87.33 & 78.75 & 86.38 \\
  \hline
\end{tabular}
\label{tab:hsi_univ_pavia}
\end{table}

\begin{table}[t]
\centering
\caption[Classification rates for the Center of Pavia test set. LSGM $z$-score = $-2.17$]{Classification rates for the Center of Pavia test set. LSGM $z$-score = $-2.17$.}
\begin{tabular}{|l|c|c|c|c|c|c|}
  \hline
  Class type & Training & Test & SVM & SVM-CK & SOMP & LSGM \\
  \hline
  Water & 745 & 64533 & 99.19 & 97.61 & 99.38 & 99.44 \\
  Trees & 785 & 5722 & 77.74 & 92.99 & 91.98 & 92.99 \\
  Meadow & 797 & 2094 & 86.72 & 97.37 & 95.89 & 96.99 \\
  Brick & 485 & 1667 & 40.37 & 79.60 & 86.44 & 87.28 \\
  Soil & 820 & 5729 & 97.52 & 98.65 & 96.75 & 97.64 \\
  Asphalt & 678 & 6847 & 94.77 & 94.37 & 93.79 & 94.54 \\
  Bitumen & 808 & 6479 & 74.37 & 97.53 & 95.06 & 96.99\\
  Tile & 223 & 2899 & 98.94 & 99.86 & 99.83 & 99.90 \\
  Shadow & 195 & 1970 & 100 & 99.89 & 98.48 & 99.34 \\
  \hline
   Overall & 5536 & 97940 & 94.63 & 96.97 & 97.82 & 98.20 \\
  \hline
\end{tabular}
\label{tab:hsi_center_pavia}
\end{table}

Hyperspectral images from the HYDICE forest radiance I data collection (FR-I) \cite{hydice_1995} are used for the experiment. The HYDICE sensor generates 210 bands across the whole spectral range from 0.4 to 2.5 $\mu$m, spanning the visible and short-wave infrared bands and including 14 targets. Only 150 of the 210 available bands are retained by removing the absorption and low-SNR bands. The target sub-dictionary $\mat D_t$ comprises 18 training spectra chosen from the leftmost target in the scene, while the background sub-dictionary $\mat D_b$ has 216 training spectra chosen using the dual window technique described in \cite{Chen_2011a}.

Four different methods are compared:
\begin{enumerate}
  \item Classical matched subspace detector (MSD) which operates on each pixel independently \cite{Scharf_1994}
  \item Composite kernel support vector machines (SVM-CK) which considers a weighted sum of spectral and spatial information \cite{camps:svm-ck_06}
  \item Simultaneous orthogonal matching pursuit (SOMP) which involves solving Eq. \eqref{eqn::joint_sparse_recovery} with a $3\times 3$ local window \cite{Chen_2011c}
  \item Proposed local-sparsity-graphical-model (LSGM) approach with the same $3\times 3$ window to generate the sparse features.
\end{enumerate}
Table \ref{tab:conf_mat_fr1} shows the confusion matrix in which detection and error rates are provided with each row representing the true class of the test pixels and each column representing the output of the specified classifier. All four approaches are compared, and the proposed LSGM methods offers better target detection performance. Improvements over SOMP can be attributed to the use of an explicit discriminative classifier in LSGM. All approaches identify the background class with a reasonably high degree of accuracy.

Fig. \ref{fig:roc_hsi} shows the ROC curve for the detection problem. The ROC curve describes the probability of detection (PD) as a function of the probability of false alarms (PFA). To calculate the ROC curve, a large number of thresholds are chosen between the minimum and maximum of the detector output, and class labels for all test pixels are determined at each threshold. The PFA is calculated as the ratio of the number of false alarms (background pixels determined as target) to the total number of pixels in the test region, while the PD is the ratio of the number of hits (target pixels correctly determined as target) to the total number of true target pixels. It can be seen that the proposed LSGM approach offers the best overall detection performance.

\subsection{Hyperspectral Target Classification}
\label{section::simulation}

\begin{figure}[]
  \centering
  \subfigure[]{\includegraphics[scale=0.4]{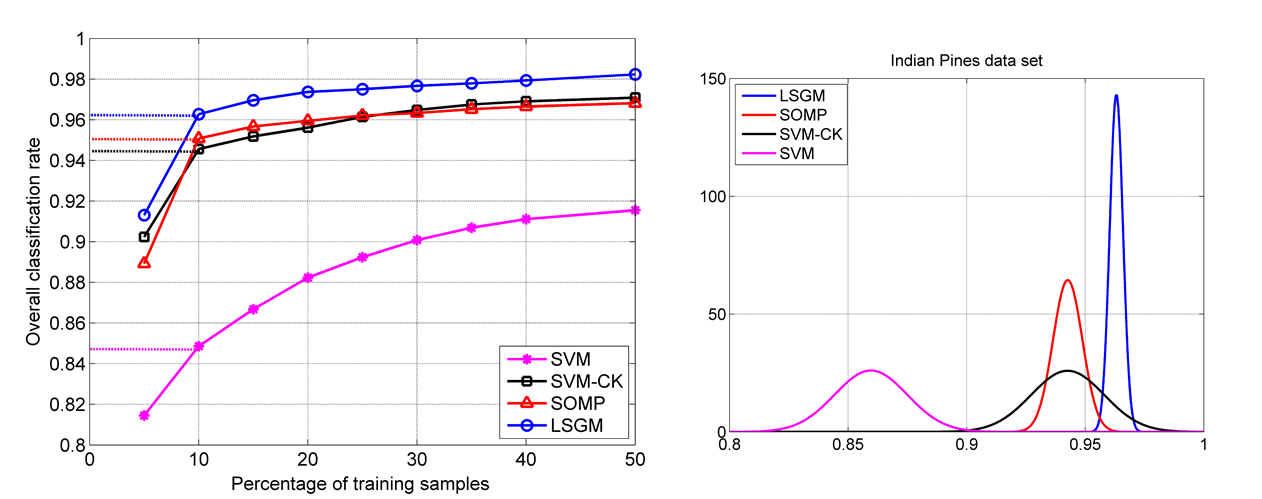}}
	\subfigure[]{\includegraphics[scale=0.4]{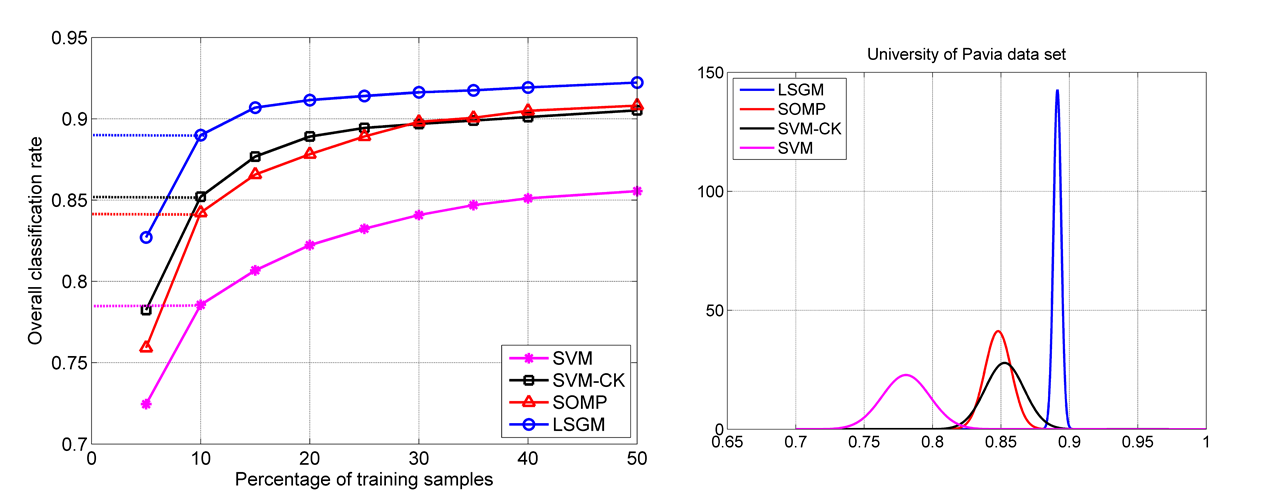}}
	\subfigure[]{\includegraphics[scale=0.4]{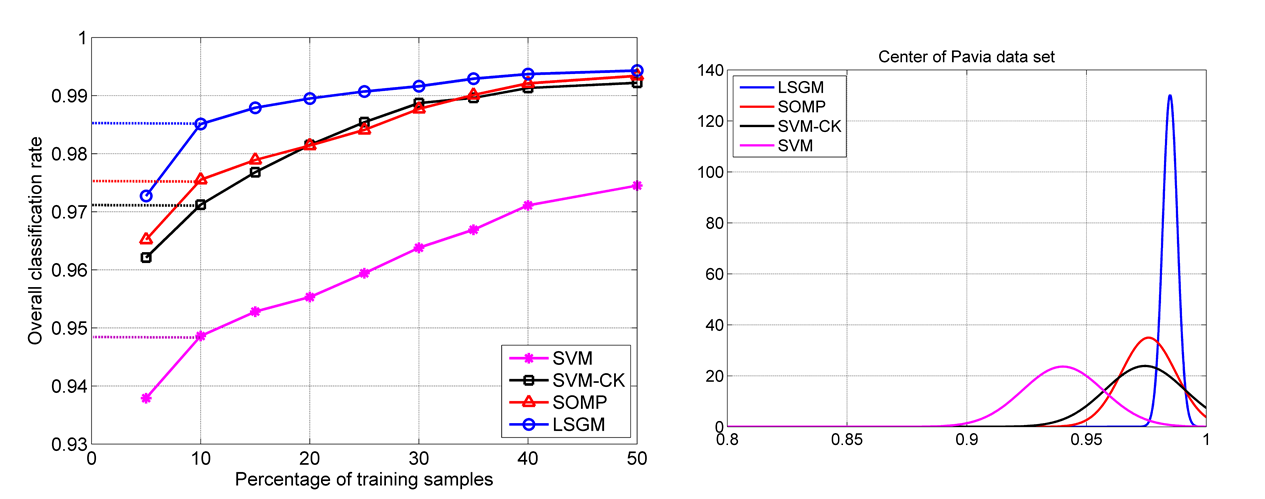}}
  \caption[Performance of different approaches as a function of number of training samples]{Performance of different approaches as a function of number of training samples provided. (a) AVIRIS image, (b) University of Pavia image, (c) Center of Pavia image. For each image, the density function of the classification rates obtained for ten different random realizations of training is plotted on the right.}
  \label{fig:trng_size_hsi}
\end{figure}

Hyperspectral target classification is a generalization of the binary detection problem to multiple classes. Here, we compare our proposed LSGM approach with three competitive methods: (i) spectral features-based SVM classifier \cite{Melgani_2004,Gualtieri_1998}, (ii) composite kernel support vector machines (SVM-CK) \cite{camps:svm-ck_06}, and (iii) joint sparsity model (SOMP) \cite{Chen_2011c}. In SVM-CK, two types of kernels are used: a spectral kernel $K_{\omega}$ for the spectral (pixel) features (in $\mathbb{R}^{200}$) and a spatial kernel $K_s$ for spatial features (in $\mathbb{R}^{400}$) which are formed by the mean and standard deviation of pixels in a neighborhood per spectral channel. A polynomial kernel (order $d$ = 7) is used for spectral features, while the RBF kernel is used for the spatial features. The $\sigma$ parameter for the RBF kernel and SVM regularization parameter $C$ are selected by cross-validation. The weighted summation kernel, $K = \mu K_{s} + (1-\mu)K_{\omega}$, effectively captures spectral and contextual spatial information, with the optimal choice $\mu = 0.4$ determined by cross-validation. A $5\times 5$ window is used for the neighborhood kernels. Parameters for SOMP are chosen as described in \cite{Chen_2011c}. The proposed LSGM approach uses a local window of dimension $3 \times 3$. For fairness of comparison, results for SOMP are also presented for the same window dimension.

We perform experiments using 3 distinct HSI data sets. Note that two flavors of the results are reported:
\begin{enumerate}
  \item Tables \ref{tab:hsi_aviris_gm}-\ref{tab:hsi_center_pavia} show classification rates for carefully selected or good training samples which amount to about 10$\%$ of available data (typical of training choices in \cite{camps:svm-ck_06,Chen_2011c})
  \item Figs.\ \ref{fig:trng_size_hsi}(a)-(c) where performance is plotted as a function of training set size and results averaged from multiple (10) random training runs. In each sub-figure, the plot on the right-hand side characterizes the distribution of the classification rates (modeled as a random variable whose value emerges as an outcome of a given run, and fit to a Gaussian). Further, we establish the statistical significance of our results by computing LSGM $z$-scores for each data set.
\end{enumerate}

\subsubsection{AVIRIS Data Set: Indian Pines}
\label{sec:aviris_expt}
The first hyperspectral image in our experiments is the Airborne Visible/Infrared Imaging Spectrometer (AVIRIS) Indian Pines image \cite{aviris}. The AVIRIS sensor generates 220 bands across the spectral range from 0.2 to 2.4 $\mu$m, of which only 200 bands are considered by removing 20 water absorption bands \cite{Gualtieri_1998}. This image has spatial resolution of 20m per pixel and spatial dimension $145 \times 145$. For well-chosen training samples, difference maps obtained using the different approaches are shown in Figs. \ref{fig:class_map_gm}(b)-(d), and classification rates for each class as well as overall accuracy are shown in Table \ref{tab:hsi_aviris_gm}. The improvement over SOMP indicates the benefits of using a discriminative classifier instead of reconstruction residuals for class assignment, while still retaining the advantages of exploiting spatio-spectral information.

Fig. \ref{fig:trng_size_hsi}(a) compares algorithm performance as a function of training set size. Our LSGM approach outperforms the competing approaches, and the difference is particularly significant in the low training regime. As expected, overall classification accuracy decreases when number of training samples is reduced. That said, LSGM offers a more graceful degradation in comparison to other approaches. From the density function plot, we see that average classification rate is the highest for LSGM, consistent with the plot on the left-hand side in Fig. \ref{fig:trng_size_hsi}(a). Further, variance is the lowest for LSGM, underlining its improved robustness against particular choice of training samples.


\subsubsection{ROSIS Urban Data Over Pavia, Italy}
The next two hyperspectral images, University of Pavia and Center of Pavia, are urban images acquired by the Reflective Optics System Imaging Spectrometer (ROSIS). The ROSIS sensor generates 115 spectral bands ranging from 0.43 to 0.86 ìm and has a spatial resolution of 1.3 m per pixel. The University of Pavia image consists of $610 \times 340$ pixels, each having 103 bands with the 12 noisiest bands removed. The Center of Pavia image consists of $1096 \times 492$ pixels, each having 102 spectral bands after 13 noisy bands are removed. For these two images, we repeat the experimental scenarios tested in Section \ref{sec:aviris_expt}.

Classification rates for the two ROSIS images are provided in Tables \ref{tab:hsi_univ_pavia} and \ref{tab:hsi_center_pavia} respectively, for the scenario of well-chosen training samples. In Table \ref{tab:hsi_univ_pavia}, the SVM-CK technique performs marginally better than LSGM in the sense of overall classification accuracy. However, for most individual classes LSGM does better and particularly in cases where training sample size is smaller. In Table \ref{tab:hsi_center_pavia}, LSGM performs better than SOMP as well as SVM-CK. From Figs. \ref{fig:trng_size_hsi}(b)-(c), we observe that the LSGM improves upon the performance of SOMP and SVM-CK by about 4$\%$, while the improvements over the baseline SVM classifier are even more pronounced.

The $z$-score for LSGM on the AVIRIS image is $-2.13$, which indicates that with a high probability $(= 0.983)$, any random selection of training samples will give results similar to the values in Table \ref{tab:hsi_aviris_gm}. For the University of Pavia and Center of Pavia images, $z$-scores are $-2.01$ and $-2.17$ respectively. The negative sign merely indicates that the experimental value is lesser than the the most likely value (Gaussian mean).

\section{Robust Face Recognition}

\subsection{Introduction}
The problem of automatic face recognition has witnessed considerable research activity in the image processing and computer vision community over the past two decades. Its significance can be gauged by the variety of practical applications that employ face recognition, like video surveillance systems, biometrics to control access to secure facilities, and online face search in social networking applications. The diversity of facial image captures, due to varying illumination conditions, pose, facial expressions, occlusion and disguise, offers a major challenge to the success of any automatic human face recognition system. A comprehensive survey of face recognition methods in literature is provided in \cite{Zhao:2003}.

One of the most popular dimensionality-reduction techniques used in computer vision is principal component analysis (PCA). In face recognition, PCA-based approaches have led to the use of eigenpictures\cite{Sirovich_1987} and eigenfaces\cite{Turk_1991} as features. Other approaches have used local facial features \cite{Zou07} like the eyes, nose and mouth, or incorporated geometrical constraints on features through structural matching. An important observation is that different (photographic) versions of the same face approximately lie in a linear subspace of the original image space \cite{Shashua_1992,Belhumeur_1997,Liu01,Basri_2003}. A variety of classifiers have been proposed for face recognition, ranging from template correlation to nearest neighbor and nearest subspace classifiers, neural networks and support vector machines \cite{Zhao_2003}.

\subsection{Motivation}
\label{sec:local_dic}
Recently, the merits of exploiting parsimony in signal representation and classification have been demonstrated in \cite{wright:tpami09,pillai:pami11,Hang_2009}. The sparsity-based face recognition algorithm \cite{wright:tpami09} yields markedly improved recognition performance over traditional efforts in face recognition under various conditions, including illumination, disguise, occlusion, and random pixel corruption. In many real world scenarios, test images for identification obtained by face detection algorithms are not perfectly registered with the training samples in the databases. The sparse subspace assumption in~\cite{wright:tpami09}, however, requires the test face image to be well aligned to the training data prior to classification. Recent approaches have attempted to address this misalignment issue in sparsity-based face recognition\cite{Huang_2008,Wagner_2009,wagner:tpami12}, usually by jointly optimizing the registration parameters and sparse coefficients and thus leading to more complex systems.

\begin{figure}[t]
  \centering
  \includegraphics[width=.8\columnwidth]{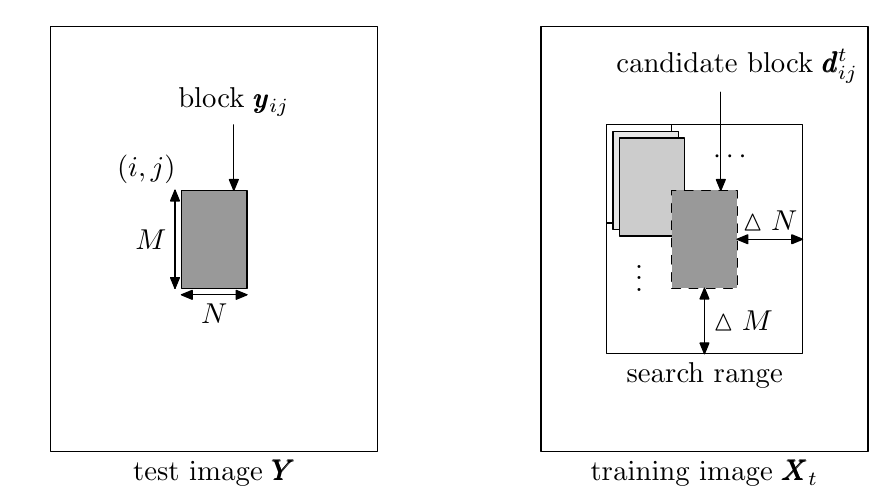}\\(a)\\
  \vspace{3pt}
  \includegraphics[width=.8\columnwidth]{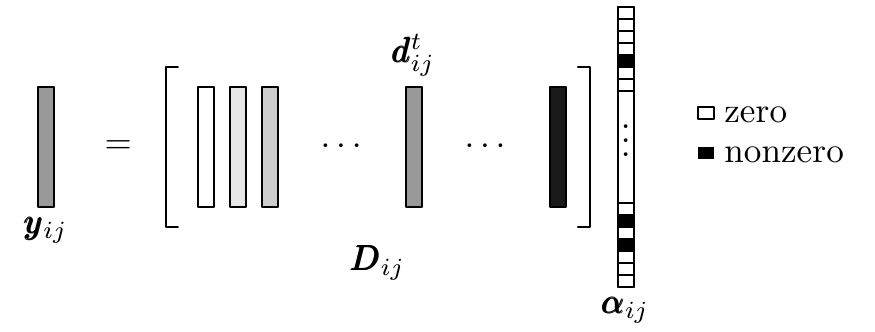}\\(b)
  \caption[Representation of a block in the test image from a locally adaptive dictionary]{Representation of a block in the test image from a locally adaptive dictionary. (a)~The blocks in the test and training images (only one training sample is displayed). (b)~Sparse representation $\vect y_{ij} = \mat D_{ij} \vect\alpha_{ij}$.}
  \label{fig::model}
\end{figure}

In order to overcome this constraint for application to practical face recognition systems, Chen \emph{et al.} \cite{chen:icip10} proposed a locally adaptive sparse representation-based approach, inspired by the inter-frame sparsity model and the observation that local image features are often more beneficial than global features in image processing applications. This is illustrated in Fig. \ref{fig::model}. Accordingly, a (vectorized) local block $\vect y_{ij}$ (indexed by its top-left pixel location $(i,j)$) in a new test image is represented by a sparse linear combination of similar blocks from the training images located within the same spatial neighborhood,
\be
\label{eqn::local_sparsity}
\vect y_{ij} = \mat D_{ij} \vect \alpha_{ij},
\ee
where $\vect \alpha_{ij}$ is a sparse vector and $\mat D_{ij}$ is an adaptive dictionary. The sparse vector is recovered by solving the following optimization problem:
\be
\label{eqn::local_l0}
\hat{\vect \alpha}_{ij} = \arg\min \norm{\vect \alpha_{ij}}_0 ~\text{subject to}~ \|\vect y_{ij} - \mat D_{ij} \vect \alpha_{ij}\|_2 < \epsilon,
\ee
and the class label is determined using the reconstruction residual, similar to the global sparsity approach. To enhance robustness to distortions, multiple local blocks are chosen and the sparse recovery problem is solved for each block individually. Each local block is assigned the label of the class with minimum residual error, and class assignments from all such local blocks are combined either by majority voting or heuristic maximum likelihood-type approaches.

Our contribution is the development of a discriminative graphical model classifier to combine the statistically \emph{correlated} local sparse features from informative local regions of the face - eyes, nose and mouth.

\section{Face Recognition Via Local Decisions From Locally Adaptive Sparse Features}
\label{section::method}

\begin{figure}[t]
  \begin{center}
  \includegraphics[scale=0.4]{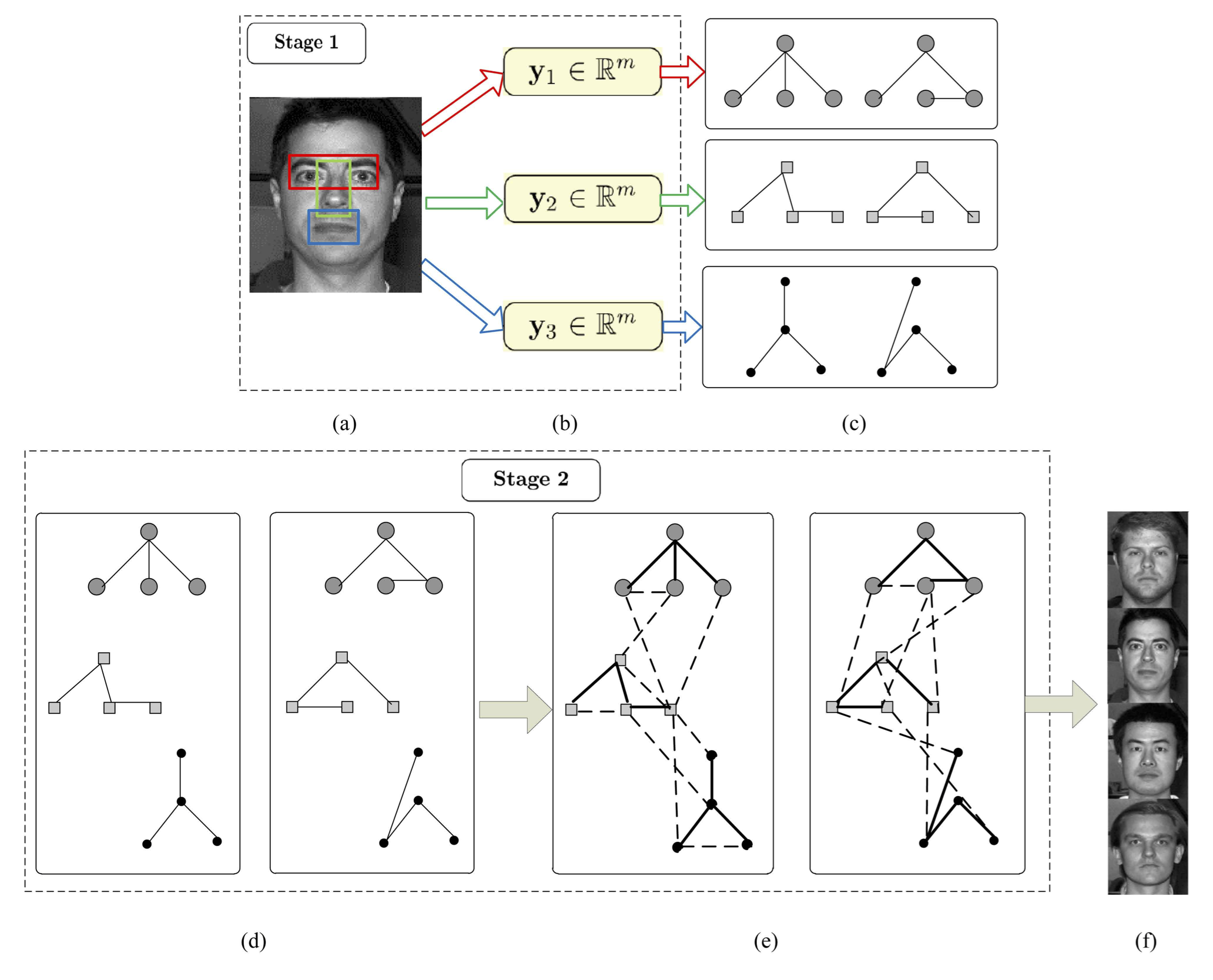}
  \caption[Proposed framework for face recognition]{Proposed framework for face recognition: (a) Target face image, (b) Local regions for extracting sparse features, (c) Initial pairs of \emph{tree} graphs for each feature set, (d) Initial sparse graph formed by tree concatenation, (e) Final pair of thickened graphs; newly learned edges represented by dashed lines, (f) Graph-based inference. In (c)-(e), the graphs on the left and right correspond to distributions $p$ (class $C_i$) and $q$ (class $\tilde{C}_i$) respectively.}
  \label{fig:igt}
  \end{center}
\end{figure}

Fig. \ref{fig:igt} shows the overall discriminative graphical model framework for face recognition. The feature extraction process in Stage 1 is designed differently for this application. We build local dictionaries using the idea described in \ref{sec:local_dic} and separately extract sparse representations corresponding to the eyes, nose and mouth. Since these features are conditionally correlated (they correspond to the same face), they form a suitable set of multiple feature representations which can be fused using our discriminative graphical model classifier. The procedure in Stage 2 is identical to the procedure described previously in Chapter \ref{chapter:gm} and in the application to hyperspectral imaging earlier in this chapter. To handle multiple classes, the graphs are learned in a one-versus-all manner and the test image is identified with the class that maximizes the likelihood ratio.

\section{Experiments and Discussion}
\label{section::simulation}

\begin{figure}[t]
\centering
\begin{tabular}{cc}
\includegraphics[scale=0.7]{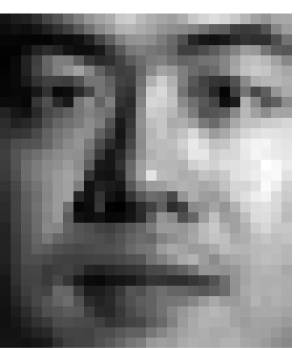} & \includegraphics[scale=0.7]{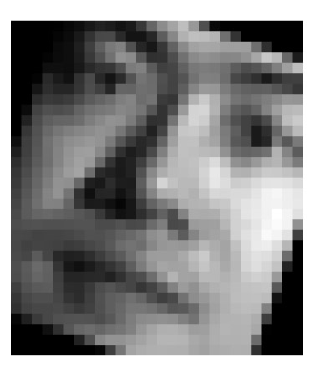}\\
(a) & (b)
\end{tabular}
\caption[An example of rotated test images]{An example of rotated test images. (a)~Original image and (b)~the image rotated by 20 degrees clockwise.}
\label{fig::rotation}
\end{figure}

\begin{figure}[t]
\centering
\includegraphics[scale=0.6]{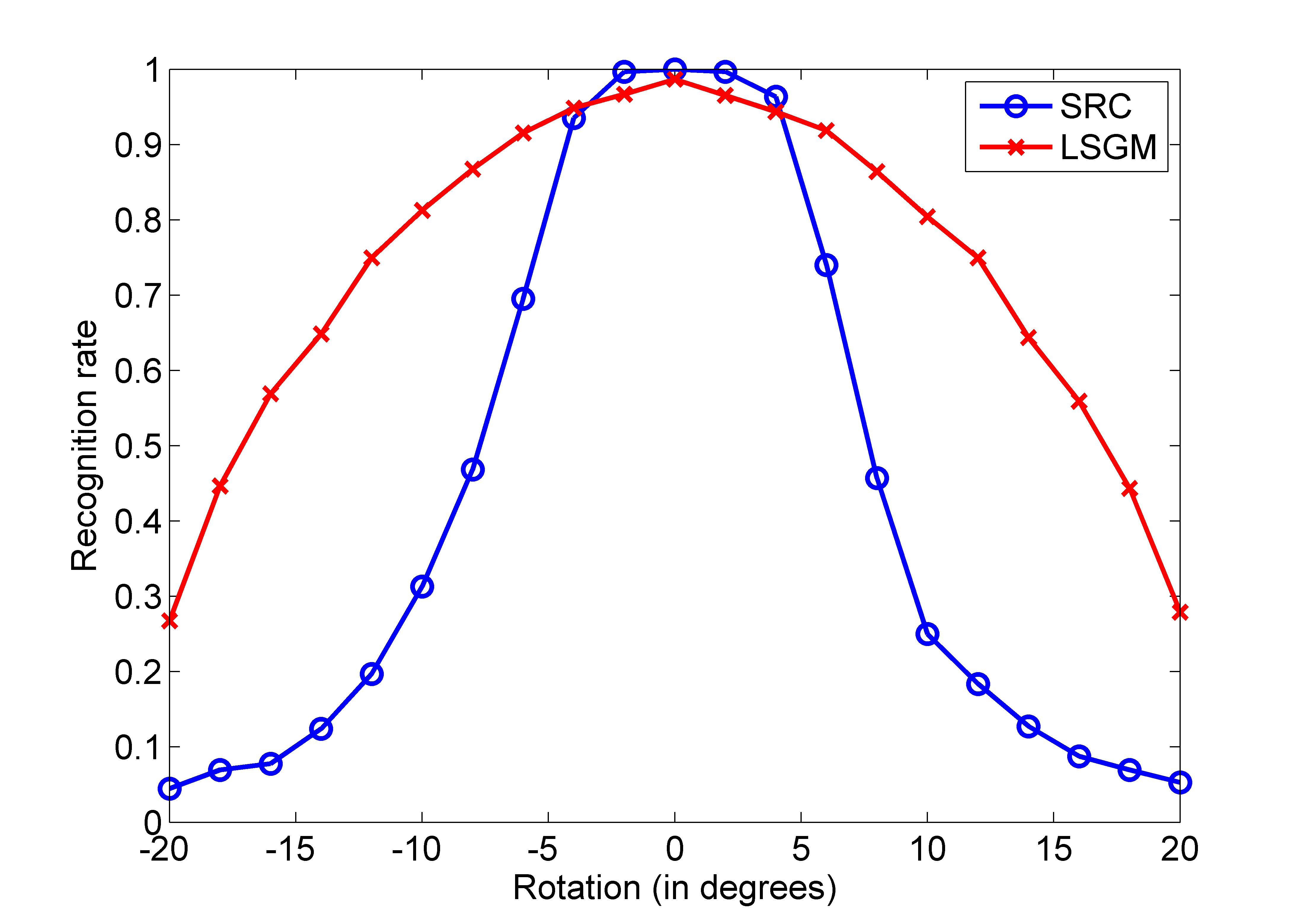}
\caption[Recognition rate for rotated test images]{Recognition rate for rotated test images.}
\label{fig::rotation_results}
\end{figure}
We test the proposed algorithm on the Extended Yale B database \cite{ExtendedYale01}, which consists of 2414 perfectly-aligned frontal face images of size $192\times 168$ of 38 individuals, 64 images per individual, under various conditions of illumination. In our experiments, for each subject we randomly choose 32 images in Subsets 1 and 2, which were taken under less extreme lighting conditions, as the training data. The remaining images are used as test data, after introducing some misalignment. All training and test samples are downsampled to size $32\times 28$.

We compare our LSGM technique against five popular face recognition algorithms: (i) sparse representation-based classification (SRC) \cite{wright:tpami09}, (ii) Eigenfaces \cite{Turk_1991} as features with nearest subspace \cite{Ho03} classifier (Eigen-NS), (iii) Eigenfaces with support vector machine \cite{vapnik:book95} classifier (Eigen-SVM), (iv) Fisherfaces \cite{Belhumeur_1997} as features with nearest subspace classifier (Fisher-NS), and (v) Fisherfaces with SVM classifier (Fisher-SVM).

\begin{figure}[t]
\centering
\begin{tabular}{cc}
\includegraphics[scale=0.7]{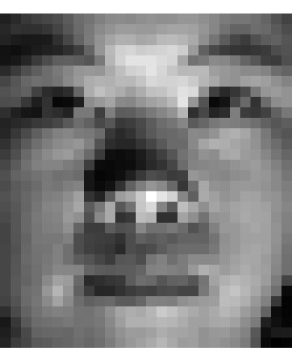} & \includegraphics[scale=0.7]{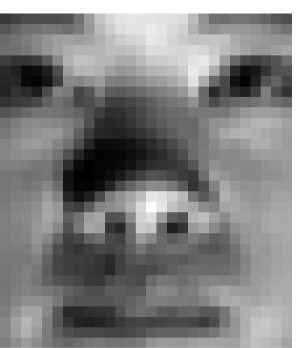}\\
(a) & (b)
\end{tabular}
\caption[An example of scaled test images]{An example of scaled test images. (a)~Original image and (b)~the image scaled by 1.313 vertically and 1.357 horizontally.}
\label{fig::zoom}
\end{figure}

\begin{table}[t]
\begin{minipage}{\columnwidth}\centering%
\caption[Recognition rate (in percentage) for scaled test images using SRC under various scaling factors]{Recognition rate (in percentage) for scaled test images using SRC \cite{wright:tpami09} under various scaling factors (SF).}
\label{table::zoom_global}
\begin{tabular}{| l |l|l|l|l|l|l|}
\hline
SF & 1 & 1.071  &  1.143 &   1.214  &  1.286  &  1.357\\
\hline
    1     & 100    &100   &  98.0 &  88.2 &  76.5 &  58.8\\
\hline
    1.063 &  99.7  & 96.5 &  86.1 &  68.5 &  50.3 &  37.6\\
\hline
    1.125 &  83.8  & 70.2 &  49.8 &  33.6 &  26.2 &  17.9\\
\hline
    1.188 &  54.5  & 43.7 &  26.8 &  20.0 &  18.0 &  12.6\\
\hline
    1.25  &  36.1  & 27.2 &  20.9 &  16.6 &  12.3 &  11.3\\
\hline
    1.313 &  31.5  & 24.3 &  16.7 &  13.9 &  10.6 &   9.8\\
\hline
\end{tabular}
\end{minipage}
\end{table}

\begin{table}[t]
\begin{minipage}{\columnwidth}\centering%
\caption[Recognition rate (in percentage) for scaled test images using proposed block-based approach under various SF]{Recognition rate (in percentage) for scaled test images using proposed block-based approach under various SF.}
\label{table::zoom_local}
\begin{tabular}{| l |l|l|l|l|l|l|}
\hline
SF & 1 & 1.071  &  1.143 &   1.214  &  1.286  &  1.357\\
\hline
    1     &  98.8   &  98.2 &  98.5 &  97.5  & 97.5  & 97.2\\
\hline
    1.063 &  97.5 &  96.7 &  96.0 &  96.0  & 93.5  & 93.4\\
\hline
    1.125 &  97.4   &  96.5 &  96.2 &  95.2  & 93.2  & 91.1\\
\hline
    1.188 &  94.9   &  92.9   &  91.6 &  89.4  & 87.1  & 83.3\\
\hline
    1.25  &  94.9 &  93.0 &  92.2   &  87.9    & 82.0  & 77.8\\
\hline
    1.313 &  90.7 &  90.4   &  84.1   &  81.0  & 75.5    & 64.2\\
\hline
\end{tabular}
\end{minipage}%
\end{table}

\subsection{Presence of Registration Errors}
\label{section::reg_rot_scale}
First, we show experimental results for test images under rotation and scaling operations. Test images are randomly rotated by an angle between -20 and 20 degrees, as illustrated by the example in Fig.~\ref{fig::rotation}. Fig.~\ref{fig::rotation_results} shows the recognition rate ($y$-axis) for each rotation degree ($x$-axis). We see that LSGM outperforms the SRC approach by a significant margin for the case of severe misalignment.

For the second set of experiments, the test images are stretched in both directions by scaling factors up to 1.313 vertically and 1.357 horizontally. An example of an aligned image in the database and its distorted version to be tested are shown in Fig.~\ref{fig::zoom}. The benefits of LSGM over SRC are apparent from Tables~\ref{table::zoom_global} and~\ref{table::zoom_local} which show the percentage of correct identification with various scaling factors. Finally, we compare the performance of our LSGM approach with five other algorithms: SRC, Eigen-NS, Eigen-SVM, Fisher-NS and Fisher-SVM, for the scenario where the test images are scaled by a horizontal factor of 1.214 and a vertical factor of 1.063. The overall recognition rates are shown in Table \ref{table::regnerr}.

\begin{table}[t]
\begin{center}
\caption[Overall recognition rate (as a percentage) for the scenario of scaling by horizontal and vertical factors of 1.214 and 1.063 respectively]{Overall recognition rate (as a percentage) for the scenario of scaling by horizontal and vertical factors of 1.214 and 1.063 respectively.}
\begin{tabular}{|c|c|c|c|c|}
  \hline
  Method & Recognition rate ($\%$)\\
  \hline
  LSGM & 89.4 \\
  SRC & 60.8 \\
  Eigen-NS & 55.5 \\
  Eigen-SVM & 56.7\\
  Fisher-NS & 54.1\\
  Fisher-SVM & 57.1 \\
  \hline
\end{tabular}
\label{table::regnerr}
\end{center}
\vspace{-3mm}
\end{table}

\begin{table}[t]
\begin{center}
\caption[Overall recognition rate (as a percentage) for the scenario where test images are scaled and subjected to random pixel corruption]{Overall recognition rate (as a percentage) for the scenario where test images are scaled and subjected to random pixel corruption.}
\begin{tabular}{|c|c|c|c|c|}
  \hline
  Method & Recognition rate ($\%$)\\
  \hline
  LSGM & 96.3 \\
  SRC & 93.2 \\
  Eigen-NS & 54.3 \\
  Eigen-SVM & 58.5\\
  Fisher-NS & 56.2 \\
  Fisher-SVM & 59.9 \\
  \hline
\end{tabular}
\label{table::pixcorr}
\end{center}
\end{table}

\subsection{Recognition Under Random Pixel Corruption}
\label{section::pixcorr}
We randomly corrupt 50\% of the image pixels in each test image. In addition, each test image is scaled by a horizontal factor of 1.071 and a vertical factor of 1.063. Local sparse features are extracted using the robust form of the $\ell_1$-minimization similar to the approach in \cite{wright:tpami09}. The overall recognition rates are shown in Table \ref{table::pixcorr}. These results reveal that under the mild scaling distortion scenario, our LSGM approach retains the robustness characteristic of the global sparsity approach (SRC), while the other competitive algorithms suffer drastic degradation in performance.

\begin{figure}[t]
\vspace{-3mm}
\centering
\includegraphics[scale=0.6]{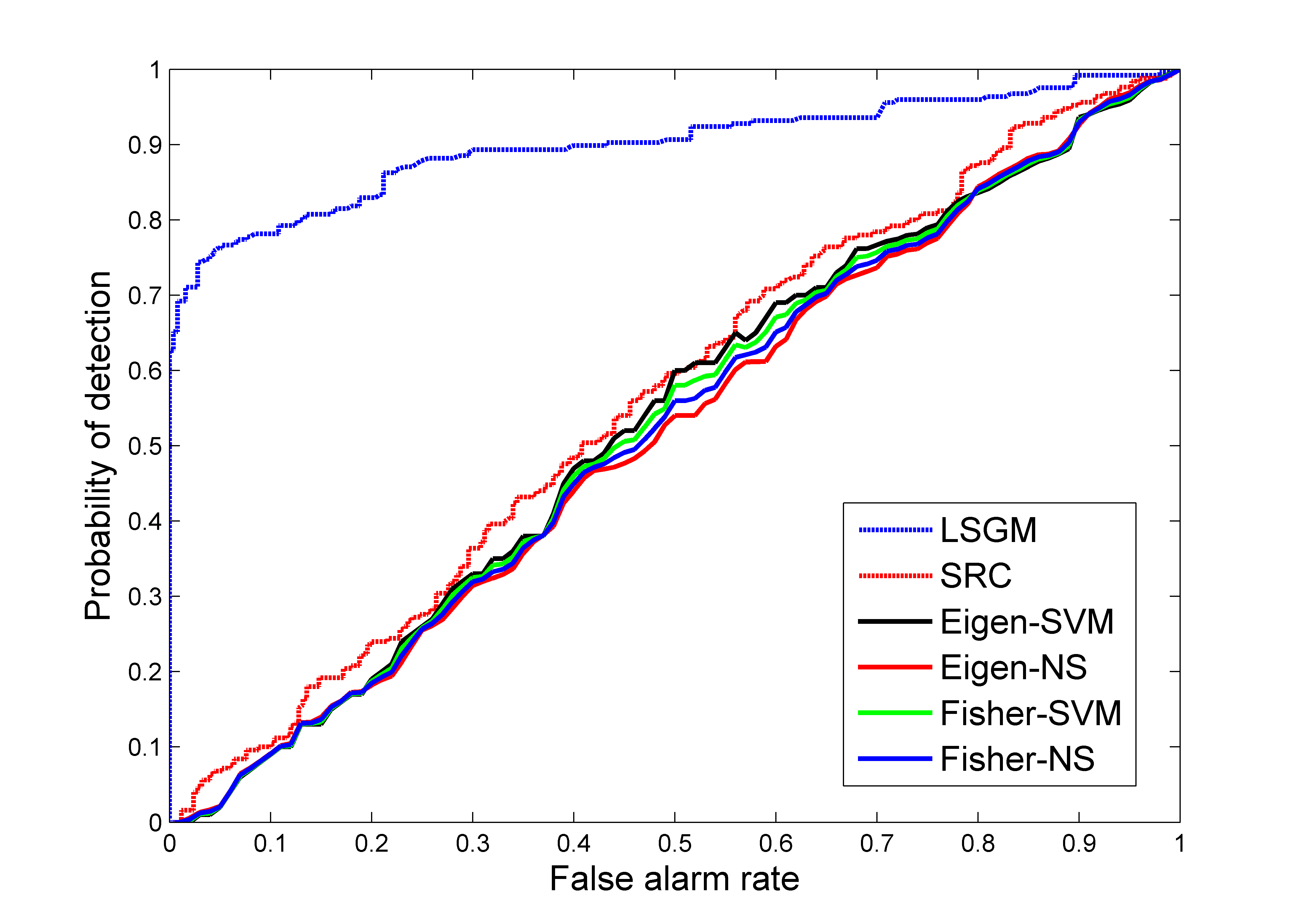}
\vspace{-3mm}
\caption[ROC curves for outlier rejection]{ROC curves for outlier rejection.}
\label{fig::ROC_rotate5}
\end{figure}

\subsection{Outlier Rejection}
\label{section::outreject}
In this experiment, samples from 19 of the 38 classes in the Yale database are included in the training set, and faces from the other 19 classes are considered outliers. For training, 15 samples per class from Subsets 1 and 2 are used~($19\times15 = 285$ samples in total), while 500 samples are randomly chosen for testing, among which 250 are inliers and the other 250 are outliers. All test samples are rotated by five degrees.

The five different competing approaches are compared with our proposed LSGM method. For the LSGM approach, we use a minimum threshold $\delta$ in the decision rule. If the maximum value of the log-likelihood ratio does not exceed $\delta$, the corresponding test sample is labeled an outlier. In the SRC approach, the Sparsity Concentration Index is used as the criterion for outlier rejection. For the other approaches under comparison which use the nearest subspace and SVM classifiers, reconstruction residuals are compared to a threshold to decide outlier rejection. The ROC curves for all the approaches are shown in Fig.~\ref{fig::ROC_rotate5}. LSGM offers the best performance, while some of the approaches are actually worse than random guessing.

\section{Conclusion}
In this chapter, we have demonstrated two additional applications of our graph-based feature fusion framework first introduced in Chapter \ref{chapter:gm}. The novelty of our contribution is in the design of multiple sets of \emph{sparse} representations that can then be fused via discriminative trees. The consistently superior performance of our approach in a variety of experimental scenarios offers proof of the wide flexibility of the fusion framework. The two central ideas in this dissertation are those of probabilistic graphical models and the theory of sparse signal representations. The contributions in this chapter constitute our first attempt to bring the discriminative benefits of these two approaches together for robust image classification. Looking ahead, the latter half of the following chapter formalizes this relationship by learning graphical spike-and-slab priors directly for sparse features in a Bayesian set-up.

\chapter{Structured Sparse Representations for Robust Image Classification}
\label{chapter:struct_sparsity}

\section{Introduction}
\label{ch4:intro}

This chapter primarily concerns itself with learning discriminative models on sparse signal representations. As outlined in Chapter \ref{chapter:intro}, our goal is to understand the discriminative structure in multiple feature representations from the standpoint of improving robustness in image classification. Here, we learn models on sparse feature representations in interesting ways.

First, we introduce a simultaneous sparsity model for classification scenarios that are multi-modal in nature. Example real-world manifestations of this scenario occur in the form of spatially local pixels for hyperspectral target classification \cite{Chen_2011c}, multiple feature sets for automatic image annotation \cite{zhang:smcb12}, kernel features for object categorization, or as query images for video-based face recognition \cite{yuan:tip12}. In the latter two examples, each event is in fact represented using multiple heterogeneous sources, resulting in multi-task versions of SRC \cite{wright:tpami09}. We consider a specific instantiation of the multi-task framework for the classification of color medical images acquired by histopathology. Digital histopathological images are comprised of three color channels - red, green and blue. The tissue staining process central to histopathology imbues the images with prominent red and blue hues. In fact, color information is used as a crucial visual cue by pathologists to identify whether a tissue is healthy or diseased. Our simultaneous sparsity model builds color-specific dictionaries and solves for sparse coefficients with constraints guided by the color channel correlations.

The success of SRC comes with a caveat. The validity of the linear representation model rests on the supposition that if enough diversity (in imaging conditions, for example) is captured by the set of training images, any new test image can be approximated very well using a small number of training images. In practice however, many applications have the limitation that rich training is not available \emph{a priori}. Examples include automatic target recognition using radar images and hyperspectral target classification. The linear representation model assumption is violated in the regime of low training. This important issue concerning sparsity-based classification methods has not been investigated thoroughly so far.

Accordingly, the second half of this chapter explores another way of learning discriminative models on sparse coefficients, this time in the form of \emph{class-specific} probabilistic priors. We learn class-specific parameters for spike-and-slab priors, which have been successful in modeling sparse signals. The main advantage of working in a Bayesian set-up is the robustness to training insufficiency. We also show that hierarchical extensions of the spike-and-slab prior lead to new optimization formulations that capture group sparsity structure for multi-task scenarios.

\section{Histopathological Image Classification: Overview}

The advent of digital pathology \cite{mendez:mp98} has ushered in an era of computer-assisted diagnosis and treatment of medical conditions based on the analysis of medical images. Of active research interest is the development of quantitative image analysis tools to complement the efforts of radiologists and pathologists towards better disease diagnosis and prognosis. This research thrust has been fueled by a variety of factors, including the availability of large volumes of patient-related medical data, dramatic improvements in computational resources (both hardware and software), and algorithmic advances in image processing, computer vision and machine learning theory. An important emerging sub-class of problems in medical imaging pertains to the analysis and classification of histopathological images \cite{boucheron:thesis08,madabhushi:im09,gurcan:rbe09,fox:jcp00}. Whole slide digital scanners process tissue slides to generate these digital images. Examples of histopathological images are shown in Figs. \ref{fig:hist_adl} and \ref{fig:hist_ibl}. It is evident that these images carry rich structural information, making them invaluable for the diagnosis of many diseases including cancer \cite{alexe:ebm09,basavanhally:tbme10,dundar:tbe11}.

\subsection{Prior Work}
Pathologists often look for visual cues at the nuclear and cellular level in order to categorize a tissue image as either healthy or diseased. Motivated by this, a variety of low-level image features have been developed based on texture, morphometric characteristics (shape and spatial arrangement) and image statistics. The gray level co-occurrence matrix by Haralick \emph{et al.} \cite{haralick:tsmc73} estimates the texture of an image in terms of the distribution of co-occurring pixel intensities at specified offset positions. Morphological image features \cite{serra:book82} have been used in medical image segmentation for detection of vessel-like patterns \cite{zana:tip01}. Image histograms are a popular choice of features for medical imaging \cite{chapelle:tnn99}. Wavelet features have been deployed for prostate cancer diagnosis in \cite{wetzel:spie99}. Esgiar \emph{et al.} \cite{esgiar:titbm02} have captured the self-similarity in colon tissue images using fractal-based features. Tabesh \emph{et al.} \cite{tabesh:tmi07} have combined color, texture and morphometric features for prostate cancer diagnosis. Doyle \emph{et al.} \cite{doyle:isbi08} introduced graph-based features using Delaunay triangulation and minimum spanning trees to exploit spatial structure. Orlov \emph{et al.} \cite{orlov:prl08,shamir:scbm08} have recently proposed a multi-purpose feature set that aggregates transform domain coefficients, image statistics and texture information. Experimental success in many different classification problems has demonstrated the versatility of this feature set. It must be mentioned that all the  features discussed above are applicable broadly for image analysis and have been particularly successful in medical imaging. For classification, these features are combined with powerful classifiers such as SVMs \cite{vapnik:book95,chapelle:tnn99} and boosting \cite{freund:jsai99,basavanhally:isbi11}. A comprehensive discussion of features and classifiers for histopathological analysis is provided in \cite{gurcan:rbe09}.

\subsection{Motivation and Challenges}
While histopathology shares some commonalities with other popular imaging modalities such as cytology and radiology, it also exhibits two principally different characteristics \cite{hipp:jpi11} that pose challenges to image analysis. First, histopathological images are invariably multi-channel in nature (commonly using three color channels - red, green and blue (RGB)). Key geometric information is spread across the color channels. It is well known that color information in the hematoxylin-eosin (H$\&$E) stained slides is essential to identify the discriminative image signatures of healthy and diseased tissue \cite{naik:isbi08,sertel:jsps09}. Specifically, the nuclei assume a bluish tinge due to hematoxylin, while the cytoplasmic structures and connective tissue appear red due to eosin. As seen from Fig. \ref{fig:hist_adl}, there is a higher density of nuclei in diseased tissue. Typically in histopathological image analysis, features are extracted from each color channel of the images \cite{tabesh:tmi07}, and the classifier decisions based on the individual feature sets are then fused for classification. Alternately, only the luminance channel information - image edges resulting mainly from illumination variations - is considered \cite{sertel:jsps09}. The former approach ignores the inherent correlations among the RGB channels while the latter strategy fails to exploit chrominance channel geometry, i.e. the edges and image textures caused by objects with different chrominance.

The second challenge posed by histopathology is the relative difficulty in obtaining good features for classification due to the geometric richness of tissue images. Tissues from different organs have structural diversity and often, the objects of interest occur at different scales and sizes \cite{gurcan:rbe09}. As a result, features are usually customized for specific classification problems, most commonly cancer of the breast and prostate.

In this chapter, we address both these challenges through a novel simultaneous sparsity model inspired by recent work using sparse representations for image classification \cite{wright:tpami09}. SRC has been proposed earlier for single-channel medical images, in cervigram segmentation \cite{zhang:isbi10,yu:isbi11} and colorectal polyp and lung nodule detection \cite{liu:miccai11}. To the best of our knowledge, ours is the first discriminative sparsity model for multi-channel histopathological images.

\subsection{Overview of Contributions}
The relevance of color information for discriminative tasks has been identified previously \cite{liu:tip10,wang:tip11}. We propose a new simultaneous Sparsity model for multi-channel Histopathological Image Representation and Classification (SHIRC). Essentially, our model recognizes the diversity of information in multiple color channels and extends the standard SRC approach \cite{wright:tpami09,Chen_2011c,zhang:taes12,wagner:tpami12,zhang:isbi10,yu:isbi11,liu:miccai11} by designing three color dictionaries, corresponding to the RGB channels. Each multi-channel histopathological image is represented as a sparse linear combination of training examples under suitable channel-wise constraints which capture color correlation information. The constraints agree with intuition since a sparse linear model for a color image necessitates identical models for each of its constituent color channels with no cross-channel contributions. 

Our approach considers a multi-task scenario that is qualitatively similar to the visual classification problem addressed very recently by Yuan \emph{et al.} \cite{yuan:tip12}. In \cite{yuan:tip12}, three types of image features - containing texture, shape and color information - are extracted from images and a joint sparsity model is proposed to classify the images. The joint (simultaneous) sparsity model employed in \cite{yuan:tip12} and the one we develop are however significantly different. First, \cite{yuan:tip12} does not consider the problem of multi-channel or color image modeling. Second and more crucially, the cost function in \cite{yuan:tip12} is a sum of reconstruction error terms from each of the feature dictionaries which results in the commonly seen {\em row sparsity} structure on the sparse coefficient matrix. The resulting optimization problem is solved using the popular Accelerated Proximal Gradient (APG) method \cite{nesterov:tr07}. In our work however, to conform to imaging physics, we introduce color channel-specific constraints on the structure of the sparse coefficients, which do {\em not} directly conform to row sparsity, leading to a new optimization problem. This in turn requires a modified greedy matching pursuit approach to solve the problem.

As discussed earlier, feature design in medical imaging is guided by the object-based perception of pathologists. Depending on the type of tissue, the objects could be nuclei, cells, glands, lymphocytes, etc. Crucially it is the presence or absence of these local objects in an image that matters to a pathologist; their absolute spatial location matters much less. As a result, the object of interest may be present in the test image as well as the representative training images, albeit at different spatial locations, causing a seeming breakdown of the image-level SHIRC. This scenario can occur in practice if the acquisition process is not carefully calibrated. So we infuse the SHIRC with a robust locally adaptive flavor by developing a Locally Adaptive SHIRC (LA-SHIRC). We rely on the pathologist's insight to carefully select multiple local regions that contain these objects of interest from training as well as test images and use them in our linear sparsity model instead of the entire images. Local image features often possess better discriminative power than image-level features \cite{zou:tip07}. LA-SHIRC is a well-founded instantiation of this idea to resolve the issue of correspondence between objects at different spatial locations. LA-SHIRC offers flexibility in the number of local blocks chosen per image and the size of each such block (tunable to the size of the object). Experimentally, a beneficial consequence of LA-SHIRC is the reduced burden on the number of training images required. 

\section{A Simultaneous Sparsity Model for Histopathological Image Representation and Classification}
\label{sec:global}

Section \ref{secch3:src} has identified the central analytical formulation underlying the simultaneous sparsity methods in literature. Typically a single dictionary $\mat D$ is used, as in \cite{Chen_2011c}. In other cases, each event is in fact characterized by multiple heterogeneous sources \cite{zhang:smcb12,yuan:tip12}, resulting in multi-task versions of SRC. Although different dictionaries are used for the different sources, the issue of correlation among different representations of the same image is not thoroughly investigated. Our contribution is an example of multi-task classification, with separate dictionaries designed from the RGB channels of histopathological images.

For ease of exposition, we consider the binary classification problem of classifying images as either healthy or diseased. $\mat D_{h}$ and $\mat D_{d}$ indicate the training dictionaries of healthy and diseased images respectively. A total of $N$ training images are chosen. We represent a color image as a matrix $\mat Y := [\vect y^r ~ \vect y^g ~ \vect y^b]\in \mathbb{R}^{n\times 3}$, where the superscripts $r,g,b$ correspond to the RGB color channels respectively. The dictionary $\mat D$ is redefined as the concatenation of three color-specific dictionaries, $\mat D := \left[\mat{D}^{r}\ \; \mat{D}^{g}\ \; \mat{D}^{b}\right]\in\mathbb{R}^{n\times3N}$. Each color dictionary $\mat D^c, c \in \{r,g,b\}$, is the concatenation of sub-dictionaries from both classes belonging to the $c$-th color channel,
\be
\mat D^c := [\mat D_{h}^{c}\ \; \; \mat D_{d}^{c}], c \in \{r,g,b\}.
\ee
\be
\Rightarrow \mat D := [\mat D^r\ \; \mat D^g\ \; \mat D^b] = [\mat D_{h}^{r}\ \; \mat D_{d}^{r}\ \; \; \mat D_{h}^{g}\ \; \mat D_{d}^{g}\ \; \; \mat D_{h}^{b}\ \; \mat D_{d}^{b}].
\ee
The color dictionaries are designed to obey column correspondence, i.e., the $i$-th column from each of the color dictionaries $\mat D^c$ taken together correspond to the $i$-th training image. Fig. \ref{fig:global} shows the arrangement of training images into channel-specific dictionaries. A test image $~\mat Y$ can now be represented as a linear combination of training samples:
\be
\mat Y = \mat D \mat S = \left[\mat D_{h}^{r}\ \; \mat D_{d}^{r}\ \; \mat D_{h}^{g}\ \; \mat D_{d}^{g}\ \; \mat D_{h}^{b}\ \; \mat D_{d}^{b}\right]\left[\vect \alpha^r ~~ \vect \alpha^g ~~\vect \alpha^b\right],
\ee
where the coefficient vectors $\vect \alpha^c \in \mathbb{R}^{3N}, c \in \{r,g,b\}$, and $\mat S = \left[\vect \alpha^r ~~ \vect \alpha^g ~~\vect \alpha^b\right] \in \mathbb{R}^{3N \times 3}$.

\begin{figure}[t]
  \centering
  \includegraphics[scale=0.6]{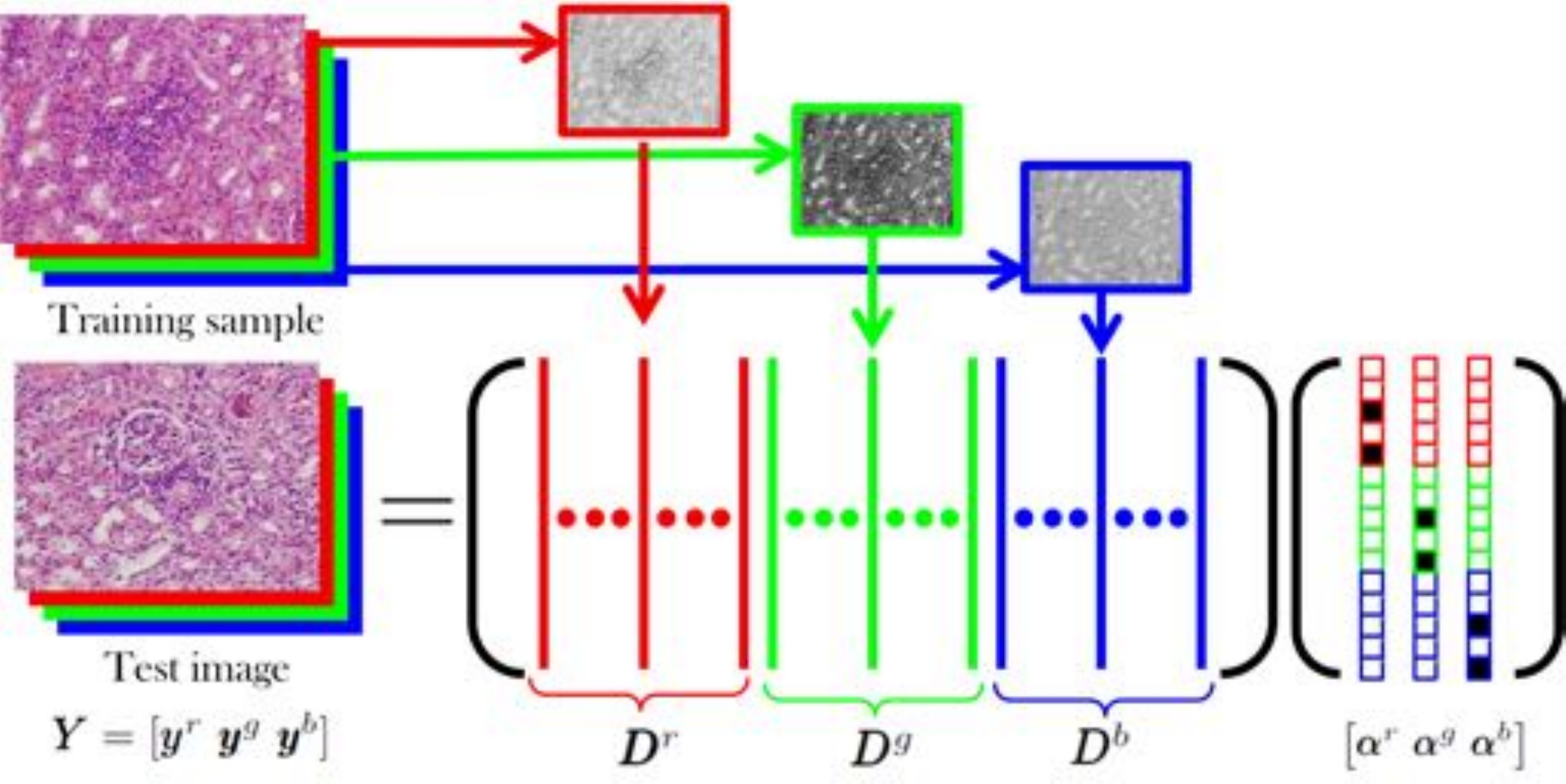}
  \caption[Color channel-specific dictionary design for SHIRC]{Color channel-specific dictionary design for SHIRC. The constituent RGB channels of the $i$-th sample training image occupy the $i$-th columns of the dictionaries $\mat D^r, \mat D^g$ and $\mat D^b$ respectively. Coefficient vectors $\vect \alpha^r, \vect \alpha^g$ and $\vect \alpha^b$ are color-coded to indicate the dictionary corresponding to each coefficient. Filled-in blocks indicate non-zero coefficients. The filling pattern illustrates that \emph{identical} linear representation models hold for each color channel of the test image, with possibly different weights in the coefficient vector.}
  \label{fig:global}
\end{figure}

A closer investigation of $\mat S$ reveals interesting characteristics:
\begin{enumerate}
  \item It is reasonable to assume that the $c$-th channel of the test image (i.e. $\vect y^c$) can be represented by the linear span of the training samples belonging to the $c$-th channel alone (i.e. only those training samples in $\mat D^c$). So the columns of $\mat S$ ideally have the following structure: \[
\vect \alpha^r = \left[\begin{array}{c}
\vect \alpha_h^r\\
\vect \alpha_d^r\\
\vect 0\\
\vect 0 \\
\vect 0\\
\vect 0
\end{array}\right],
\vect \alpha^g = \left[\begin{array}{c}
\vect 0\\
\vect 0\\
\vect \alpha_h^g\\
\vect \alpha_d^g\\
\vect 0\\
\vect 0
\end{array}\right],
\bm{\alpha}_{b}=\left[\begin{array}{c}
\vect 0\\
\vect 0\\
\vect 0\\
\vect 0 \\
\vect \alpha_h^b\\
\vect \alpha_d^b
\end{array}\right],
\]
where $\vect 0$ denotes the conformal zero vector. In other words, $\mat S$ exhibits block-diagonal structure.
  \item Each color channel representation $\vect y^c$ of the test image is in fact a \emph{sparse} linear combination of the training samples in $\mat D^c$. Suppose the image belongs to class $h$ (healthy); then only those coefficients in $\vect \alpha^c$ that correspond to $\mat D_h^c$ are expected to be non-zero.
  \item The locations of non-zero weights of color training samples in the linear combination exhibit one-to-one correspondence across channels. If the $j$-th training sample in $\mat D^r$ has a non-zero contribution to $\vect y^r$, then for $c \in \{g,b\}$, $\vect y^c$ has non-zero contribution from the $j$-th training sample in $\mat D^c$.
\end{enumerate}

This immediately suggests a joint sparsity model similar to \eqref{eqn::joint_sparse_recovery}. However, the row sparsity constraint leading to the SOMP solution is not obvious from this formulation. Instead, we introduce a new matrix $\mat S' \in \mathbb{R}^{N \times 3}$ as the transformation of $\mat S$ with the redundant zero coefficients removed,
\be
\mat S' = \left[\begin{array}{ccc}
\vect \alpha_{h}^{r} & \vect \alpha_{h}^{g} & \vect \alpha_{h}^{b}\\
\vect \alpha_{d}^{r} & \vect \alpha_{d}^{g} & \vect \alpha_{d}^{b}\\
\end{array}\right].
\ee
This is possible by first defining $\mat H \in \mathbb{R}^{3N \times 3}$ and $\mat J \in \mathbb{R}^{N\times 3N}$,
\be
\mat H = \left[\begin{array}{ccc}
\vect 1_N & \vect 0 & \vect 0\\
\vect 0 & \vect 1_N & \vect 0\\
\vect 0 & \vect 0 & \vect 1_N
\end{array}\right],~\mat J = \left[\mat I_N\quad \mat I_N\quad \mat I_N\right],
\ee
where $\vect 1_N \in \mathbb{R}^{N}$ is the vector of all ones, and $\mat I_N$ denotes the $N$-dimensional identity matrix. Now,
\be
\mat S' = \mat J \left(\mat H\circ\mat S\right),
\label{eq:S-prime}
\ee
where $\circ$ denotes the Hadamard product, $\left(\mat H \circ \mat S\right)_{ij}\triangleq h_{ij}s_{ij}~\forall~ i,j$. Finally, we formulate a sparsity-enforcing optimization problem that is novel to the best of our knowledge:
\begin{eqnarray}
\hat{\mat S}' = \arg\min \left\Vert \mat S'\right\Vert_{\text{row},0} & \mbox{subject to} & \left\Vert\mat Y - \mat D \mat S \right\Vert_{F}\leq\epsilon.
\label{eq:sparsity_main}
\end{eqnarray}
Solving the problem in (\ref{eq:sparsity_main}) presents a challenge in that a straightforward application of SOMP \cite{tropp:sp06a} is not possible due to the non-invertibility of the Hadamard operator. We have developed a greedy algorithmic modification of SOMP that fares well in practice. The analytical details of the algorithm are presented in Appendix \ref{sec:appA}.

The final classification decision is made by comparing the class-specific reconstruction errors to a threshold $\tau$,
\be
R(\mat Y) = \norm{\mat Y - \mat D_h \hat{\mat S}_h}_2 - \norm{\mat Y - \mat D_d \hat{\mat S}_d}_2 \underset{d}{\overset{h}{\gtrless}} \tau,
\label{eq:new_class_residual}
\ee
where $\hat{\mat S}_h$ and $\hat{\mat S}_d$ are matrices whose entries are those in $\hat{\mat S}$ associated with $\mat D_h$ and $\mat D_d$ respectively. Our approach extends to the $K$-class scenario in a straightforward manner by incorporating additional class-specific dictionaries in $\mat D$. The classification rule is then modified as follows:
\be
\mbox{Class}(\mat Y) = \arg\min_{k = 1,\ldots,K} \left\Vert \mat Y - \mat D\delta_{k}(\hat{\mat S})\right\Vert_{F},
\label{eq:residual multiclass}
\ee
where $\delta_{k}(\hat{\mat S})$ is the matrix whose only non-zero entries are the same as those in $\hat{\mat S}$ associated with class $C_k$.

\section{LA-SHIRC: Locally Adaptive SHIRC}
\label{sec:local}

Some histopathological image collections present a unique practical challenge in that the presence or absence of desired objects (e.g. cells, nuclei) in images is more crucial - compared to their actual locations - for pathologists for disease diagnosis. Consequently, if these discriminative objects (sub-images) are not in spatial correspondence in the test and training images, it would seem that SHIRC cannot handle this scenario. Fig. \ref{fig:la-shirc} illustrates this for sample images from the IBL data set.

This issue can be handled practically to some extent by careful pre-processing that manually segments out the objects of interest for further processing \cite{dundar:tbe11}. However this approach causes a loss in contextual information that is also crucial for pathologists to make class decisions. We propose a robust algorithmic modification of SHIRC, known as Locally Adaptive SHIRC (LA-SHIRC), to address this concern.
\begin{figure}[t]
  \centering
  \includegraphics[scale=0.75]{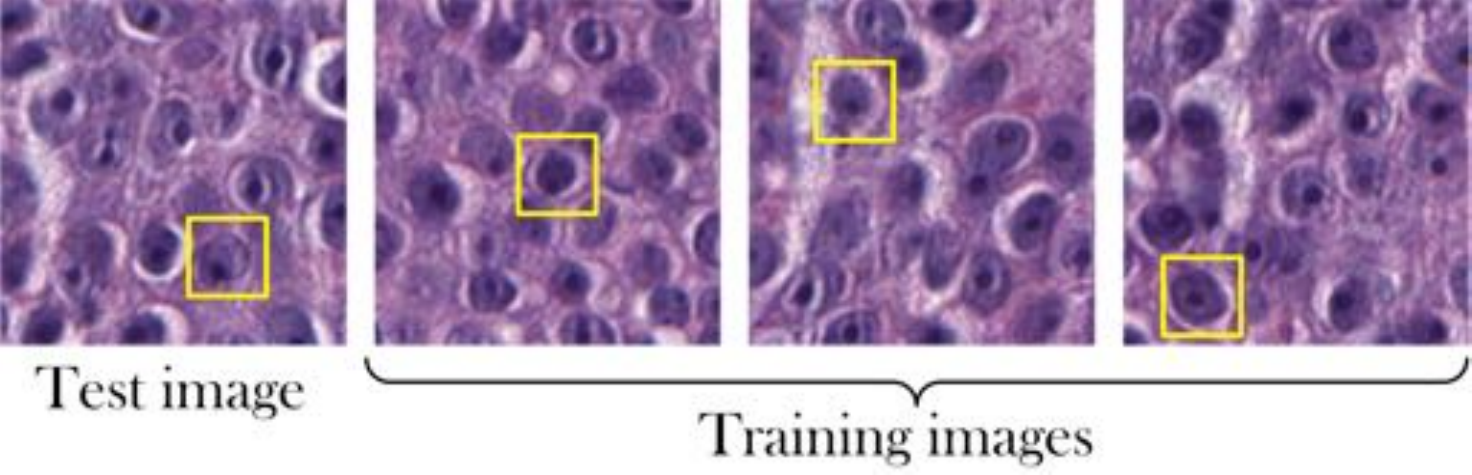}
  \caption[Illustration to motivate LA-SHIRC]{Illustration to motivate LA-SHIRC. Shown here are four images from the IBL data set. The test image cannot be represented accurately as a linear combination of the training images. However, the four local regions marked in yellow (individual cells in the tissue) have structural similarities, although they are located at different spatial locations in the image.}
  \label{fig:la-shirc}
\end{figure}

\begin{figure}[t]
  \centering
  \subfigure{\includegraphics[scale=1]{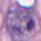}}
	\subfigure{\includegraphics[scale=1]{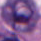}}
	\subfigure{\includegraphics[scale=1]{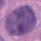}}
	\subfigure{\includegraphics[scale=1]{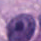}}
	\subfigure{\includegraphics[scale=1]{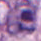}}
	\subfigure{\includegraphics[scale=1]{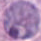}}
	\subfigure{\includegraphics[scale=1]{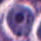}}
	\subfigure{\includegraphics[scale=1]{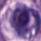}} \\
  \subfigure{\includegraphics[scale=1]{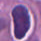}}
	\subfigure{\includegraphics[scale=1]{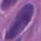}}
	\subfigure{\includegraphics[scale=1]{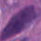}}
	\subfigure{\includegraphics[scale=1]{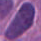}}
	\subfigure{\includegraphics[scale=1]{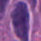}}
	\subfigure{\includegraphics[scale=1]{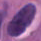}}
	\subfigure{\includegraphics[scale=1]{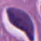}}
	\subfigure{\includegraphics[scale=1]{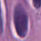}}
  \caption[Local cell regions selected from the IBL data set]{Local cell regions selected from the IBL data set. Top row: DCIS (actionable), bottom row: UDH (benign). Structural differences between the two classes are readily apparent.}
  \label{fig:local_roi}
\end{figure}

It is well known that local image features are often more useful than global features from a discriminative standpoint \cite{zou:tip07}. This also conforms with the intuition behind pathologists' decisions. Accordingly, we modify SHIRC to classify local image objects instead of the global image. Several local objects of the same dimension (vectorized versions lie in $\mathbb{R}^m, m \ll n$) are identified in each training image based on the recommendation of the pathologist. Fig. \ref{fig:local_roi} shows individual cells from the IBL data set. In fact, these obvious differences in cell structure have been exploited for classification \cite{dundar:tbe11} by designing morphological features such as cell perimeter and ratio of major-to-minor axis of the best-fitting ellipse.

The dictionary $\mat D$ in SHIRC is now replaced by a new dictionary $\bar{\mat D}$ that comprises the local blocks. In Fig. \ref{fig:local}, the yellow boxes indicate the local regions. Assuming $N$ full-size training images, the selection of $B$ local blocks per image results in a training dictionary of size $NB$. Note that even for fixed $N$, the dictionary $\bar{\mat D} \in \mathbb{R}^{m\times NB}$ has more samples (or equivalently, leads to a richer representation) than $\mat D \in \mathbb{R}^{n \times N}$. Therefore, a test image block is expressed as a sparse linear combination of \emph{all} local image blocks from the training images. $B$ blocks are identified in every test image, and a class decision is obtained for each such block by solving \eqref{eq:sparsity_main} but with the dictionary $\bar{\mat D}$. Finally the identity of the overall image is decided by combining the local decisions from its constituent blocks. The \emph{adaptive} term in the name LA-SHIRC indicates the flexibility offered by the algorithm in terms of choosing the number of blocks $B$ and their dimension $m$. Objects of interest in histopathological image analysis exist at different scales \cite{gurcan:rbe09} and the tunability of LA-SHIRC makes it amenable to a variety of histopathological classification problems.

\begin{figure}[t]
  \centering
  \includegraphics[scale=0.6]{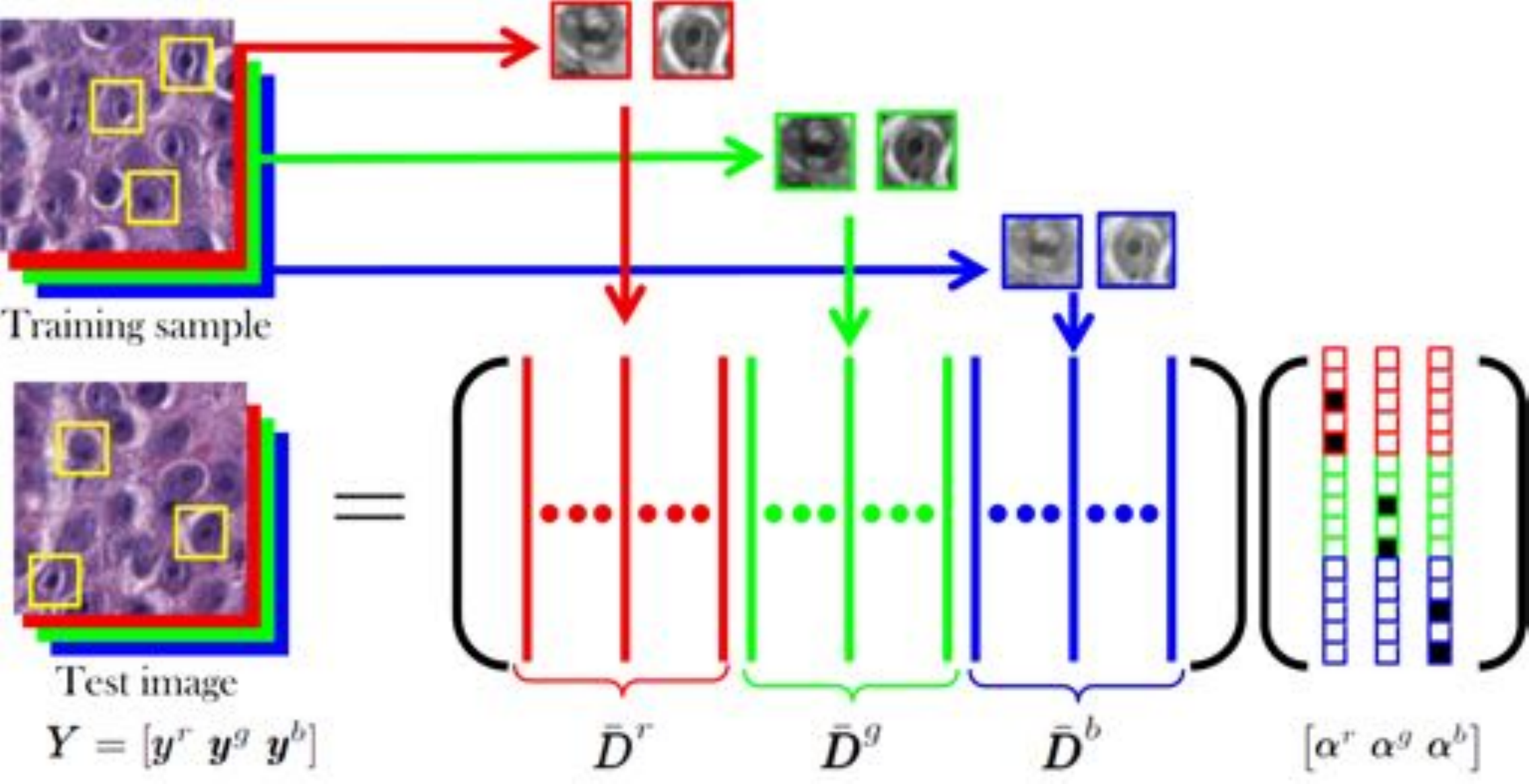}
  \caption[LA-SHIRC: Locally adaptive variant of SHIRC]{LA-SHIRC: Locally adaptive variant of SHIRC. The black boxes indicate local objects of interest such as cells and nuclei. The new dictionary $\bar{\mat D}$ is built using multiple local blocks from each training image. In every test image, the local objects are classified using the simultaneous sparsity model and their decisions are fused for overall image-level classification.}
  \label{fig:local}
\end{figure}

LA-SHIRC satisfactorily handles the issue of spatial correspondence of objects. Additionally, as will be demonstrated through experiments in Section \ref{sec:trng_size}, it ensures high classification accuracy even with a small number of (global) training images. This has high practical relevance since generous number of training histopathological images per condition (healthy/diseased) may not always be available. The improved performance however comes at the cost of increased computational complexity since $B$ optimization problems need to be solved and the dictionary $\bar{\mat{D}}$ has considerably more columns than $\mat{D}$.\\

\noindent \textbf{Decision Fusion:} A natural way of combining local class decisions is majority voting. Suppose $\mat Y_i, i = 1,\ldots,B$, represent the $B$ local blocks from image $\mat Y$. Then,
\be
\text{Class}\left(\mat Y\right) = \max_{k = 1,\ldots,K}\left|\left\{i: \text{Class}\left(\mat Y_i\right) = k\right\}\right|, i = 1,\ldots,B,
\label{eq:voting}
\ee
where $|\cdot|$ denotes set cardinality and $\text{Class}\left(\mat Y_i\right)$ is determined by \eqref{eq:residual multiclass}. We use a more intuitive approach to fuse individual decisions based on a heuristic version of maximum likelihood. Let $\hat{\mat S}_i$ be the recovered sparse representation matrix of the block $\mat Y_i$. The probability of $\mat Y_i$ belonging to the $k$-th class is defined to be inversely proportional to the residual associated with the dictionary atoms in the $k$-th class:
\be
p^k_i = P\left(\text{Class}\left(\mat Y_i\right) = k\right) = \frac{1/R^k_i}{\sum_{k=1}^K\left(1/R^k_i\right)},
\label{eq:prob}
\ee
where $R^k_i = \norm{\mat Y_i-\bar{\mat D}\vect \delta_k\left(\hat{\mat S}_{i}\right)}_2$. The identity of the test image $\mat Y$ is then given by:
\be
\text{Class}\left(\mat Y\right) = \arg\max_{k = 1,\ldots,K}\left(\prod_{i=1}^B p^k_i\right).
\label{eq:local_prob_combine}
\ee

\section{Validation and Experimental Results}
\label{ch4:results}

\subsection{Experimental Set-Up: Image Data Sets}
\label{sec:datasets}

We compare the performance of SHIRC and LA-SHIRC against state-of-the-art alternatives for two challenging real-world histopathological image data sets.

\begin{figure}[t]
  \centering
  \subfigure[Healthy lung.]{\includegraphics[scale=0.13]{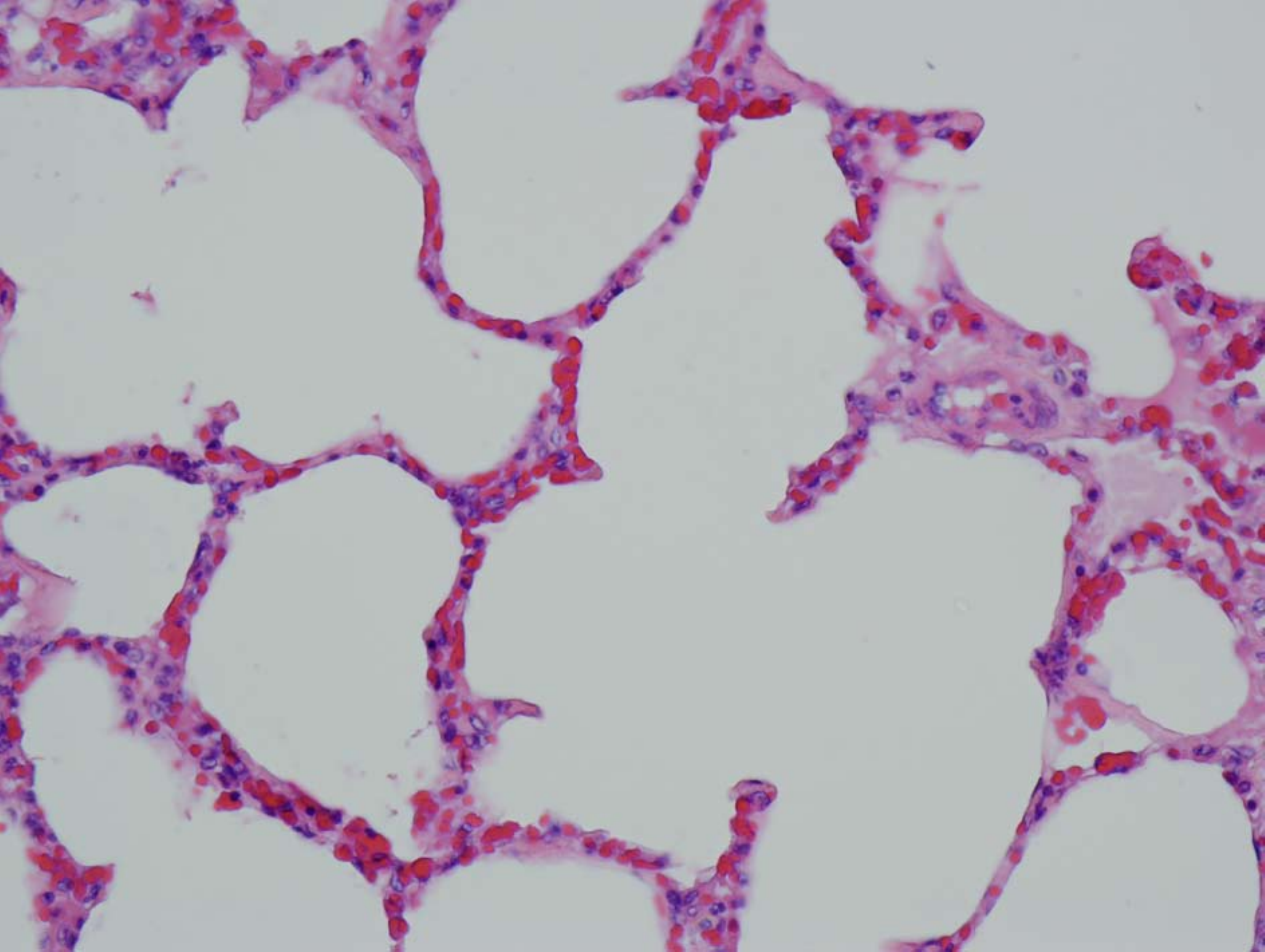}} \hspace{2mm}
	\subfigure[Healthy lung.]{\includegraphics[scale=0.13]{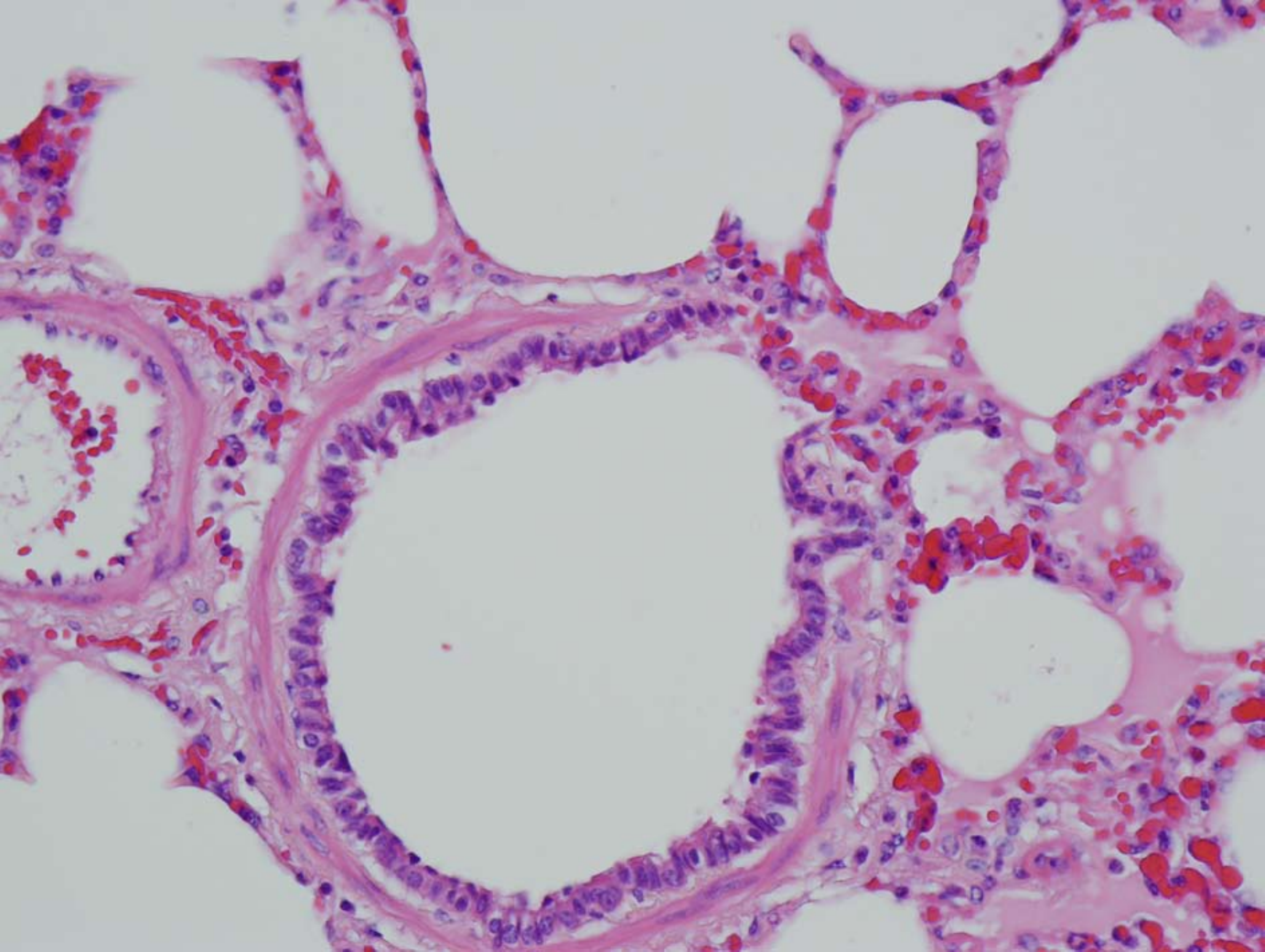}} \hspace{2mm}
	\subfigure[Inflamed lung.]{\includegraphics[scale=0.13]{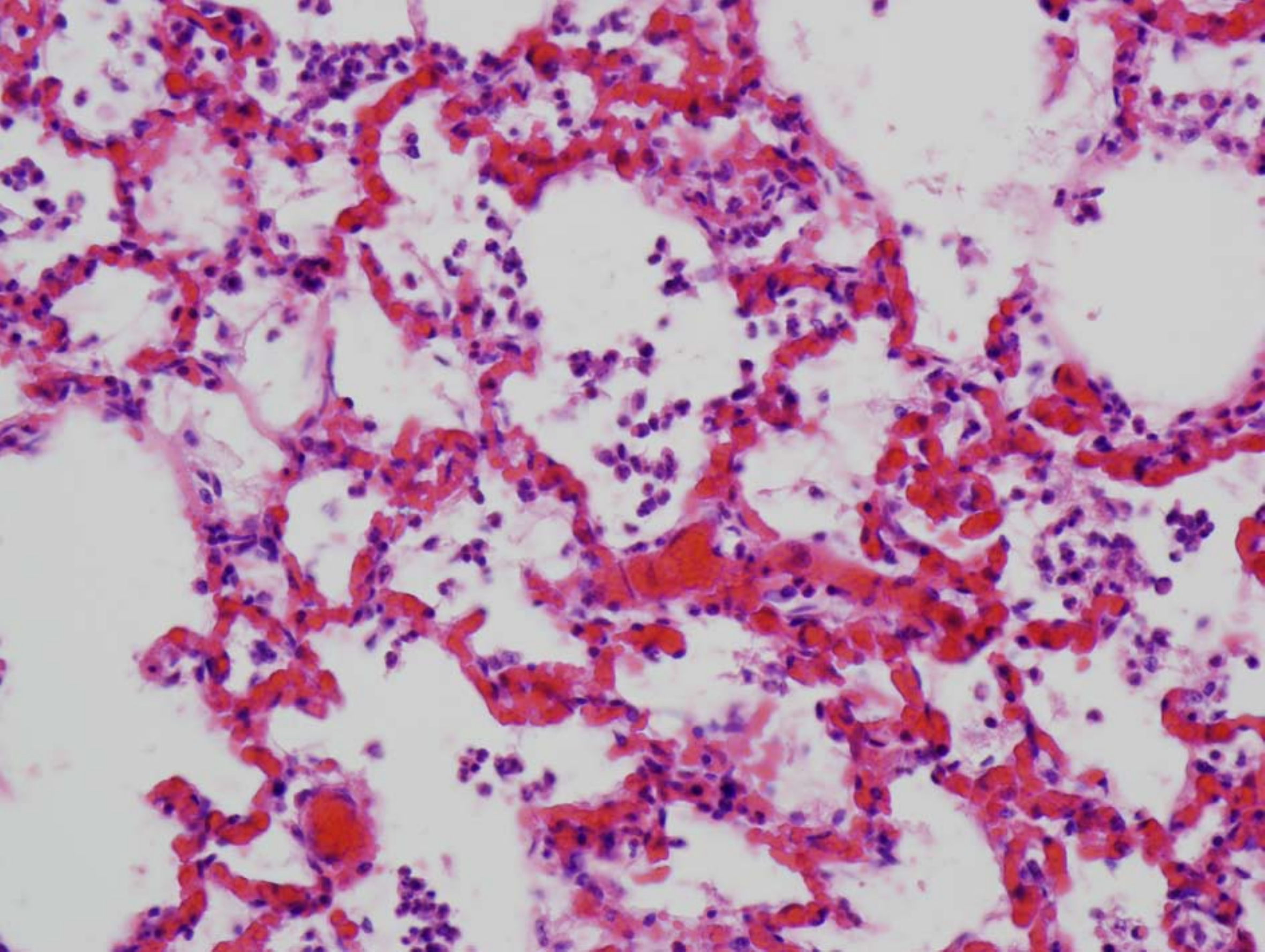}} \hspace{2mm}
	\subfigure[Inflamed lung.]{\includegraphics[scale=0.13]{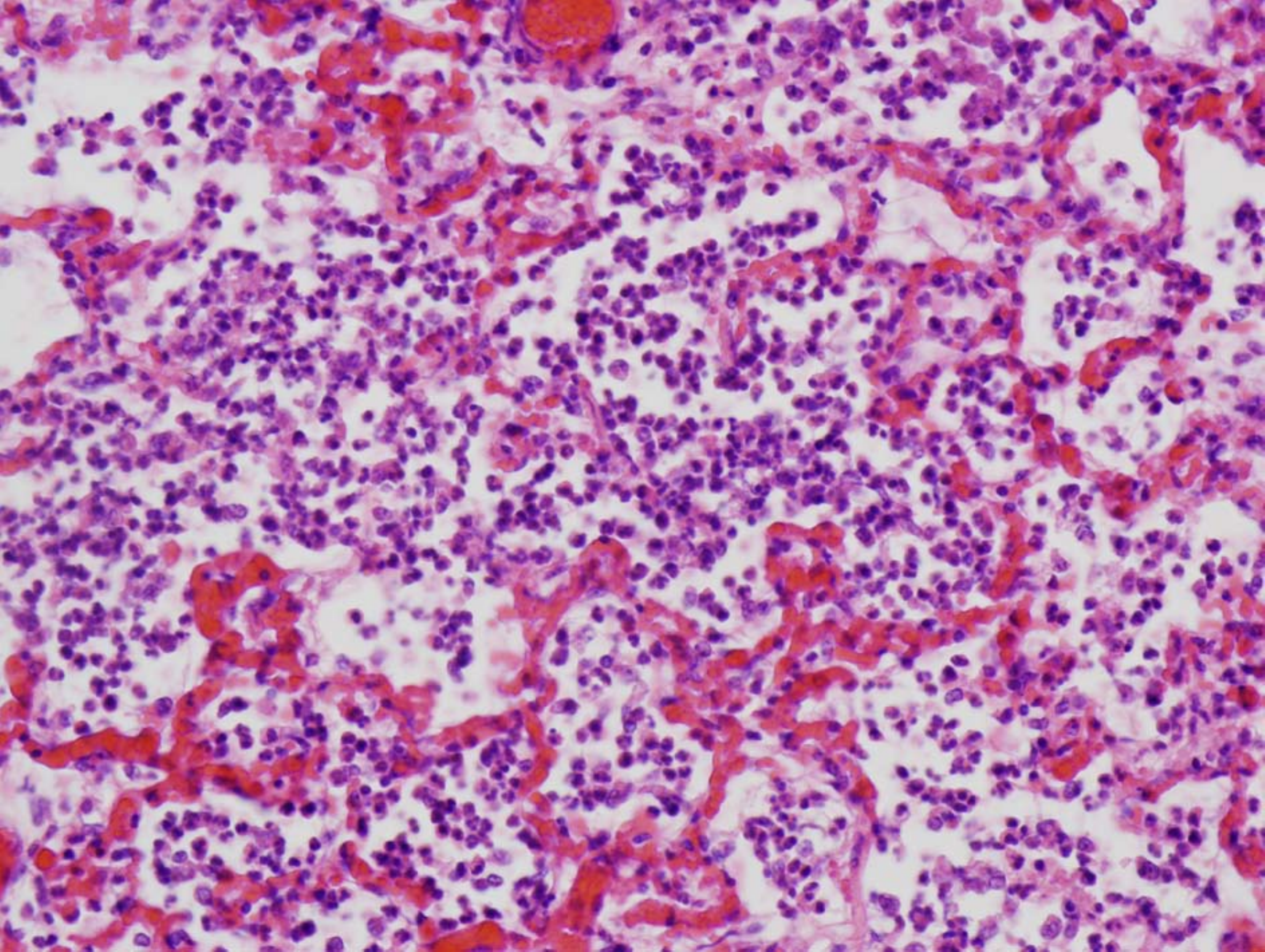}} \\
	\subfigure[Healthy kidney.]{\includegraphics[scale=0.13]{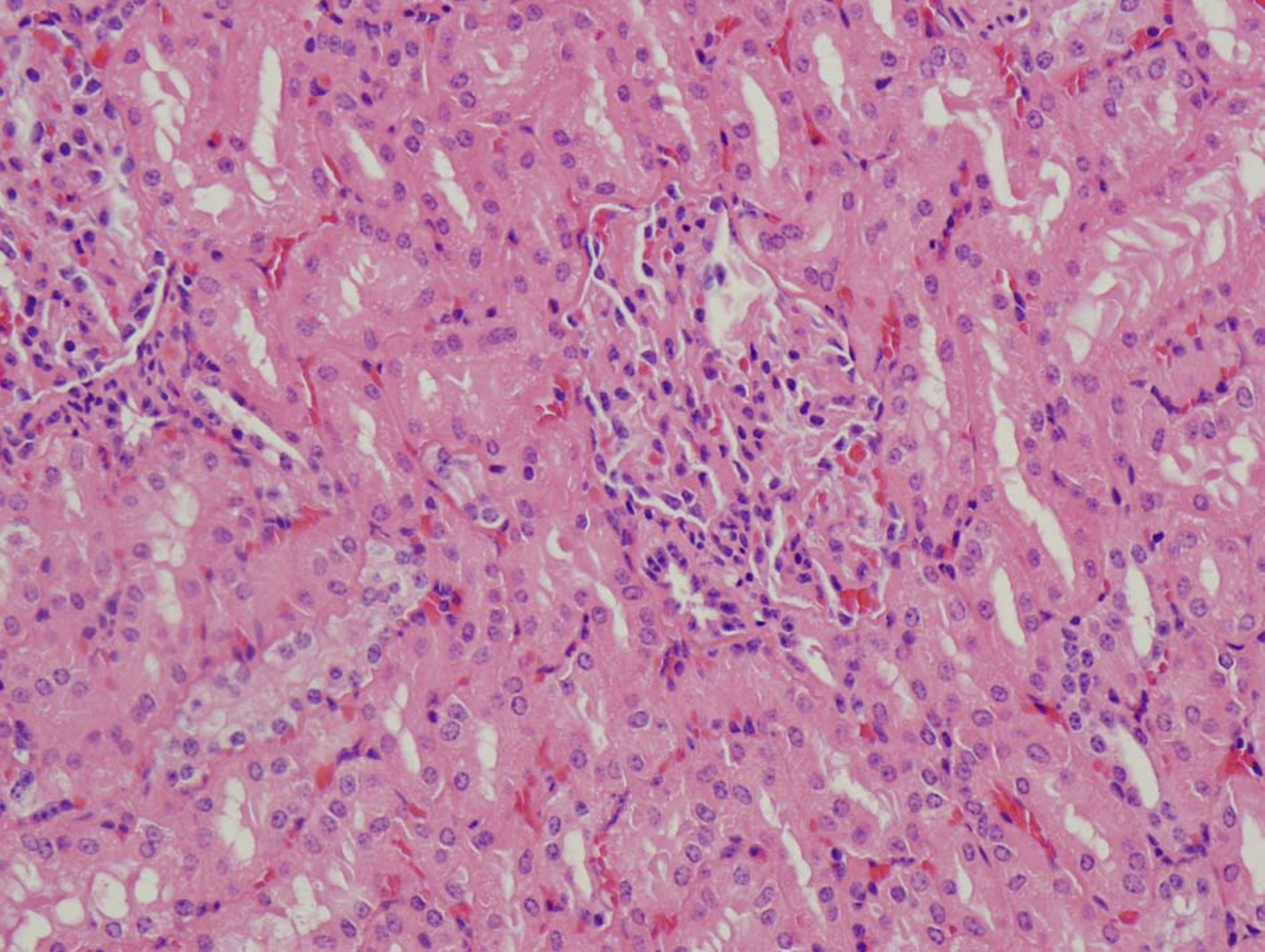}} \hspace{2mm}
	\subfigure[Healthy kidney.]{\includegraphics[scale=0.13]{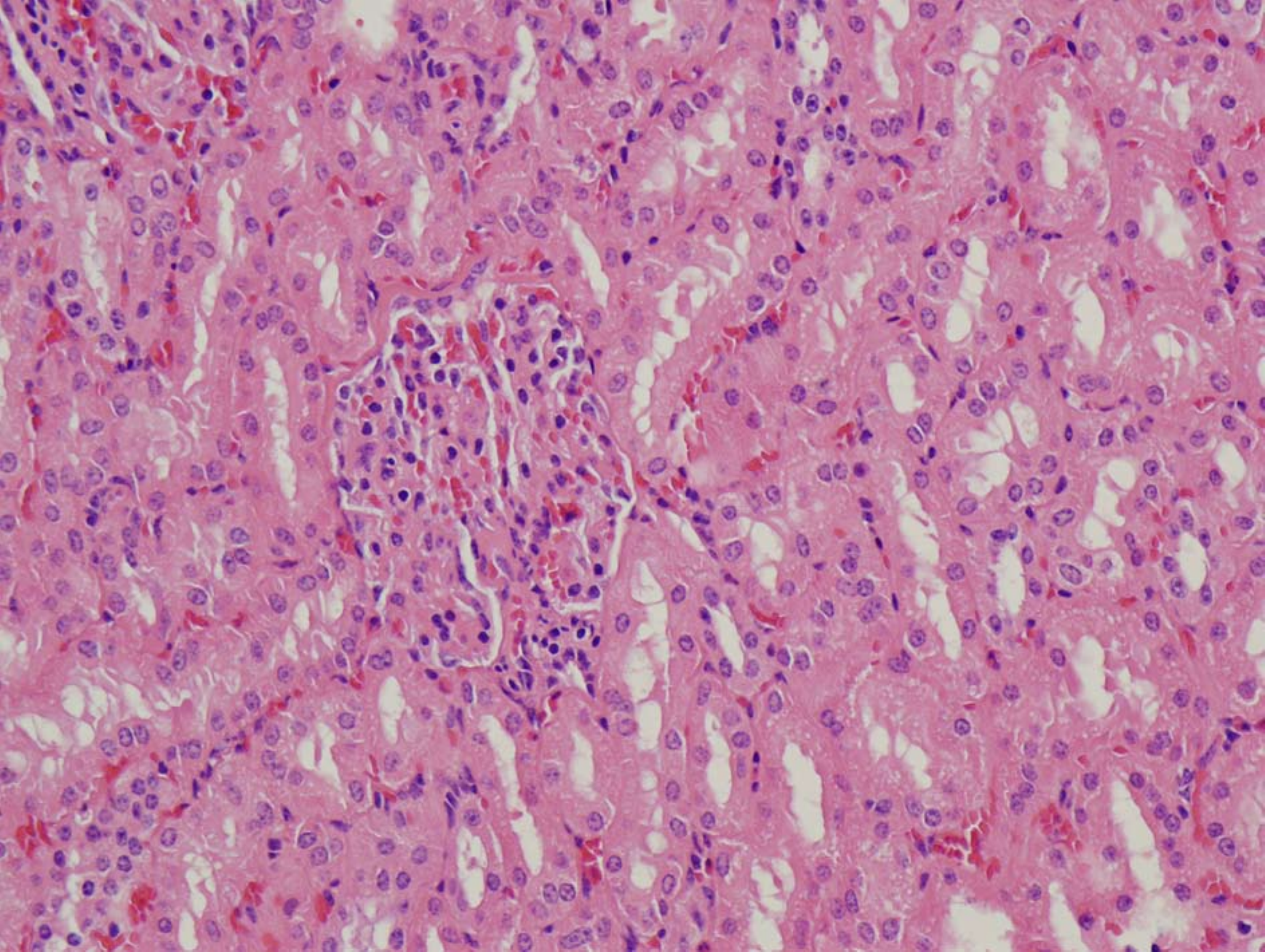}} \hspace{2mm}
	\subfigure[Inflamed kidney.]{\includegraphics[scale=0.13]{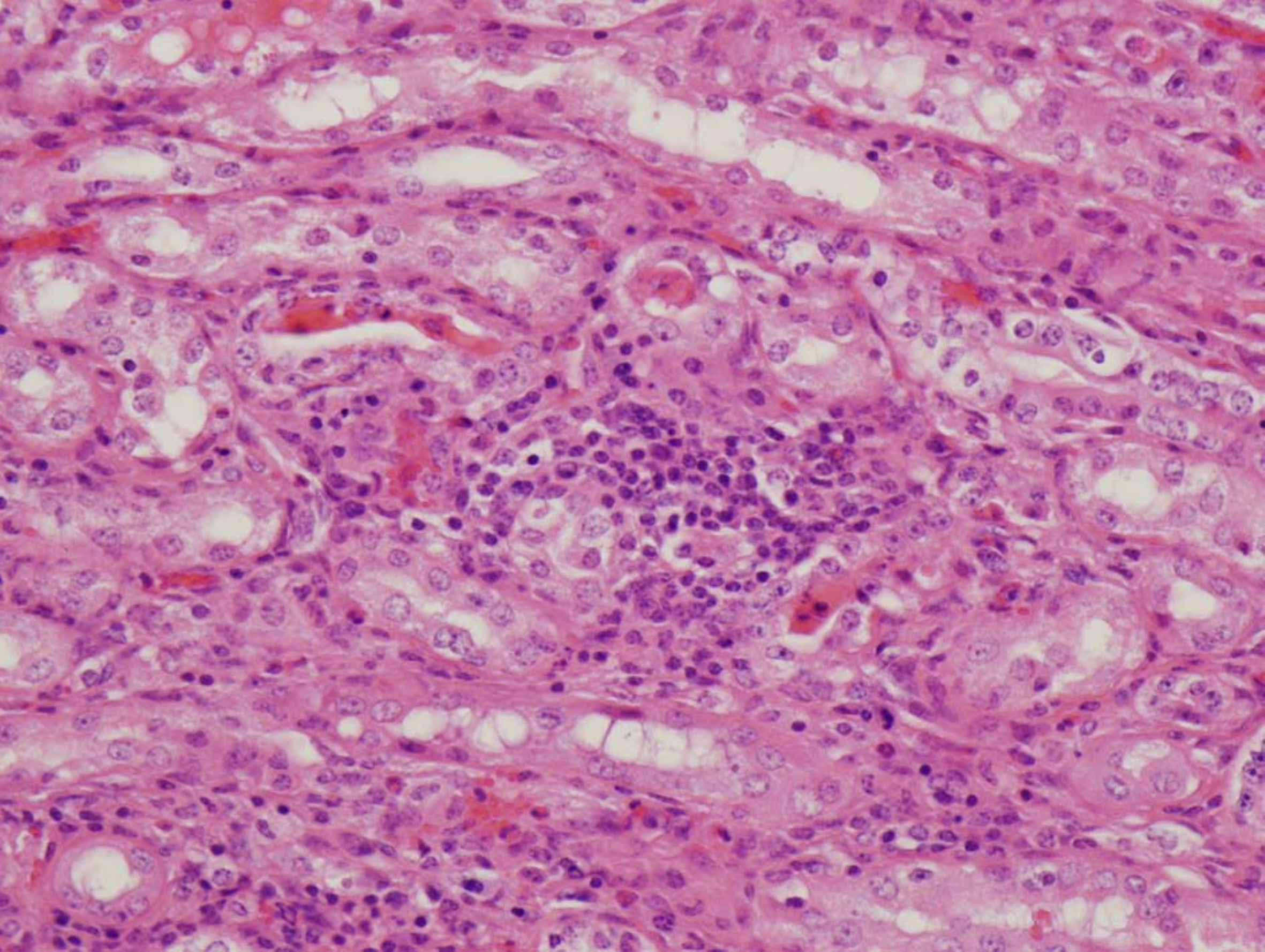}} \hspace{2mm}
	\subfigure[Inflamed kidney.]{\includegraphics[scale=0.13]{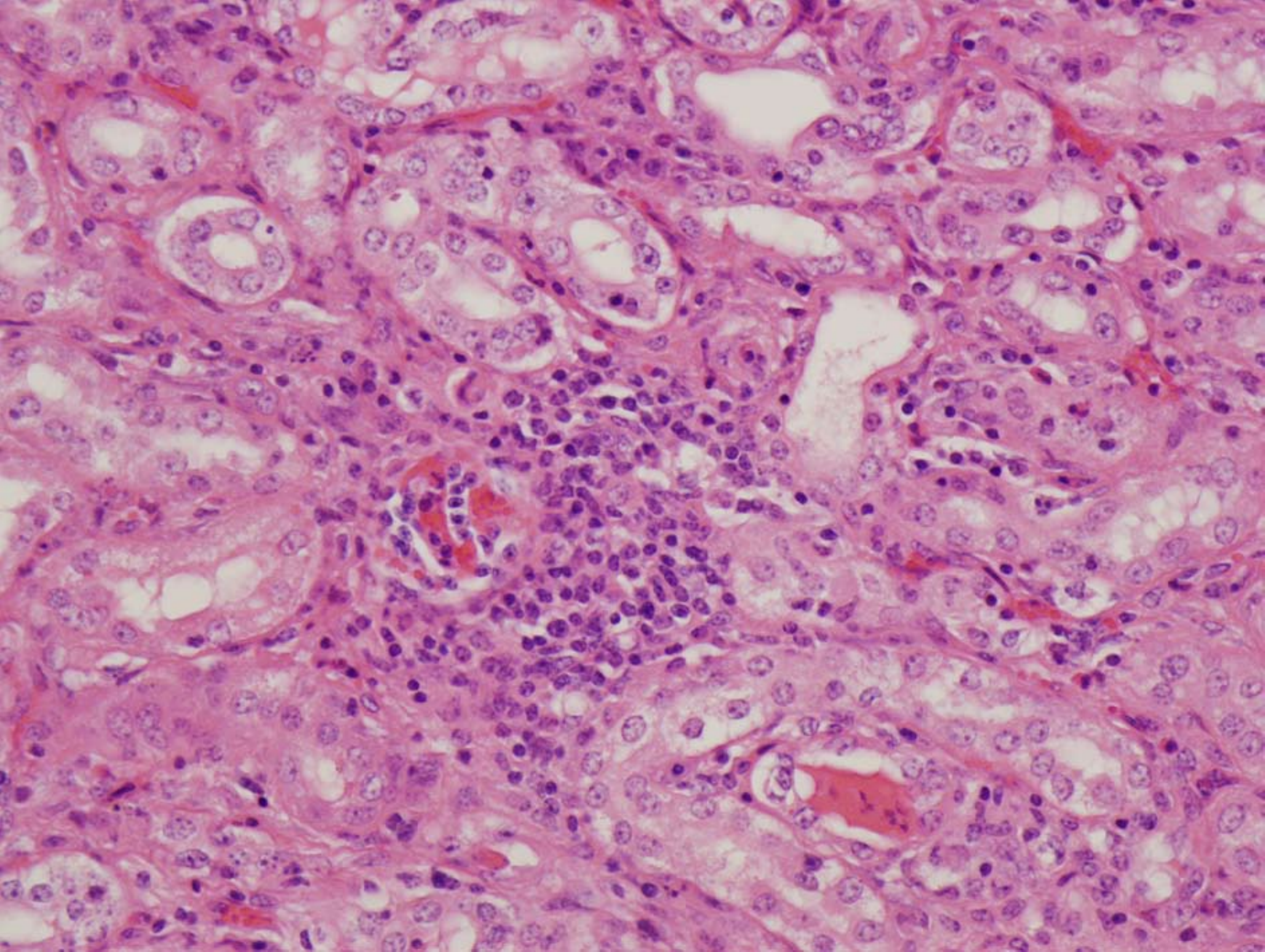}} \\
	\subfigure[Healthy spleen.]{\includegraphics[scale=0.13]{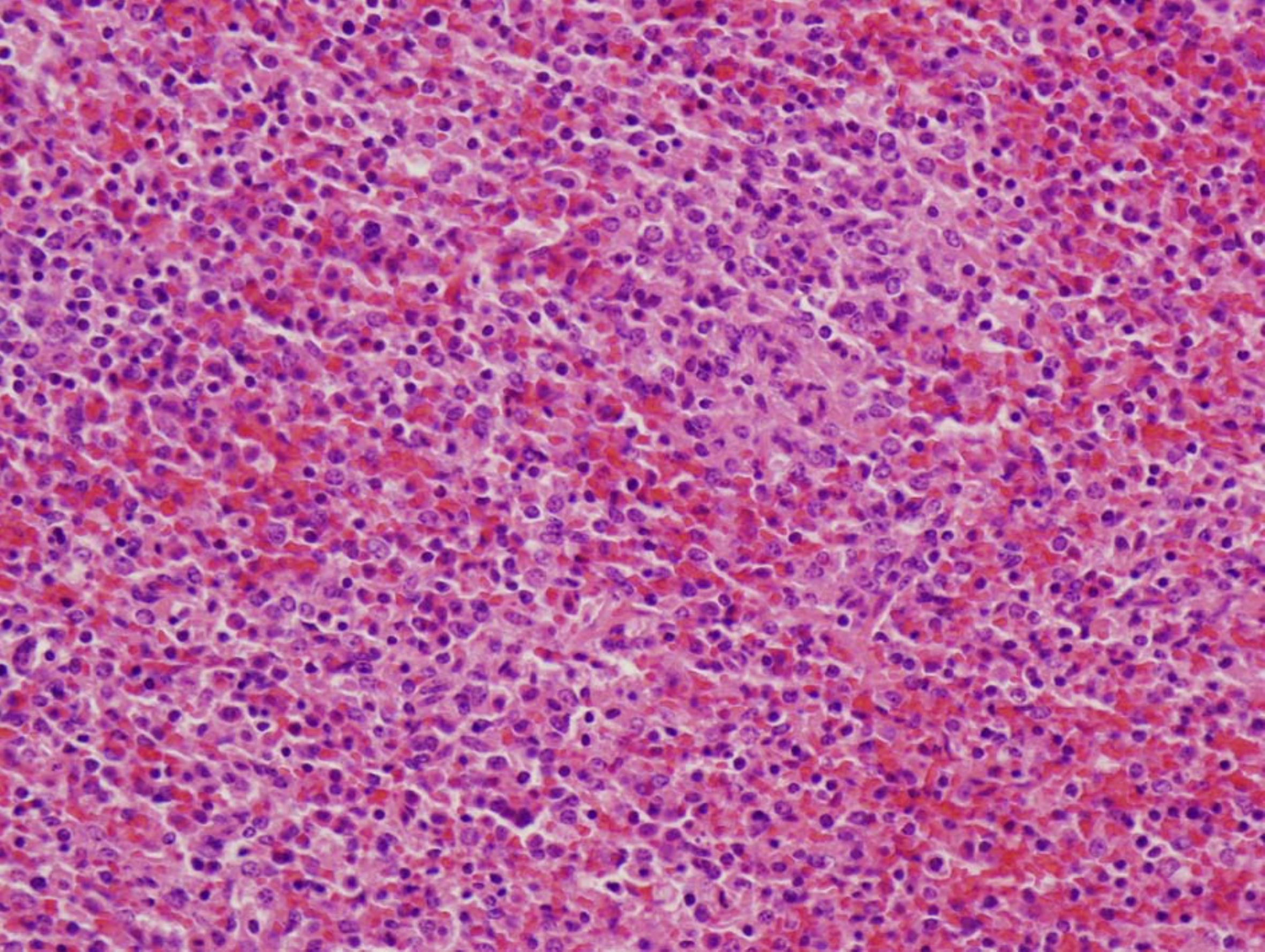}} \hspace{2mm}
	\subfigure[Healthy spleen.]{\includegraphics[scale=0.13]{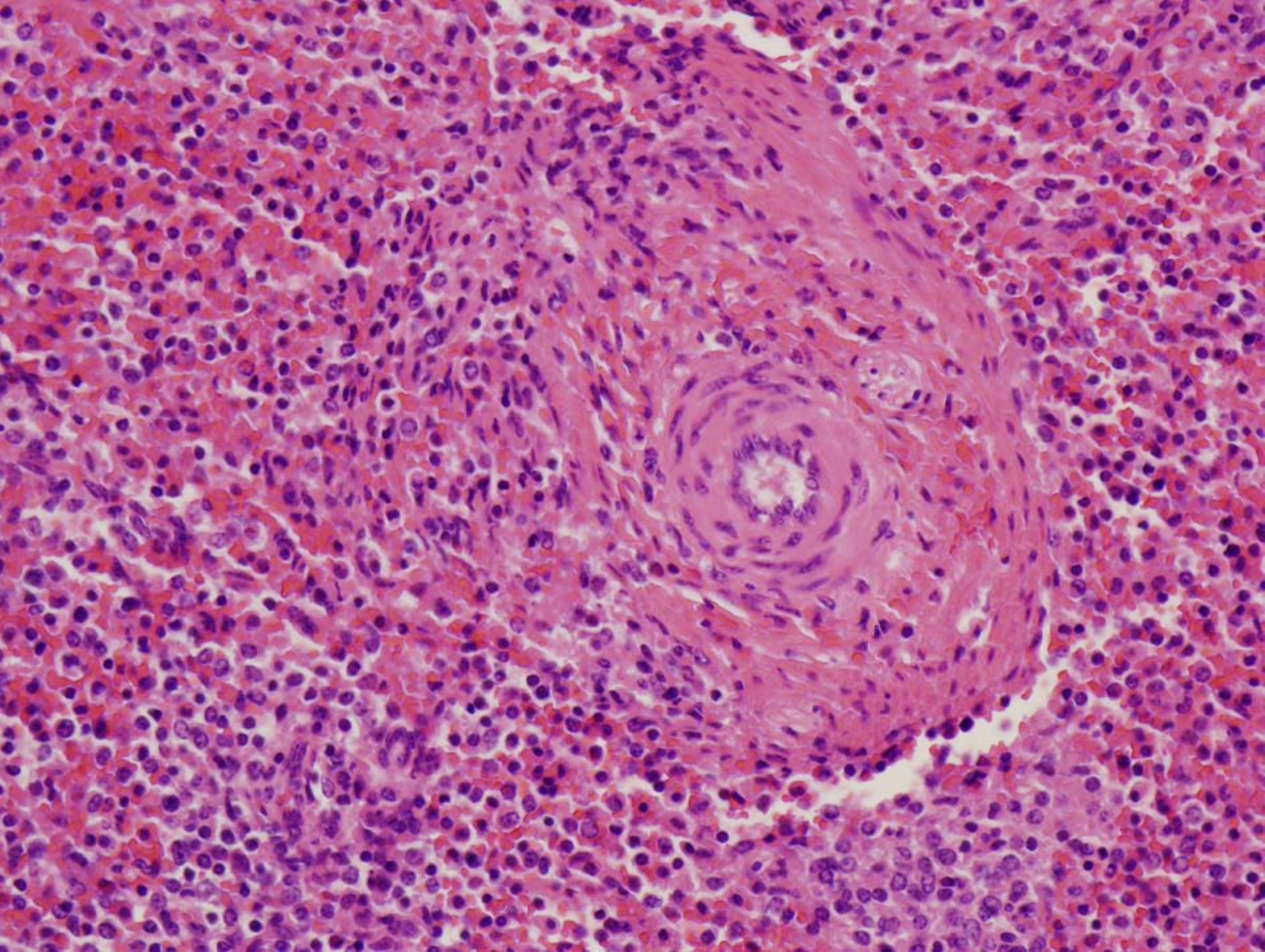}} \hspace{2mm}
	\subfigure[Inflamed spleen.]{\includegraphics[scale=0.13]{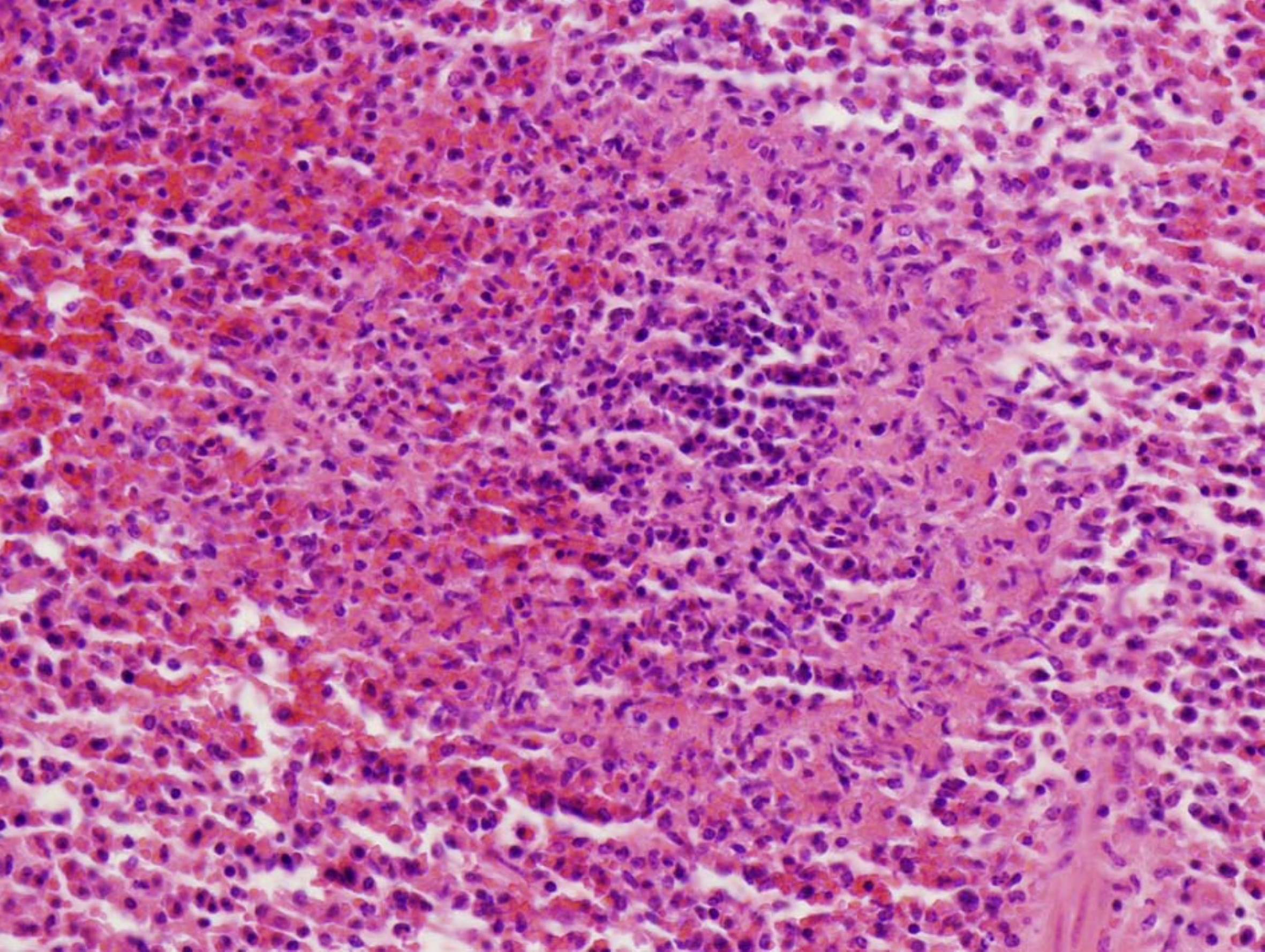}} \hspace{2mm}
	\subfigure[Inflamed spleen.]{\includegraphics[scale=0.13]{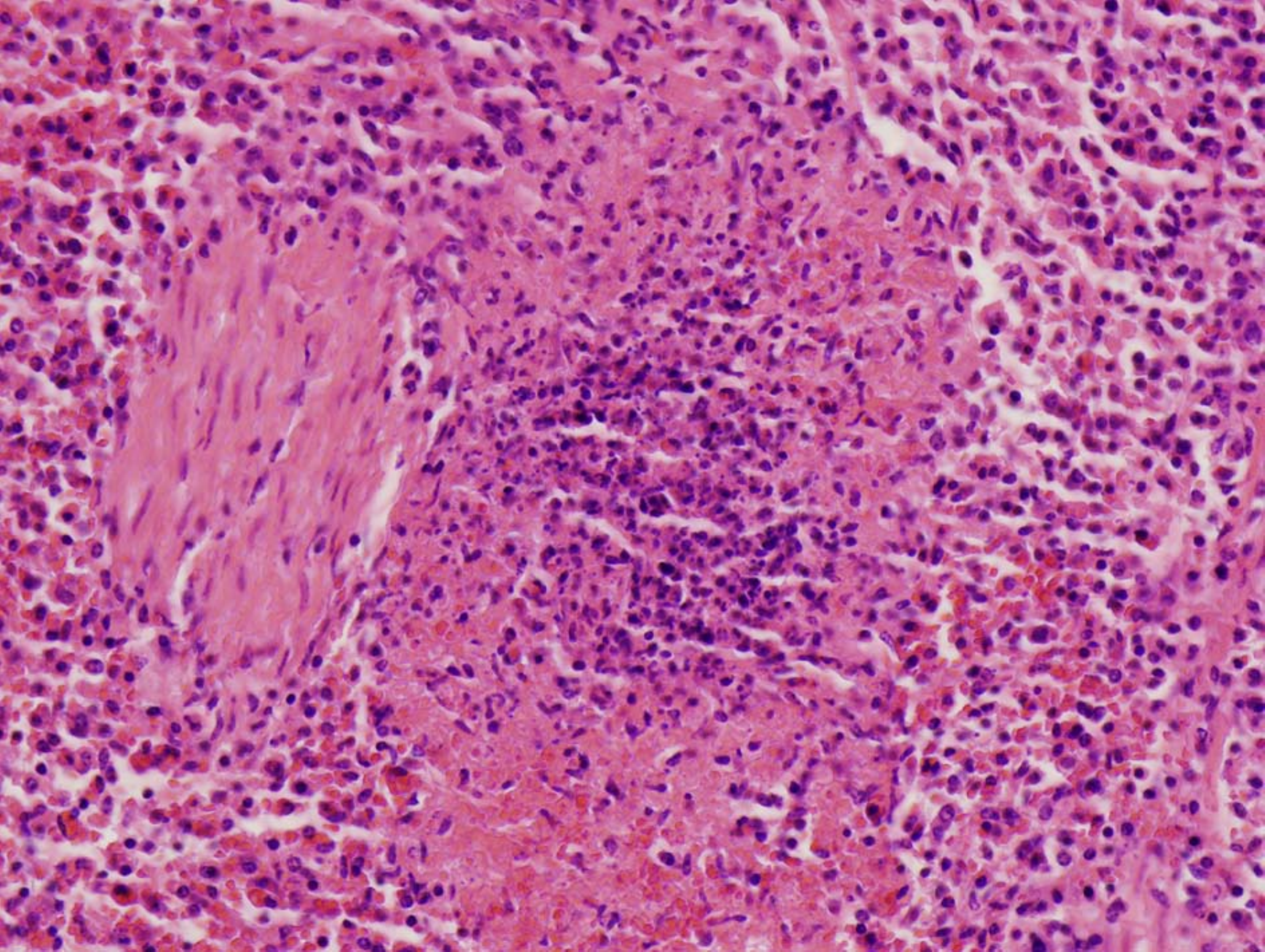}}
  \caption[Sample images from the ADL data set]{Sample images from the ADL data set. Each row corresponds to tissues from one mammalian organ.}
  \label{fig:hist_adl}
\end{figure}

\begin{figure}[t]
  \centering
  \subfigure{\includegraphics[scale=0.8]{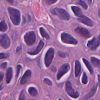}}
	\subfigure{\includegraphics[scale=0.8]{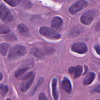}}
	\subfigure{\includegraphics[scale=0.8]{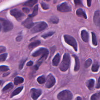}}
	\subfigure{\includegraphics[scale=0.8]{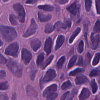}}
	\subfigure{\includegraphics[scale=0.8]{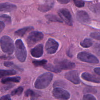}}
	\vspace{3mm}
	\subfigure{\includegraphics[scale=0.8]{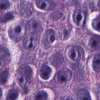}}
	\subfigure{\includegraphics[scale=0.8]{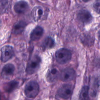}} 	
	\subfigure{\includegraphics[scale=0.8]{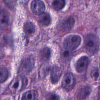}} 	
	\subfigure{\includegraphics[scale=0.8]{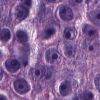}} 	
	\subfigure{\includegraphics[scale=0.8]{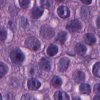}}
	\caption[Sample breast lesion images from the IBL data set]{Sample breast lesion images from the IBL data set. Top row: healthy (UDH) lesions, bottom row: cancerous (DCIS) lesions.}
  \label{fig:hist_ibl}
\end{figure}

\noindent{\textbf{ADL data set:}} These images are provided by pathologists at the Animal Diagnostics Lab, Pennsylvania State University. The tissue images have been acquired from three different mammalian organs - kidney, lung, and spleen. For each organ, images belonging to two categories - healthy or inflammatory - are provided. The H$\&$E-stained tissues are scanned using a whole slide digital scanner at 40x optical magnification, to obtain digital images of pixel size $4000 \times 3000$. All images are downsampled to $100 \times 75$ pixels in an aliasing-free manner for the purpose of computational speed-up. The algorithm in fact works at any image resolution. Example images\footnote{While the entire data set cannot be made publicly available, sample full-resolution images can be viewed at: \url{http://signal.ee.psu.edu/histimg.html}.} are shown in Fig. \ref{fig:hist_adl}. There are a total of 120 images for each organ, of which 40 images are used for training and the rest for testing. The ground truth labels for healthy and inflammatory tissue are assigned following manual detection and segmentation performed by ADL pathologists. We present classification results separately for each organ.

It is worthwhile to briefly understand the biological mechanisms underlying the different conditions in these images. Inflammatory cell tissue in cattle is often a sign of a contagious disease, and its cause and duration are indicated by the presence of specific types of white blood cells. Inflammation due to allergic reactions, bacteria, or parasites is indicated by the presence of eosinophils. Acute infections are identified by the presence of neutrophils, while macrophages and lymphocytes indicate a chronic infection. In Fig. \ref{fig:hist_adl}, we observe that a healthy lung is characterized by large clear openings of the alveoli, while in the inflamed lung, the alveoli are filled with bluish-purple inflammatory cells. Similar clusters of dark blue nuclei indicate the onset of inflammation in the other organs.\\

\noindent{\textbf{IBL data set:}} The second data set comprises images of human intraductal breast lesions \cite{dundar:tbe11}. It has been provided by the Clarian Pathology Lab and Computer and Information Science Dept., Indiana University-Purdue University Indianapolis. The images belong to either of two well-defined categories: usual ductal hyperplasia (UDH) and ductal carcinoma in situ (DCIS). UDH is considered benign and the patients are advised follow-up check-ups, while DCIS is actionable and the patients require surgical intervention. Ground truth class labels for the images are assigned manually by the pathologists. A total of 40 patient cases - 20 well-defined DCIS and 20 UDH - are identified for experiments in the manner described in \cite{dundar:tbe11}. Each case contains a number of Regions of Interest (RoIs), and we have chosen a total of 120 images (RoIs), consisting of a randomly selected set of 60 images for training and the remaining 60 RoIs for test. Each RoI represents a full-size image for our experiments. Smaller local regions are chosen carefully within each such RoI for LA-SHIRC as described in \ref{sec:local}, using a classical morphology-based blob detection technique \cite{serra:book82}.

We compare the performance of SHIRC and LA-SHIRC with two competing approaches:
\begin{enumerate}
  \item \emph{SVM:} this method combines state-of-the-art feature extraction and classification. We use the collection of features from WND-CHARM \cite{orlov:prl08,shamir:scbm08} which is known to be a powerful toolkit of features for medical images. A support vector machine is used for decisions unlike weighted nearest neighbor in \cite{orlov:prl08} to further enhance classification. We pick the most relevant features for histopathology \cite{gurcan:rbe09}, including but not limited to (color channel-wise) histogram information, image statistics, morphological features and wavelet coefficients from each color channel. The source code for WND-CHARM is made available by the National Institutes of Health online at: \texttt{http://ome.grc.nia.nih.gov/wnd-charm/}.
  \item \emph{SRC:} the single-channel sparse representation-based classification approach reviewed in Section \ref{secch3:src}. Specifically, we employ SRC directly on the luminance channel (obtained as a function of the RGB channels) of the histopathological images, as proposed initially for face recognition and applied widely thereafter.
\end{enumerate}
For results on the IBL data set, we also directly report corresponding numbers from \cite{dundar:tbe11} - the multiple-instance learning (MIL) algorithm - which is a full image analysis and classification system customized for the IBL data set.

In supervised classification, it is likely that some particularly well-chosen training sets can lead to high classification accuracy. In order to mitigate this issue of \emph{selection bias}, we perform 10 different trials of each experiment. In each trial, we randomly select a set of training images -- all results reported are the average of the classification accuracies from the individual trials.

\subsection{Validation of Central Idea: Overall Classification Accuracy}
\label{sec:validation}

\begin{figure}[t]
  \centering
  \subfigure[ADL data set.]{\includegraphics[scale=0.32]{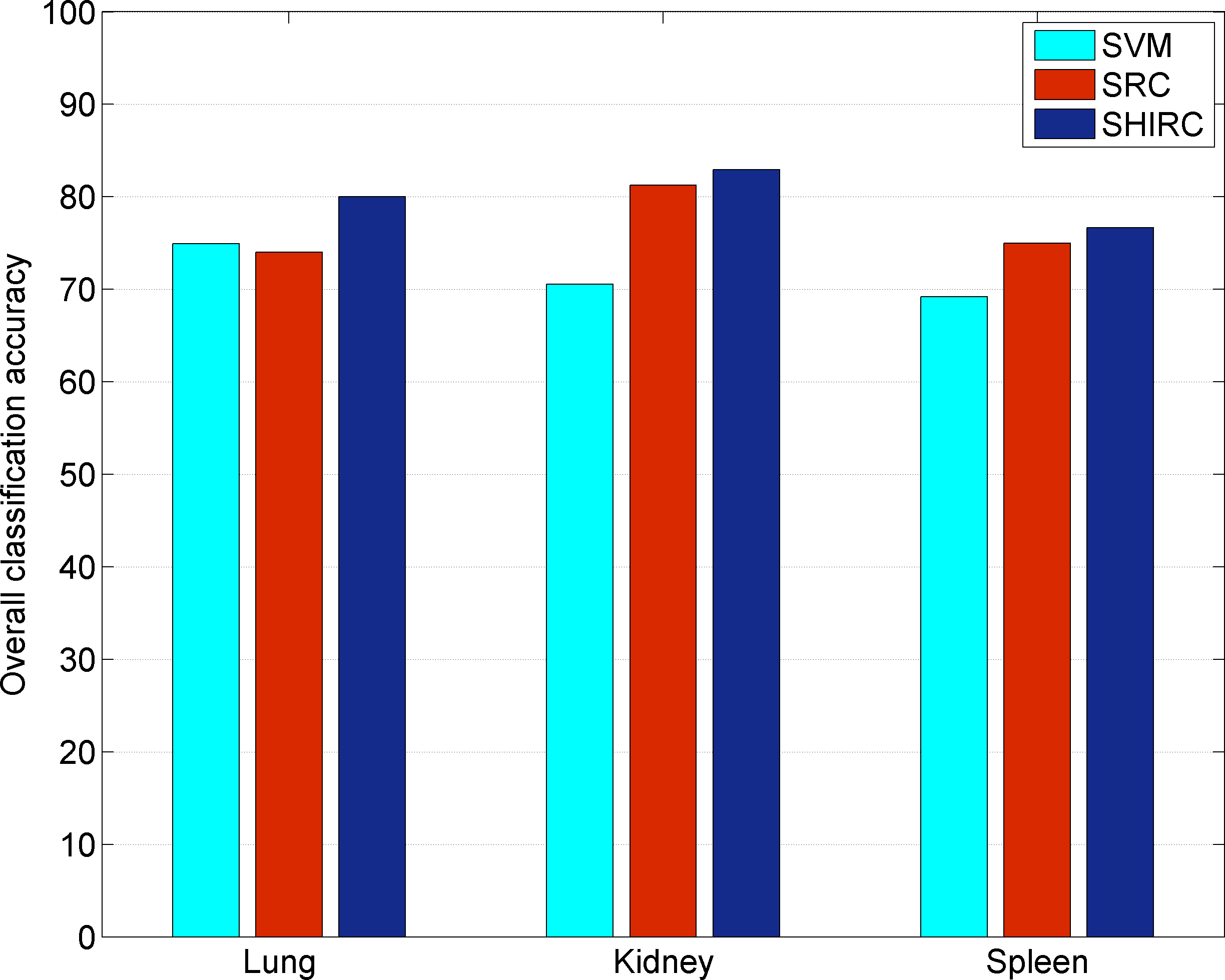}} \hspace{10mm}
  \subfigure[IBL data set.]{\includegraphics[scale=0.32]{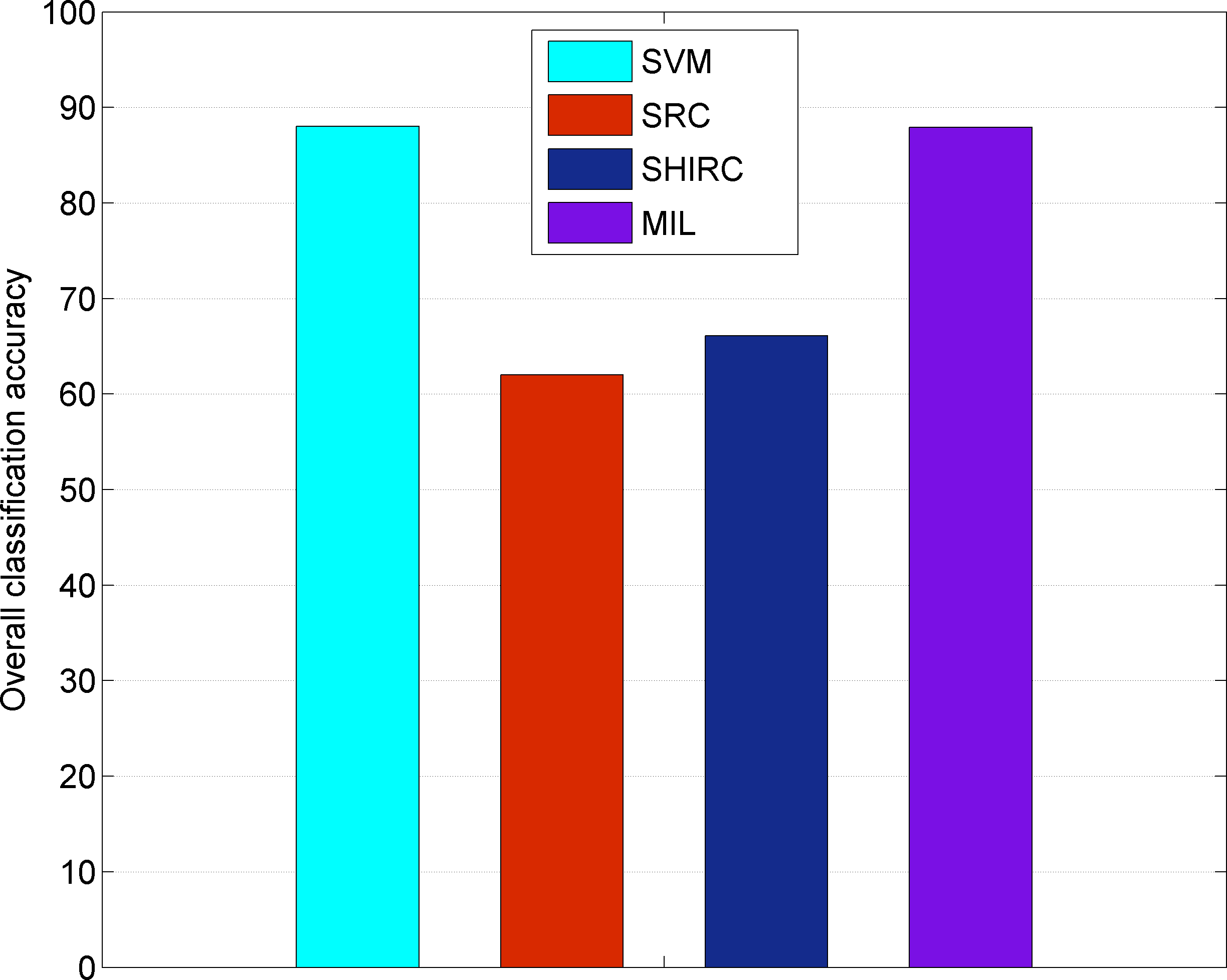}}
	\caption[Bar graphs indicating the overall classification accuracies of the competing methods]{Bar graphs indicating the overall classification accuracies of the competing methods.}
  \label{fig:bargraph}
\end{figure}

\begin{table}[t]
  \centering
  \caption[Confusion matrix: Lung]{Confusion matrix: Lung.}
  \label{tab:conf_lung}
  \begin{tabular}{|c|c|c||c|}
  \hline
  Class & Healthy & Inflammatory & Method \\
  \hline
  Healthy & \textbf{0.8875} & 0.1125 & SVM \\
  & 0.7250 & 0.2750 & SRC \\
  & 0.7500 & 0.2500 & SHIRC \\
  \hline
  Inflammatory & 0.3762 & 0.6238 &  SVM \\
  & 0.2417 & 0.7583 &  SRC \\
  & 0.1500 & \textbf{0.8500} & SHIRC \\
  \hline
  \end{tabular}
\end{table}

\begin{table}[t]
  \centering
  \caption[Confusion matrix: Kidney]{Confusion matrix: Kidney.}
  \label{tab:conf_kidney}
  \begin{tabular}{|c|c|c||c|}
  \hline
  Class & Healthy & Inflammatory & Method \\
  \hline
  Healthy & 0.6925 & 0.3075 & SVM \\
  & \textbf{0.8750} & 0.1250 & SRC \\
  & 0.8250 & 0.1750 & SHIRC \\
  \hline
  Inflammatory & 0.2812 & 0.7188 & SVM \\
  & 0.2500 & 0.7500 &  SRC \\
  & 0.1667 & \textbf{0.8333} & SHIRC \\
  \hline
  \end{tabular}
\end{table}

\begin{table}[t]
  \centering
  \caption[Confusion matrix: Spleen]{Confusion matrix: Spleen.}
  \label{tab:conf_spleen}
  \begin{tabular}{|c|c|c||c|}
  \hline
  Class & Healthy & Inflammatory & Method \\
  \hline
  Healthy & 0.5112 & 0.4888 & SVM \\
  &\textbf{0.7083}  &0.2917  & SRC \\
  &0.6500  &0.3500 & SHIRC \\
  \hline
  Inflammatory & 0.1275 & 0.8725 & SVM \\
  &0.2083  &0.7917  &  SRC \\
  &0.1167  & \textbf{0.8833}  & SHIRC \\
  \hline
  \end{tabular}
\end{table}

\begin{table}[t]
  \centering
  \caption[Confusion matrix: Intraductal breast lesions]{Confusion matrix: Intraductal breast lesions.}
  \label{tab:conf_breast}
  \begin{tabular}{|c|c|c||c|}
  \hline
  Class & UDH & DCIS & Method \\
  \hline
  UDH & 0.8636 & 0.1364 & SVM \\
  & 0.6800 & 0.3200 & SRC \\
  & 0.6818 & 0.3182 & SHIRC \\
  & \textbf{0.9333} & 0.0667 & LA-SHIRC \\
  \hline
  DCIS & 0.0909 & 0.9091 & SVM \\
  & 0.4400 & 0.5600 & SRC \\
  & 0.3600 & 0.6400 & SHIRC \\
  & 0.1000 & \textbf{0.9000} & LA-SHIRC \\
  \hline
  \end{tabular}
\end{table}

\begin{figure}[t]
  \centering
  \subfigure[Lung (ADL).]{\includegraphics[scale=0.32]{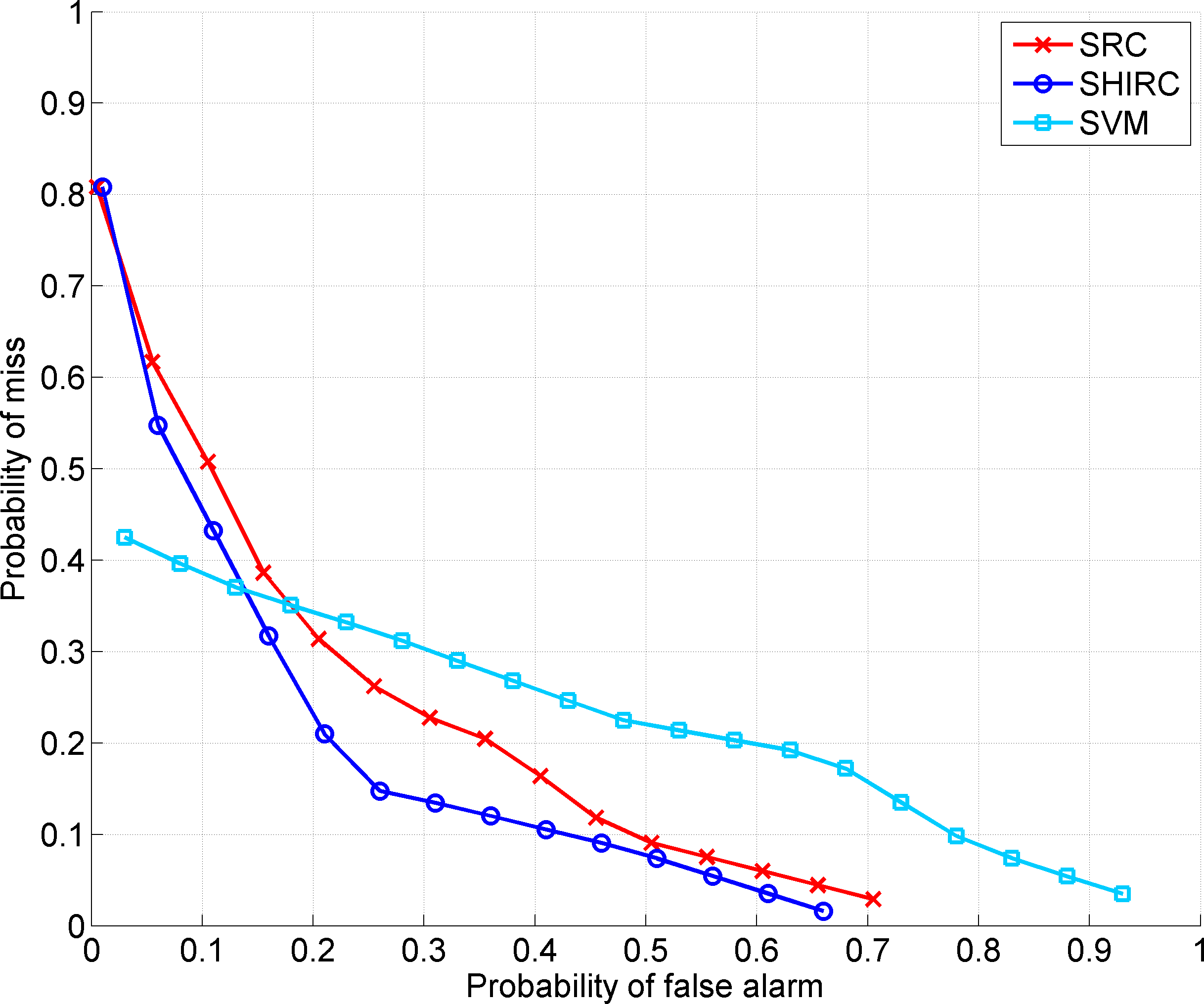}} \hspace{5mm}
  \subfigure[Kidney (ADL).]{\includegraphics[scale=0.30]{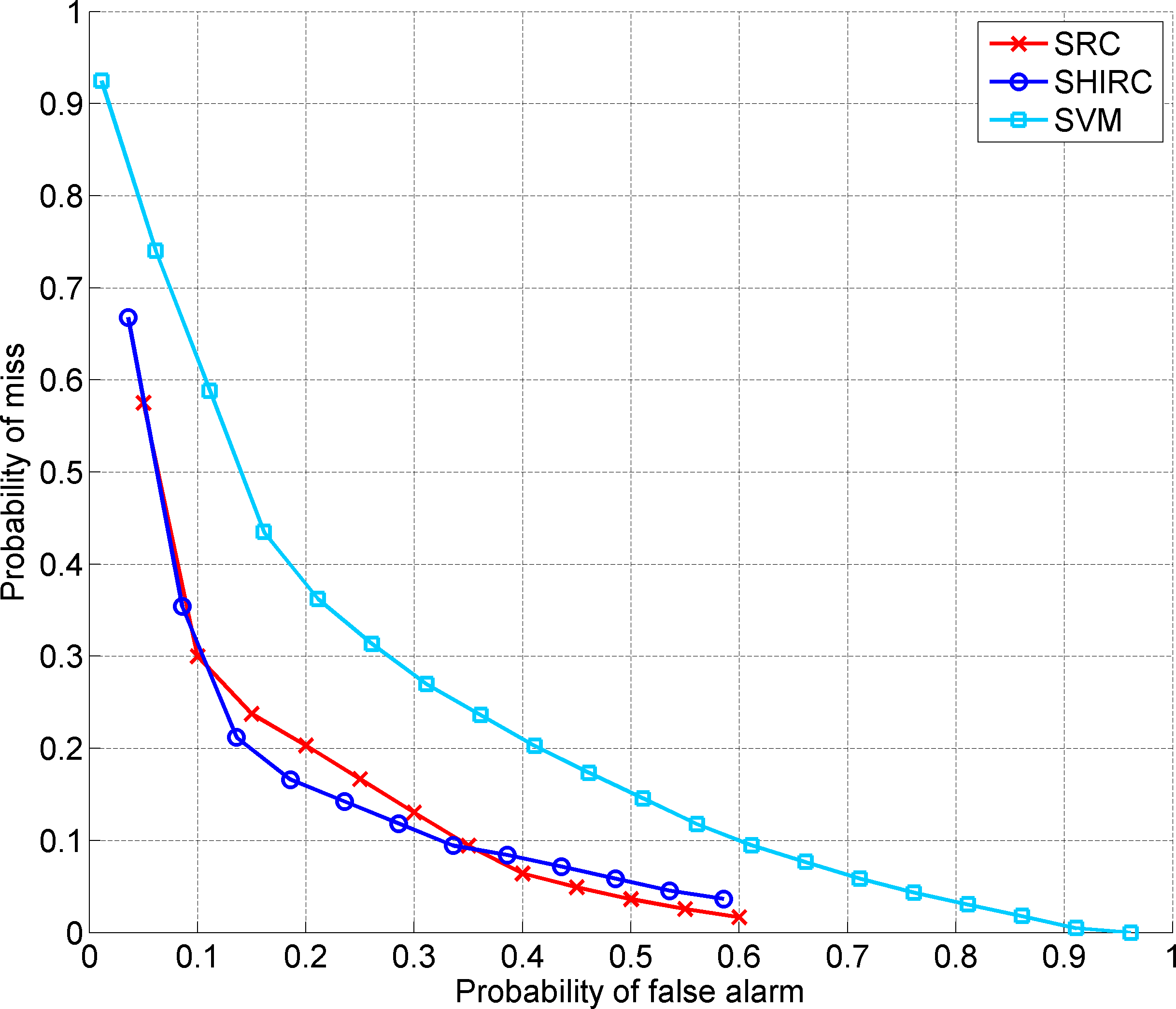}}\\
  \subfigure[Spleen (ADL).]{\includegraphics[scale=0.32]{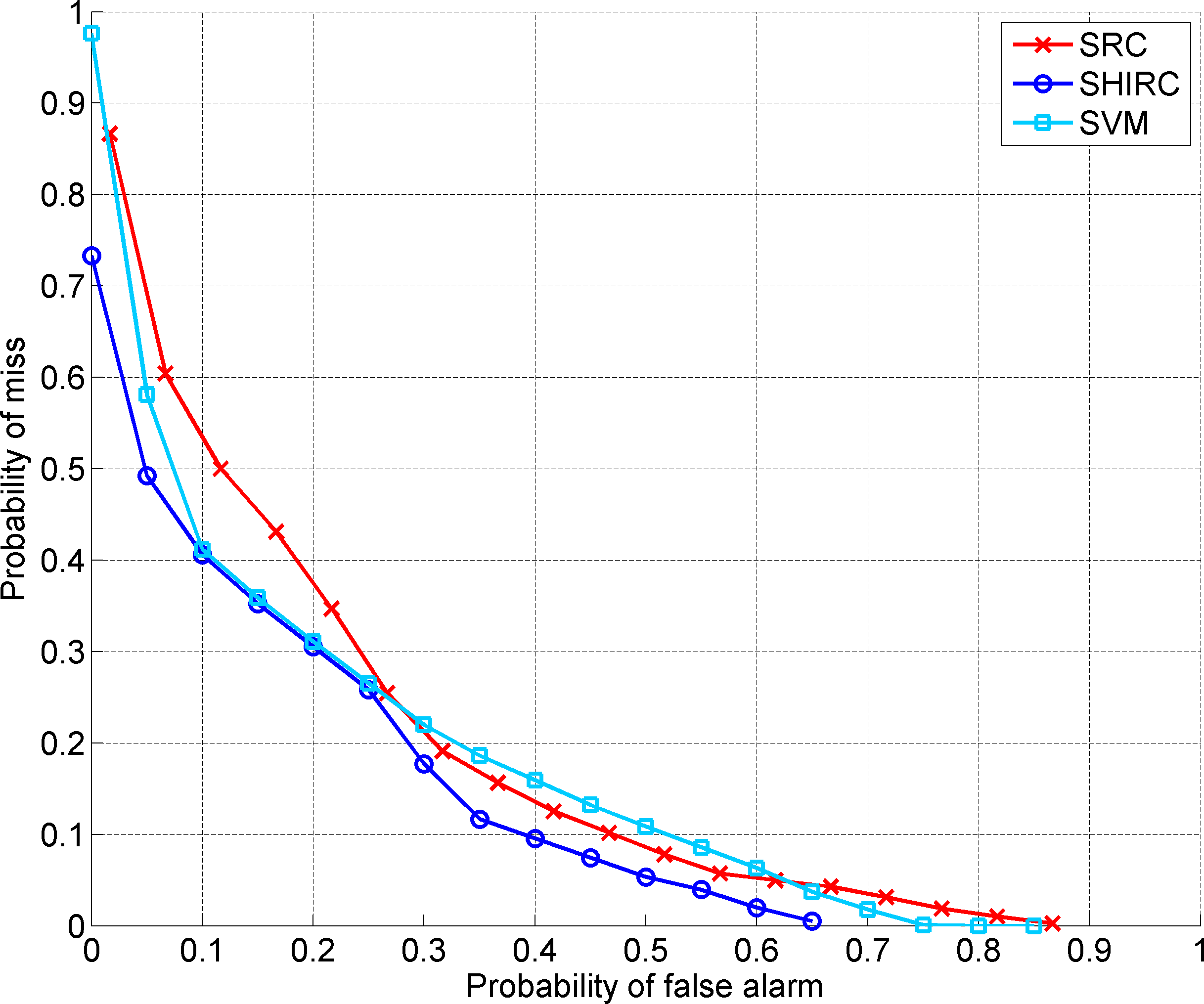}} \hspace{5mm}
  \subfigure[IBL.]{\includegraphics[scale=0.30]{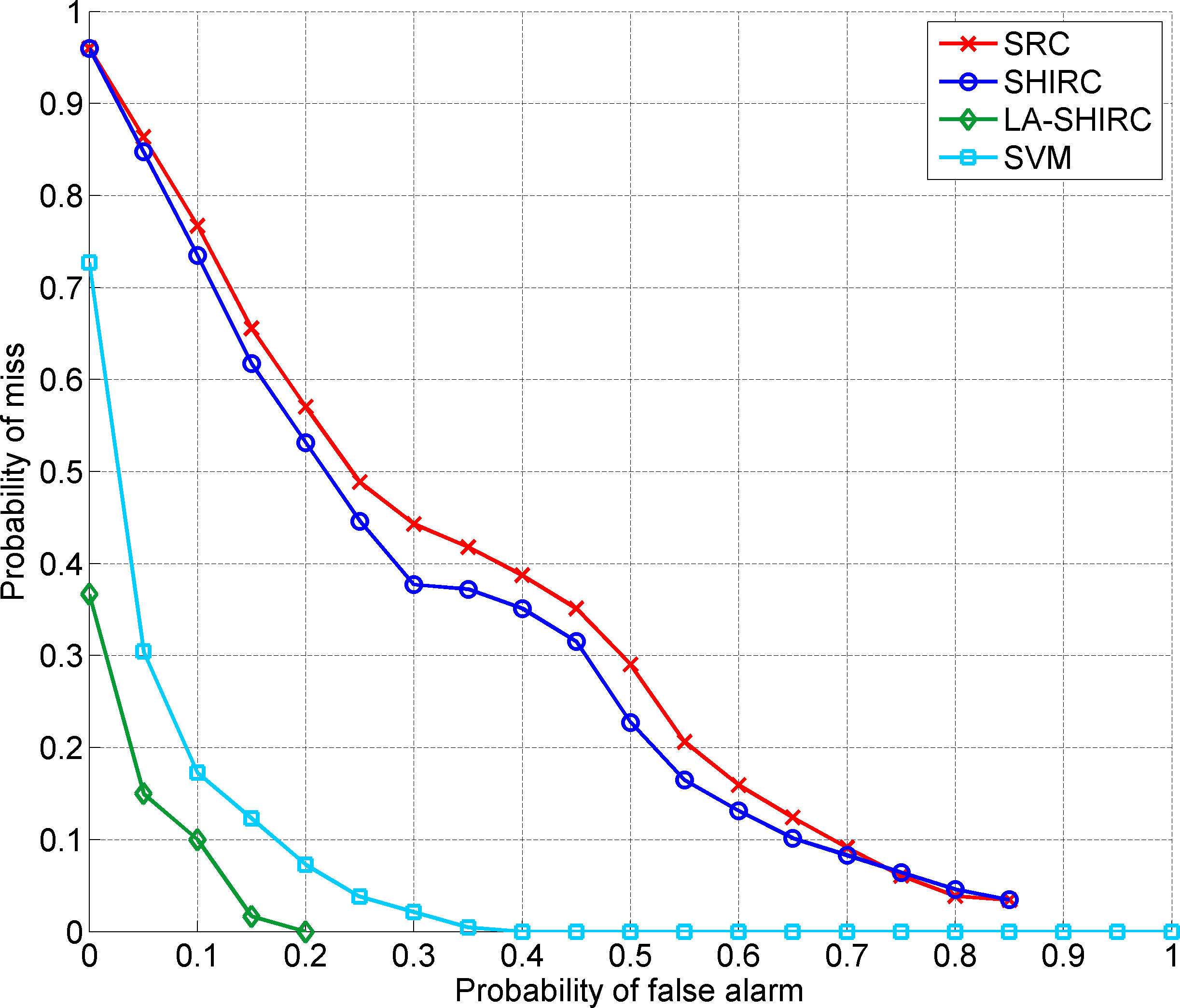}}
  \caption[Receiver operating characteristic curves for different organs]{Receiver operating characteristic (ROC) curves for different organs.}
  \label{fig:roc}
\end{figure}

\begin{table}[t]
  \centering
  \caption[False alarm probability for fixed detection rate]{False alarm probability for fixed detection rate.}
  \label{tab:fixed_det}
  \begin{tabular}{|c|c|c|c|c|c|}
		\hline
		\multirow{2}{*}{Images} & Fixed rate & \multicolumn{4}{c}{False alarm rate} \\
		\cline{3-6}
		& of detection & SVM & SRC & SHIRC & LA-SHIRC \\
    \hline
		Lung (ADL) & 0.15 & 0.71 & 0.42 & 0.26 & 0.26\\
		Kidney (ADL) & 0.15 & 0.50 & 0.27 & 0.22 & 0.22 \\
		Spleen (ADL) & 0.15 & 0.45 & 0.40 & 0.33 & 0.33 \\
		IBL & 0.10 & 0.17 & 0.69 & 0.65 & 0.10\\
		\hline
	\end{tabular}
\end{table}

First, we provide experimental validation of our central hypothesis: that exploiting color information in a principled manner through simultaneous sparsity models leads to better classification performance over existing techniques for histopathological image classification. To this end, we present overall classification accuracy for the three organs from the ADL data set, in the form of bar graphs in Fig. \ref{fig:bargraph}(a). SHIRC outperforms SVM and SRC in each of the three organs, thereby confirming the merit of utilizing color correlation information. The selection of application-specific features coupled with the inclusion of features from the RGB channels ensures that the SVM classifier performs competitively, particularly for the lung.

A similar experiment using the full-size images from the IBL data set illustrates the variability in histopathological imagery. Each image in the data set contains multiple cells at different spatial locations, as seen in Fig. \ref{fig:hist_ibl}. SHIRC is not designed to handle this practical challenge. The bar graph in Fig. \ref{fig:bargraph}(b) shows that the SVM classifier and the systemic MIL approach in \cite{dundar:tbe11} offer the best classification accuracy. This is not surprising because MIL \cite{dundar:tbe11} incorporates elaborate segmentation and pre-processing followed by feature extraction strategies customized to the acquired set of images. This experimental scenario occurs frequently enough in practice and serves as our motivation to develop LA-SHIRC.

\subsection{Detailed Results: Confusion Matrices and ROC Curves}
\label{sec:roc}

Next, we present a more elaborate interpretation of classification performance in the form of confusion matrices and ROC curves. Each row of a confusion matrix refers to the actual class identity of test images and each column indicates the classifier output.

Tables \ref{tab:conf_lung}-\ref{tab:conf_spleen} show the mean confusion matrices for the ADL data set. In continuation of trends from Fig. \ref{fig:bargraph}(a), SHIRC offers the best disease detection accuracy - a quantitative metric of high relevance to pathologists - for each organ, while maintaining high classification accuracy for healthy images too. An interesting observation can be made from Table \ref{tab:conf_spleen}. The SVM classifier reveals a tendency to classify the diseased tissue images much more accurately than the healthy tissues. In other words, there is a high false alarm rate (healthy image mistakenly classified as inflammatory) associated with the SVM classifier. SHIRC however offers a more consistent class-specific performance, resulting in the best overall performance. The corresponding results using LA-SHIRC are identical to SHIRC and hence not shown, since a single block (i.e. the entire image) was deemed by pathologists to have sufficient discriminative information.

Table \ref{tab:conf_breast} shows the mean confusion matrix for the IBL data set. SHIRC provides an average classification accuracy of $66.09 \%$, in comparison with about $87.9 \%$ using the MIL approach \cite{dundar:tbe11}. However, LA-SHIRC results in a significant improvement in performance, even better than the rates reported using SVM, MIL or SRC. For LA-SHIRC, we identify 9 local objects per image corresponding to individual cells. It is noteworthy that a pre-processing stage involving careful image segmentation is performed prior to feature extraction in MIL \cite{gurcan:rbe09}, implying that MIL is representative of state-of-the-art classification techniques using local image information.

Typically in medical image classification problems, pathologists desire algorithms that reduce the probability of miss (classifying diseased image as healthy) while also ensuring that the false alarm rate remains low. However, there is a trade-off between these two quantities, conveniently described using a receiver operating characteristic (ROC) curve. Fig. \ref{fig:roc} shows the ROC curves for the ADL and IBL data sets. The lowest curve (closest to the origin) has the best overall performance and the optimal operating point minimizes the sum of the miss and false alarm probabilities. In Figs. \ref{fig:roc}(a)-(c), the curves for LA-SHIRC are not shown since they are identical to SHIRC (only one global image block is sufficient). In each case, SHIRC offers the best trade-off. In Fig. \ref{fig:roc}(d), the LA-SHIRC outperforms SVM, and both methods are much better than SRC and SHIRC\footnote{Note that ROCs for MIL \cite{dundar:tbe11} could not be reported because the image analysis and classification system in \cite{dundar:tbe11} has a variety of pre-processing, segmentation and other image processing and classification steps which makes exact reproduction impossible in the absence of publicly available code.}.

\begin{figure}[t]
  \centering
  \subfigure[Kidney (ADL).]{\includegraphics[scale=0.33]{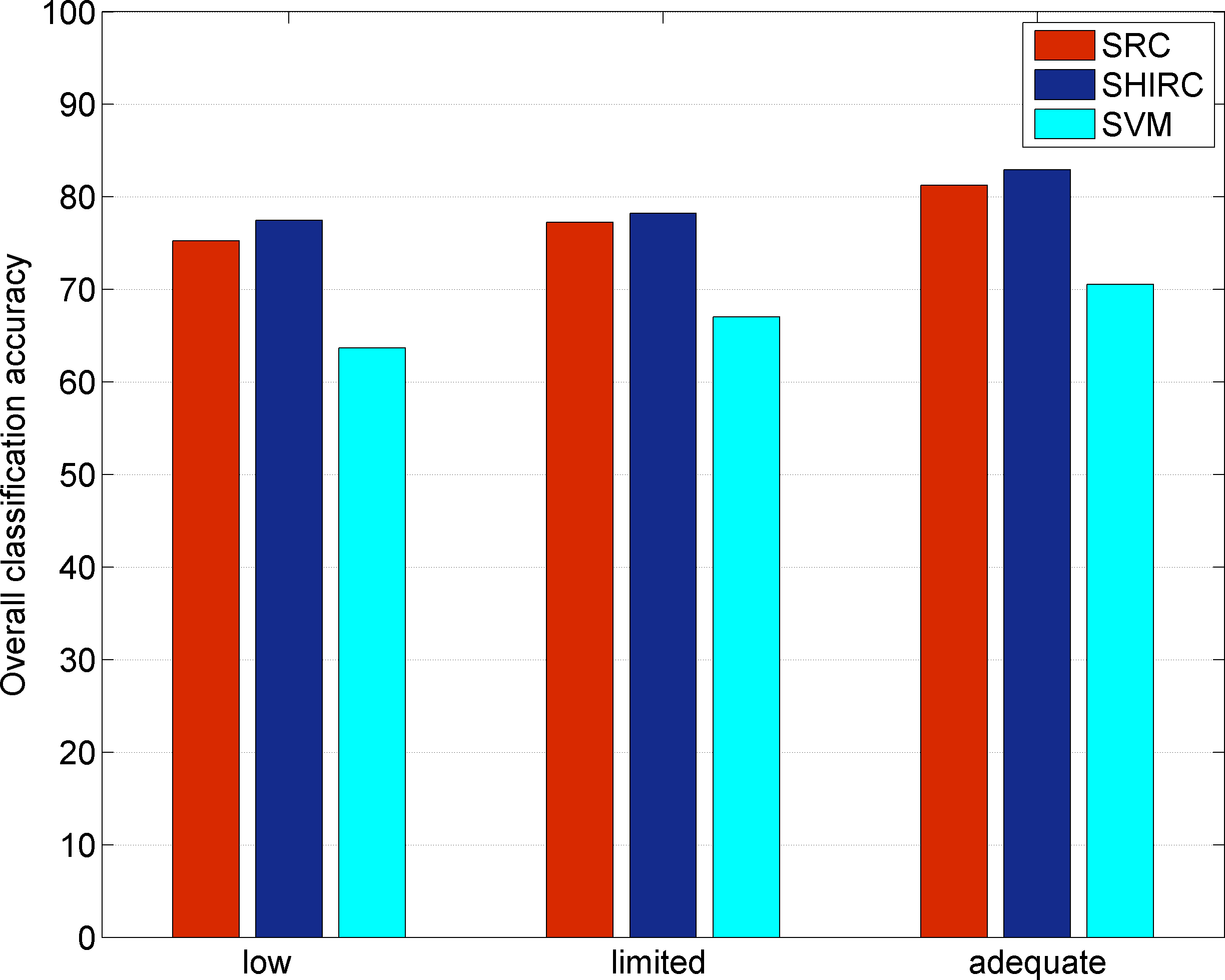}} \hspace{5mm}
  \subfigure[IBL.]{\includegraphics[scale=0.30]{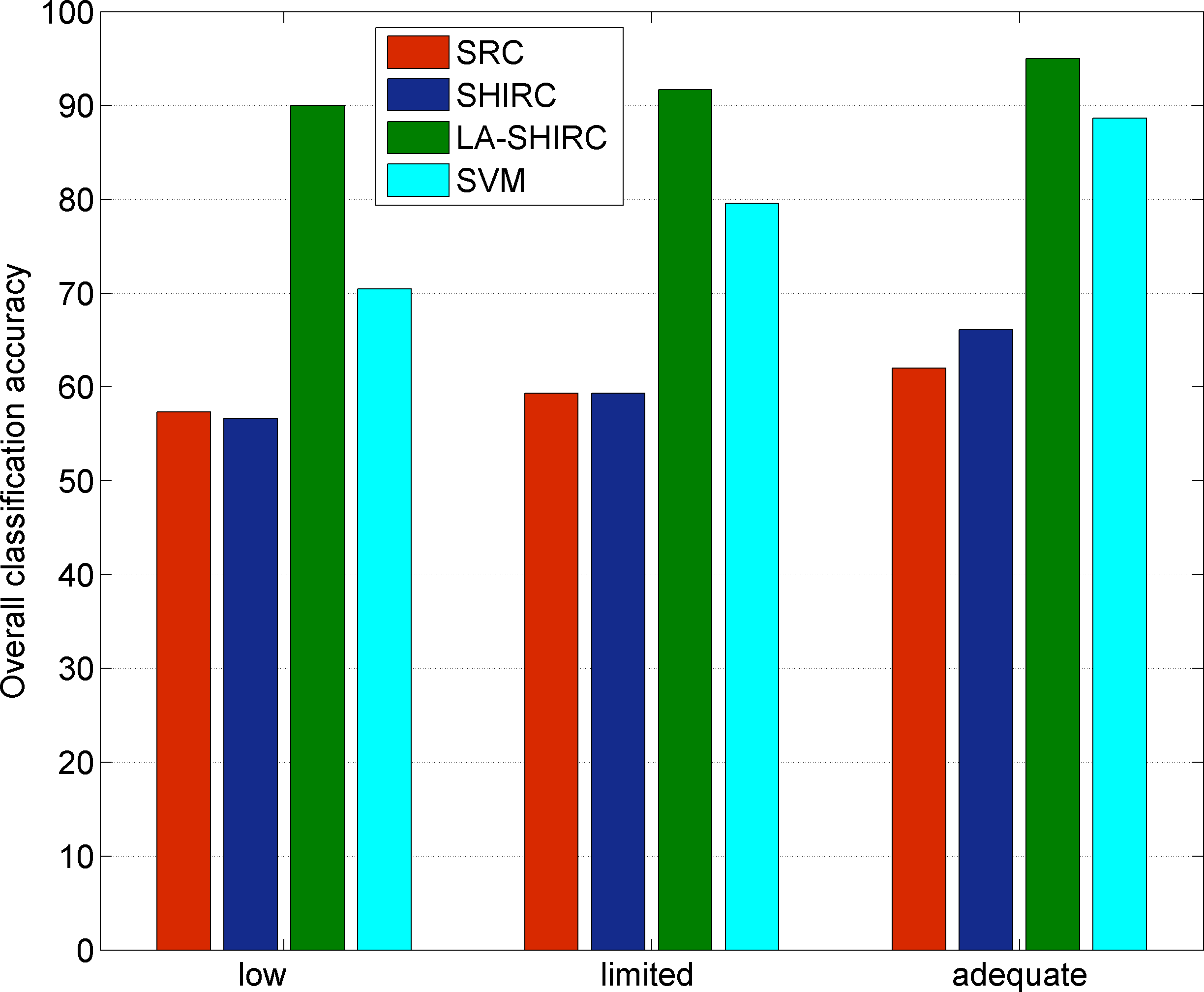}}
  \caption[Overall classification accuracy as a function of training set size for three different scenarios: low, limited and adequate training]{Overall classification accuracy as a function of training set size for three different scenarios: low, limited and adequate training.}
  \label{fig:trng_plots}
\end{figure}
Depending on the inherent degree of difficulty in classifying a particular image set and the severity of the penalty for misclassifying a diseased image\footnote{For example, DCIS requires immediate surgical attention, while a mild viral infection may only prolong for a few more days if not diagnosed early.}, a pathologist can choose an acceptable probability of miss and corresponding false alarm rate for each method. Table \ref{tab:fixed_det} shows that for each organ in the ADL data set, a higher false alarm must be tolerated with the SVM method, compared to SRC and SHIRC, in order to maintain a fixed rate of miss. For the IBL data set, the LA-SHIRC incurs the lowest false alarm rate to achieve a miss rate of $10\%$.

\subsection{Performance as Function of Training Set Size}
\label{sec:trng_size}

This experiment offers new insight into the practical performance of our algorithms.  Real-world classification tasks often suffer from the lack of availability of large training sets. We present a comparison of overall classification accuracy as a function of the training set size for the different methods.

We identify three scenarios of interest: (i) low training, (ii) limited training, and (iii) adequate training. In Fig. \ref{fig:trng_plots}(a), overall classification accuracy is reported for ADL data set (kidney) corresponding to the three training scenarios: 20 (low), 30 (limited) and 40 (adequate) training images  respectively. As before, comparison is made against the single channel SRC and state-of-the-art feature extraction plus SVM classifier. Unsurprisingly, all three methods suffer in performance as training is reduced.

Fig. \ref{fig:trng_plots}(b) reports analogous results for the IBL data set. Here the regime of low, limited and adequate training images are defined by 20, 40 and 60 images  respectively. Analyzing the results in Fig. \ref{fig:trng_plots}(b) for the IBL data set, a more interesting trend reveals itself. As discussed before in Section \ref{sec:local}, LA-SHIRC can lead to richer dictionaries made out of local image blocks even as the number of training images is not increased. This allows  LA-SHIRC to perform extremely well even under low training - offering about $90\%$ accuracy - as is evident from Fig. \ref{fig:trng_plots}(b). This benefit however comes at the cost of increased computational complexity at the time of inference because the dictionary size (number of columns) is significantly increased in LA-SHIRC vs. SHIRC.

\section{Structured Sparse Priors for Image Classification}

\subsection{Related Work in Model-based Compressive Sensing}
\label{sec:model_based_cs}

We introduced sparse signal representations in Chapter \ref{chapter:sparsity_gm} by discussing the compressive problem and the sparse representation-based classification framework. For the sake of clarity, it is worthwhile to reproduce here the key equations from Section \ref{sec:sparsity_sig_proc}:

\be
(\mbox{P}_0) \quad \min_{\vect x}\|\vect x\|_0 ~\mbox{subject to}~ \vect y = \mat A\vect x.
\label{eqch4:l0}
\ee
\be
(\mbox{P}_1) \quad \min_{\vect x}\|\vect x\|_1 ~\mbox{subject to}~ \vect y = \mat A\vect x.
\label{eqch4:l1}
\ee
\be
(\mbox{P}_2) \quad \min_{\vect x}\|\vect x\|_1 ~\mbox{subject to}~ \|\vect y - \mat A\vect x\|_2 < \epsilon.
\label{eqch4:l1_noise}
\ee

The optimization problem in (P$_2$) can be interpreted as maximizing the probability of observing $\vect x$ given $\vect y$ under the assumption that the coefficients of $\vect x$ are modeled as i.i.d. Laplacians. Thus, sparsity can be interpreted as a prior for signal recovery. This is a particular example of the broader Bayesian perspective: signal comprehension can be enhanced by incorporating contextual information (in this case, sparsity) as priors. Estimating the sparse coefficients in a Bayesian framework has the benefits of probabilistic predictions and automatic estimation of model parameters. The relevance vector machine \cite{tipping:jmlr01} has inspired more sophisticated substitutes for the Laplacian via hierarchical priors \cite{ji_bcs:tsp08,babacan:tip10}.

Sparsity is in fact a first-order description of structure in signals. However, often there is \emph{a priori} structure inherent to the sparse signals that is exploited for better representation, compression or modeling. As an illustration, a connected tree structure can be enforced on wavelet coefficients to capture the multi-scale dependence \cite{baraniuk_modelbasedcs:tit10}. Other such structured prior models have also been integrated into the CS framework \cite{carin_spike:tsp09,Blumensath_2009, Eldar_2009,Ji_2008,Stojnic_2009,Baron05distributedcompressed,duarte:tsp11}. The wavelet-based Bayesian approach in \cite{carin_spike:tsp09} employs a ``spike-and-slab'' prior \cite{ishwaran_spike:as05,george_spike:jasa93,chipman_spike:cjs96,carvalho_spike:jasa08}, which is a mixture model of two components representing the zero and nonzero coefficients, and dependencies are encouraged in the mixing weights across resolution scales.
Structure on sparse coefficients can also be enforced via probabilistic prior models, as demonstrated in \cite{cevher09:nips,cevher:spm10} by solving the following optimization problem:
\be
(\mbox{P}_4) \quad \max_{\vect{x}} f(\vect{x})~\mbox{subject to}~\|\vect{y}-\mat{A}\vect{x}\|_2 < \epsilon.
\label{eq:gm_sparsity}
\ee
The pdf $f$ \emph{simultaneously} captures the sparsity and structure (joint distributions of coefficients) of $\vect x$. In comparison, the standard CS recovery (P$_2$) captures only the sparse nature of $\vect x$.

\subsection{Overview of Contribution}

Consider a binary classification problem (classes $C_0$ and $C_1$) with the two class conditional pdfs represented by $f_{C_0}$ and $f_{C_1}$ respectively. We choose a fixed dictionary matrix $\mat A = [\mat A_0, ~\mat A_1]$. Suppose that we have access to $T$ labeled training vectors $\{\vect y_{i,t}\}_{t=1}^{T}$ and corresponding sparse features $\{\vect x_{i,t}\}_{t=1}^{T}$ from each class, where the index $i \in \{0,1\}$ refers to $C_0$ and $C_1$. Given a test vector $\vect y$, we solve a constrained posterior maximization problem \emph{separately for each class}:
\be
(\mbox{P}_5) \quad \vect{\hat{x}}^{(i)} = \arg\max_{\vect x} f_{C_i}(\vect x)~\mbox{s.t.} ~\|\vect y-\mat A\vect x\|_2 < \epsilon, \quad i = 0,1.
\label{eq:prior_class}
\ee
The two spike-and-slab priors $f_{C_i}$ are learned separately, in a class-specific manner, from training samples of the two classes.
Class assignment is performed as follows:
\be
\mbox{Class}(\vect y) = \arg\max_{i \in \{0,1\}} f_{C_i}(\vect{\hat{x}}^{(i)}).
\label{eq:ml_compare}
\ee

The evolution of this formulation can be traced organically through (P$_0$)-(P$_5$), and it represents a consummation of ideas developed for model-based CS into a general framework for sparse model-based classification. Owing to its proven success in modeling sparsity, the spike-and-slab prior is an excellent initial choice for the $f_{C_i}$ in this framework. This is validated next by theoretical analysis and experiments.

\section{Design of Discriminative Spike-and-slab Priors}
\label{sec:contrib}
\begin{figure}[t]
  \centering
  \includegraphics[scale=0.23]{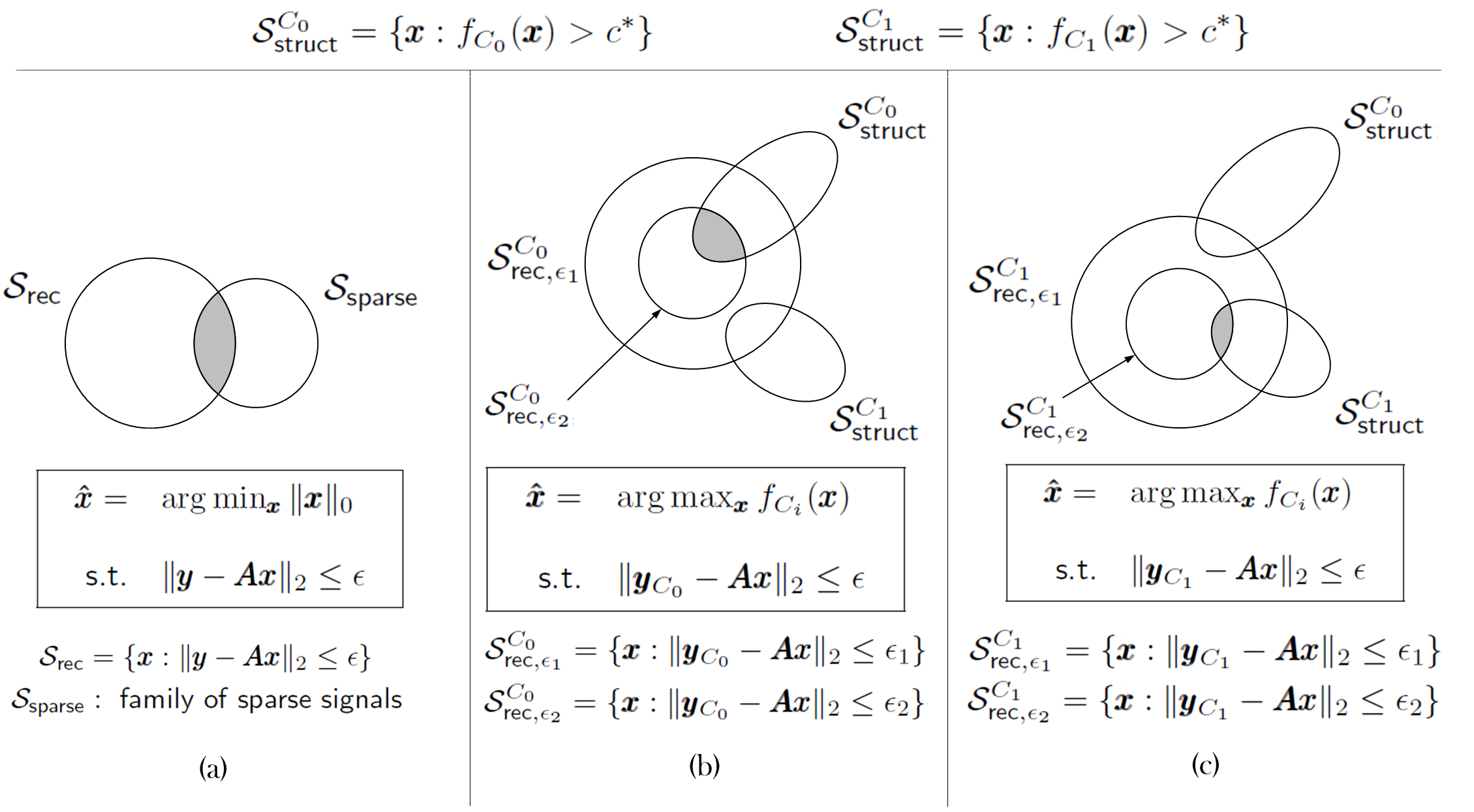}
  \caption[SSPIC: Set-theoretic comparison]{Set-theoretic comparison: (a) traditional CS recovery, and (b)-(c) our proposed framework - structured sparsity using class-specific priors. In (b), the test vector $\vect y_{C_0}$ is actually from class $C_0$, while in (c), the test vector $\vect y_{C_1}$ is from $C_1$.}
  \label{fig:set_schematic}
\end{figure}

\subsection{Set-theoretic Interpretation}
\label{sec:set_theory}

Fig. \ref{fig:set_schematic} offers a set-theoretic viewpoint to illustrate the central idea of our SSPIC framework. Fig. \ref{fig:set_schematic}(a) represents the traditional CS recovery problem. $\mathcal{S}_{\mbox{\scriptsize{rec}}}$ is the sub-level set of all vectors that lead to reconstruction error less than a specific tolerance $\epsilon$. $\mathcal{S}_{\mbox{\scriptsize{sparse}}}$ is the set of all vectors with only a few non-zero coefficients. The vectors that lie in the intersection of these two sets (shaded region in Fig. \ref{fig:set_schematic}(a)) are exactly the set of solutions to \eqref{eq:l0}.

Figs. \ref{fig:set_schematic}(b) and \ref{fig:set_schematic}(c) describe our idea for binary classification. Now we have two sub-level sets $\mathcal{S}_{\mbox{\scriptsize{rec}},\epsilon_1}$ and $\mathcal{S}_{\mbox{\scriptsize{rec}},\epsilon_2}$, which correspond to vectors $\vect x$ leading to reconstruction error not greater than $\epsilon_1$ and $\epsilon_2$ respectively ($\epsilon_1 > \epsilon_2$).

Our contribution is the introduction of the two sets $\mathcal{S}_{\mbox{\scriptsize{struct}}}^{C_i}, i = 0,1$. These sets enforce additional class-specific structure on the sparse coefficients. Let us first consider Fig. \ref{fig:set_schematic}(b), where the sample test vector $\vect y_{C_0}$ is in fact from class $C_0$. The two sets $\mathcal{S}_{\mbox{\scriptsize{struct}}}^{C_0}$ and $\mathcal{S}_{\mbox{\scriptsize{struct}}}^{C_1}$ are defined by priors $f_{C_0}$ and $f_{C_1}$ respectively, which \emph{simultaneously encode sparsity and class-specific structure}. For a relaxed reconstruction error tolerance $\epsilon_1$, both these sets have non-zero intersection with $\mathcal{S}_{\mbox{\scriptsize{rec}},\epsilon_1}^{C_0}$. As a result, both the class-specific optimization problems in \eqref{eq:prior_class} are feasible and this increases the possibility of the test vector being misclassified. However, as the error bound is tightened to $\epsilon_2$, we see that only $\mathcal{S}_{\mbox{\scriptsize{struct}}}^{C_0}$ intersects with $\mathcal{S}_{\mbox{\scriptsize{rec}},\epsilon_2}^{C_0}$, and the solution to \eqref{eq:prior_class} correctly identifies the class of the test vector as $C_0$.

An analogous argument holds in Fig. \ref{fig:set_schematic}(c), where the test vector $\vect y_{C_1}$ is now from $C_1$. As the reconstruction error tolerance is reduced, only $\mathcal{S}_{\mbox{\scriptsize{struct}}}^{C_1}$ intersects with $\mathcal{S}_{\mbox{\scriptsize{rec}},\epsilon_2}^{C_1}$.

The proposed framework extends to multi-class classification in a natural manner, by defining multiple such structured priors, one per class, and solving the corresponding optimization problems in parallel. In terms of computational complexity, it is similar to the requirements of SRC.

\subsection{Spike-and-slab Priors}
\label{sec:spike-slab}

\begin{figure}[t]
  \centering
  \includegraphics[scale=0.4]{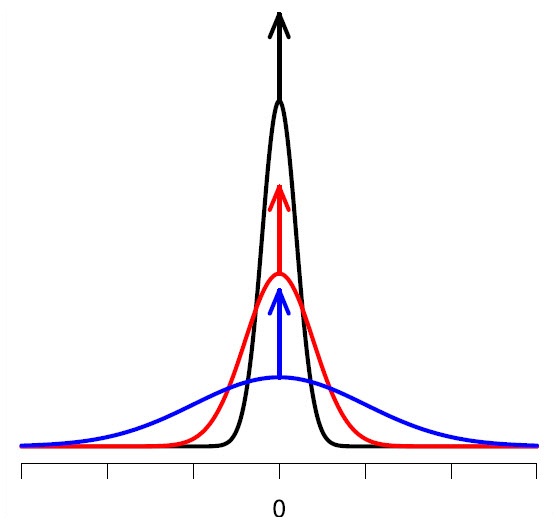}
  \caption{Probability density function of a spike-and-slab prior.}
  \label{fig:spike_slab}
\end{figure}
What priors do we choose per class? We remind ourselves that the priors $f_{C_i}$ should be chosen to simultaneously capture the sparse nature of and structure inherent to $\vect x$. A particularly well-suited example of probabilistic sparsity prior is the spike-and-slab prior which is widely used in Bayesian regression models \cite{ishwaran_spike:as05,george_spike:jasa93,chipman_spike:cjs96,carvalho_spike:jasa08}. In fact, the spike-and-slab prior is acknowledged to be the gold standard for sparse inference in the Bayesian set-up \cite{titsias:nips11}. Each individual coefficient $x_i$ of $\vect x$ is modeled as a mixture of two components:
\be
x_i \sim (1-\gamma_i)\delta_0 + \gamma_if(x_i),
\ee
where $\delta_0$ is a point mass concentrated at zero (the ``spike''), and $f_i$ (the ``slab'') is any suitable distribution on the non-zero coefficient (e.g. a Gaussian). Fig. \ref{fig:spike_slab} shows an illustration of a spike-and-slab prior. Structural sparsity is encoded by the parameter $\gamma_i \in [0,1]$. For example, the slab term $(f_i)$ is expected to dominate for a non-zero coefficient and this can be enforced by choosing $\gamma_i$ closer to 1; likewise $\gamma_i$ is chosen closer to 0 to encourage a zero coefficient.

Fig. \ref{fig:sparsity} illustrates a binary classification problem for the task of face recognition. $\mat A_1$ and $\mat A_2$ are built using training samples from the two respective classes. For $\vect y$ from class 1, the first half of coefficients in $\vect \alpha$ are expected to be active. For our choice of spike-and-slab priors, this leads to weights $\gamma = 0$ for the corresponding nodes in the graph. Similarly, the inactive coefficients are identified with weights $\gamma = 1$.

\subsection{Analytical Development}
\label{sec:opt_prob}
\begin{figure}[t]
  \centering
  \includegraphics[scale=0.11]{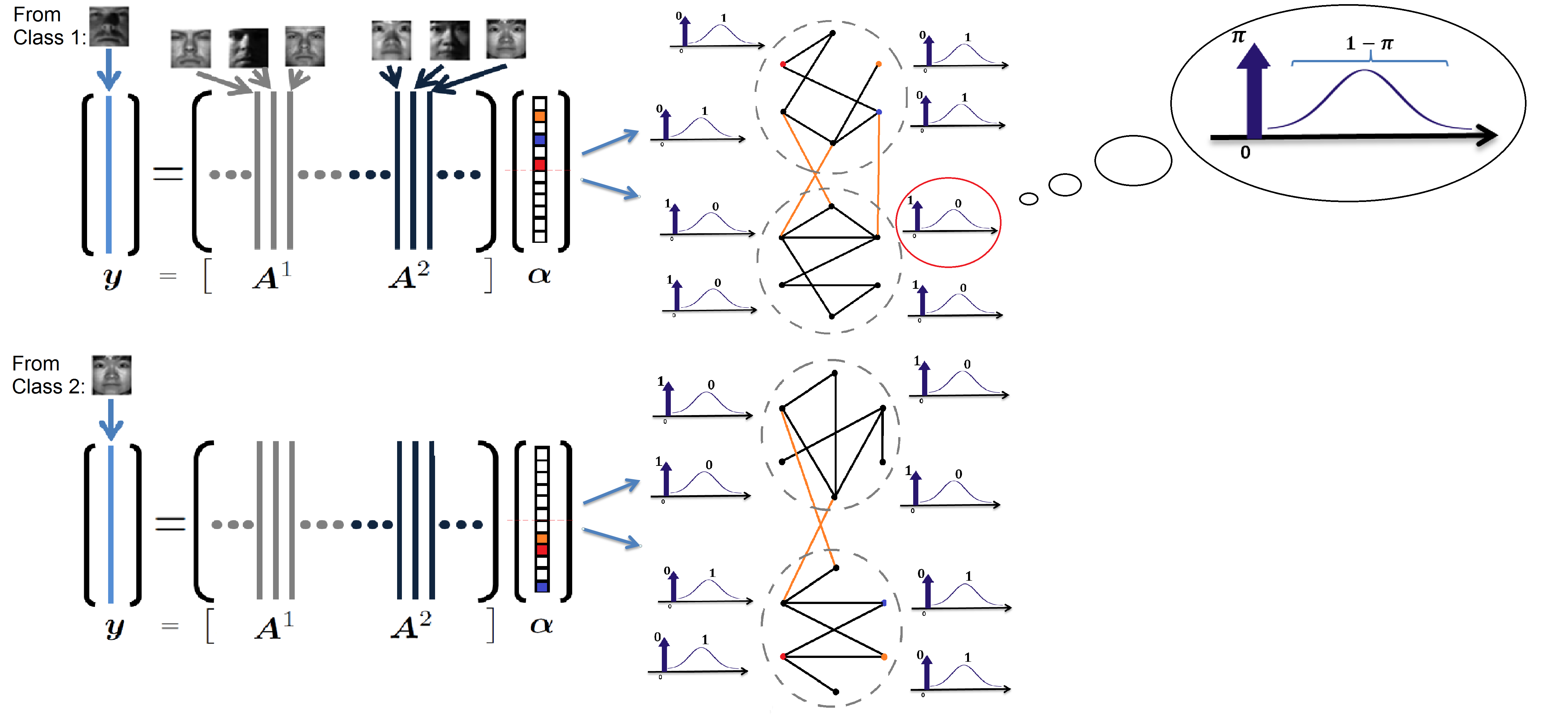}
  \caption[Structured spike-and-slab priors for sparse representation-based classification]{Structured spike-and-slab priors for sparse representation-based classification.}
  \label{fig:sparsity}
\end{figure}

Inspired by a recent \emph{maximum a posteriori} (MAP) estimation approach to variable selection using spike-and-slab priors\cite{yen:as11a}, we develop the corresponding Bayesian framework for classification. Another model is developed analogously for the other class, albeit with different parameters; considering the model for a single class keeps notation simpler without an additional class index term. We consider the following linear representation model:
\be
\vect y = \mat A\vect x + \vect n,
\ee
where $\vect y \in \mathbb{R}^m, \vect x \in \mathbb{R}^n, \mat A \in \mathbb{R}^{m\times n}$, and $\vect n \in \mathbb{R}^m$ models Gaussian noise. We define $\gamma_i := \mathbb{I}(x_i \neq 0), i = 1,\ldots,n$, i.e. $\gamma_i$ is binary-valued. It is an indictor variable which takes the value one if the corresponding coefficient $x_i \neq 0$ in $\vect x$, and zero otherwise. We set up the Bayesian formulation as follows:
\begin{subequations}
\begin{align}
\vect y | \mat A, \vect x, \vect \gamma, \sigma^2 & \sim \mathcal{N}\left(\mat A\vect x, \sigma^2\mat I \right) \label{eq:y}\\
x_i | \sigma^2, \gamma_i, \lambda & \sim \gamma_i\mathcal{N}(0,\sigma^2\lambda^{-1}) + (1-\gamma_i)\mathbb{I}(x_i = 0) \label{eq:spike-slab2}\\
\sigma^2 | \tau_1,\tau_2 & \sim \Gamma^{-1}(\tau_1,\tau_2) \\
\gamma_i | \kappa_i & \sim \mbox{Bernoulli}(\kappa_i),~i = 1,\ldots,n.
\end{align}
\end{subequations}
Here, $\mathcal{N}(\cdot)$ represents the Gaussian distribution. It can be seen that \eqref{eq:spike-slab2} represents the i.i.d. spike-and-slab prior on each coefficient of $\vect x$. The slab term is a zero-mean Gaussian with variance $\sigma^2\lambda^{-1}$. The role of $\lambda$ as a regularization parameter will become clear shortly. In all generality, we select different $\kappa_i$ to parameterize each coefficient $\gamma_i, i = 1,\ldots, n$. It can be seen that \eqref{eq:spike-slab2} represents the spike-and-slab prior on $x_i$.

Exploiting the mixture form of the prior, we can write the pdf for $x_i$ in \eqref{eq:spike-slab2} as:
\be
x_i \sim~\left(\mathcal{N}(0,\sigma^2\lambda^{-1})\right)^{\gamma_i}.\left(\mathbb{I}(x_i = 0)\right)^{1-\gamma_i}
\label{eq:spike-slab-prod}
\ee

The joint posterior density is then given by:
\bea
f(\vect x,\vect \gamma,\sigma^2|\mat A,\vect y,\lambda,\tau_1,\tau_2,\vect \kappa) &\propto &f(\vect y |\mat A,\vect x,\vect \gamma,\sigma^2)f(\vect x|\vect \gamma,\sigma^2,\lambda) \nonumber\\
& & f(\sigma^2|\tau_1,\tau_2)f(\vect \gamma | \vect \kappa).
\label{eq:joint-posterior}
\eea
The optimal $\vect x^\ast, \vect \gamma^\ast, \sigma^{2\ast}$ are obtained by MAP estimation as:
\be
(\vect x^\ast, \vect \gamma^\ast, \sigma^{2\ast}) = \arg \min_{\vect x,\vect \gamma,\sigma^2} \left\{-2\log f(\vect x,\vect \gamma,\sigma^2|\mat A,\vect y,\lambda,\tau_1,\tau_2,\vect \kappa)\right\}.
\label{eq:map-estimate}
\ee
We now evaluate each of the terms separately on the right hand side of Eq. \eqref{eq:joint-posterior}.
\be
f(\vect y |\mat A,\vect x,\vect \gamma,\sigma^2) = \frac{1}{(2\pi)^{m/2}\sigma^m}\exp\left\{-\frac{1}{2\sigma^2}(\vect y - \mat A\vect x)^T(\vect y - \mat A\vect x)\right\}
\ee
\bea
\Rightarrow -2\log f(\vect y |\mat A,\vect x,\vect \gamma,\sigma^2) & = &\frac{1}{\sigma^2}(\vect y - \mat A\vect x)^T(\vect y - \mat A\vect x) \nonumber\\
&& + m\log \sigma^2 + m\log (2\pi).
\eea

\bea
f(\vect x|\vect \gamma,\sigma^2,\lambda) & = &\prod_{i=1}^{n}\left\{\left(\frac{1}{\sqrt{2\pi\sigma^2/\lambda}}\right)^{\gamma_i}\exp\left(-\frac{\gamma_ix_i^2}{2\sigma^2\lambda^{-1}}\right)\right\}\nonumber\\
& & \prod_{i=1}^{n}\mathbb{I}(x_i = 0)^{1-\gamma_i}\\
& = &\left(\frac{2\pi\sigma^2}{\lambda}\right)^{-\frac{1}{2}\sum_{i=1}^{n}\gamma_i}\exp\left\{-\frac{1}{2\sigma^2\lambda^{-1}}\vect x^T\vect x\right\}\nonumber\\
& &\prod_{i=1}^{n}\mathbb{I}(x_i = 0)^{1-\gamma_i}
\eea
\be
\Rightarrow -2\log f(\vect x|\vect \gamma,\sigma^2,\lambda) = \frac{\vect x^T\vect x}{\sigma^2\lambda^{-1}} + \left(\sum_{i=1}^{n}\gamma_i\right)\log\left(\frac{2\pi\sigma^2}{\lambda}\right) -2\sum_{i=1}^{n}(1-\gamma_i)\log\mathbb{I}(x_i = 0).
\ee
The final term on the right hand side evaluates to zero, since $\mathbb{I}(x_i = 0) = 1 \Rightarrow \log\mathbb{I}(x_i = 0) = 0$, and $\mathbb{I}(x_i = 0) = 0 \Rightarrow x_i \neq 0 \Rightarrow \gamma_i = 1 \Rightarrow (1-\gamma_i) = 0$. Therefore,
\be
-2\log f(\vect x|\vect \gamma,\sigma^2,\lambda) = \frac{\vect x^T\vect x}{\sigma^2\lambda^{-1}} + \left(\sum_{i=1}^{n}\gamma_i\right)\log\left(\frac{2\pi\sigma^2}{\lambda}\right).
\ee

\bea
f(\sigma^2|\tau_1,\tau_2) & = &\frac{\tau_2^{\tau_1}}{\Gamma(\tau_1)}\sigma^{2(-\tau_1-1)}\exp\left(-\frac{\tau_2}{\sigma^2}\right)\\
\Rightarrow -2\log f(\sigma^2|\tau_1,\tau_2) & = &\frac{2\tau_2}{\sigma^2} + 2(\tau_1+1)\log \sigma^2 - 2\tau_1\log\tau_2 + 2\log \Gamma(\tau_1).
\eea

Finally,
\begin{subequations}
\begin{align}
f(\vect \gamma | \vect \kappa) & = \prod_{i=1}^{n}\kappa_i^{\gamma_i}(1-\kappa_i)^{1-\gamma_i}\\
\Rightarrow -2\log f(\vect \gamma | \vect \kappa) & = \sum_{i=1}^{n}\gamma_i\log\left(\frac{(1-\kappa_i)^2}{\kappa_i^2}\right) - \sum_{i=1}^{n}\log(1-\kappa_i)^2. \label{eq:indiv_kappa}
\end{align}
\end{subequations}

Collecting all these expressions together:
\bea
(\vect x^\ast, \vect \gamma^\ast, \sigma^{2\ast}) & = &\arg \min \frac{1}{\sigma^2}(\vect y - \mat A\vect x)^T(\vect y - \mat A\vect x) + m\log \sigma^2 \nonumber\\
& &  + \frac{\vect x^T\vect x}{\sigma^2\lambda^{-1}} + \left({\sum_{i=1}^{n}\gamma_i}\right)\log\left(\frac{2\pi\sigma^2}{\lambda}\right) + \frac{2\tau_2}{\sigma^2} \nonumber\\
& & + 2(\tau_1+1)\log \sigma^2 + {\sum_{i=1}^{n}\gamma_i}\log\left(\frac{(1-\kappa_i)^2}{\kappa_i^2}\right) \label{eq:opt_prob_ki}.
\eea

Note that in the analytical development so far, we have introduced the formulation for a single class. The formulations for the other classes in a general multi-class scenario are obtained in exactly the same manner, but with different sets of model parameters per class.\\

\noindent \textbf{Tractability of optimization problem:} We observe that \eqref{eq:opt_prob_ki} comprises a collection of terms that: (i) lacks direct interpretation in terms of modeling sparsity, and (ii) leads to a difficult optimization problem. So, we introduce the simplifying assumption of choosing a single scalar $\kappa$ per class. With this simplification, we now have:
\begin{subequations}
\begin{align}
f(\vect \gamma | \kappa) & = \prod_{i=1}^{n}\kappa^{\gamma_i}(1-\kappa)^{1-\gamma_i} = \kappa^{\sum_{i=1}^{n}\gamma_i}(1-\kappa)^{n-\sum_{i=1}^{n}\gamma_i}\\
\Rightarrow -2\log f(\vect \gamma | \kappa) = & \sum_{i=1}^{n}\gamma_i\log\left(\frac{(1-\kappa)^2}{\kappa^2}\right) - 2n\log(1-\kappa).
\end{align}
\end{subequations}

Plugging this back into \eqref{eq:opt_prob_ki}:
\bea
(\vect x^\ast, \vect \gamma^\ast, \sigma^{2\ast}) & = &\arg \min_{\vect x,\vect \gamma,\sigma^2} \frac{1}{\sigma^2}(\vect y - \mat A\vect x)^T(\vect y - \mat A\vect x) + m\log \sigma^2 \nonumber \\
& & + \frac{\vect x^T\vect x}{\sigma^2\lambda^{-1}} + \left(\sum_{i=1}^{n}\gamma_i\right)\log\left(\frac{2\pi\sigma^2}{\lambda}\right) + \frac{2\tau_2}{\sigma^2} \nonumber\\
& & + 2(\tau_1+1)\log \sigma^2 + \sum_{i=1}^{n}\gamma_i\log\left(\frac{(1-\kappa)^2}{\kappa^2}\right).
\eea
Now we observe another interesting aspect of this formulation. The term $\sum_{i=1}^{n}\gamma_i$ counts the number of $\gamma_i$ which are equal to 1, since $\gamma_i$ is a binary variable. In turn, this is equivalent to counting the number of entries of $\vect x$ that are non-zero, i.e. the $l_0$-norm of $\vect x$. So,
\bea
L(\vect x, \vect \gamma, \sigma^2) & = &\frac{1}{\sigma^2} \left\{\|\vect y - \mat A\vect x\|_2^2 + \lambda\|\vect x\|_2^2 + \rho_{\sigma^2,\lambda,\kappa}\|\vect x\|_0\right. \nonumber\\
& & \left. + (m+2\tau_1+2)\sigma^2\log \sigma^2 + 2\tau_2\right\},
\label{eq:final-opt}
\eea
where $\rho_{\sigma^2,\lambda,\kappa} := \sigma^2\log\left(\frac{2\pi\sigma^2(1-\kappa)^2}{\lambda\kappa^2}\right)$. For fixed $\sigma^2$, the cost function reduces to:
\be
L(\vect x; \sigma^2) = \|\vect y - \mat A\vect x\|_2^2 + \lambda\|\vect x\|_2^2 + \rho_{\sigma^2,\lambda,\kappa}\|\vect x\|_0.
\label{eq:fixed-sigma}
\ee

In fact, we obtain multiple such cost functions $L_0(\vect x, \vect \gamma, \sigma^2)$ and $L_1(\vect x, \vect \gamma, \sigma^2)$, corresponding to to each class. Different sets of data-dependent parameters $\rho_{\sigma^2,\lambda,\kappa}$ and $\lambda$ are learned from the training images of each class. The general form of the classification rule for multiple ($K$) classes is as follows:
\be
\mbox{Class}(\vect y) = \arg\max_{i \in \{1,\ldots,K\}} f_{C_i}(\vect{\hat{x}}^{(i)}).
\label{eq:ml_compare_k}
\ee

\subsection{SSPIC: Some Observations}
\label{sec:observations}

For fixed $\sigma^2$, the cost function reduces to:
\be
(\mbox{P}_5) \quad L(\vect x; \sigma^2) = \|\vect y - \mat A\vect x\|_2^2 + \lambda\|\vect x\|_2^2 + \rho_{\sigma^2,\lambda,\kappa}\|\vect x\|_0.
\label{eq:fixed-sigma2}
\ee
To summarize our analytical contribution, we initially choose a sparsity-inducing spike-and-slab prior per class and perform MAP estimation. With reasonable simplifications on model structure, we obtain the final formulation \eqref{eq:fixed-sigma} which explicitly captures sparsity in the form an $\l_0$-norm minimization term. This is intuitively satisfying, since we are looking for sparse vectors $\vect x$. Thereby, we offer a Bayesian perspective to SRC.

We can immediately draw parallels to SRC. However, our framework has two key differences when compared to SRC. One, the $\|\vect x\|_2$-term, which enforces smoothness in $\vect x$, is absent in SRC.  Extensions to SRC have considered the addition of regularizers such as the $\|\vect x\|_2$-term to induce group sparsity \cite{yu:isbi11}. However these have largely been heuristic yet meaningful choices to improve classification accuracy over SRC. On the other hand, our SSPIC framework handles the classification problem in a completely Bayesian setting. Secondly, the regularization parameter in SRC is chosen uniformly to be the same for images from all classes. However, our framework learns different parameters per class, leading to the solution of multiple optimization problems in parallel. On a related note, (P$_5$) is identical to the elastic net proposed for statistical regression problems \cite{zou:jrss05} if the $l_0$-term is relaxed to its $l_1$-counterpart. Of course, it must be mentioned that elastic net was proposed in the context of sparse signal modeling and not for classification tasks.

What benefit does the Bayesian approach buy? A limitation of SRC is its requirement of abundant training (highly overcomplete dictionary $\mat A$). However in many real-world problems such as hyperspectral image classification, there is a scarcity of available training. The use of carefully selected class-specific priors alleviates the burden on the number of training images required, as we shall see in Section \ref{sec:experiments}.

\subsection{Parameter Learning}

Different sets of parameters $\rho_{\sigma^2,\lambda,\kappa}$ and $\lambda$ are learned for different classes. The classification accuracy is tied to the accuracy of estimating these parameters which encode discriminative information. A common way of selecting these parameters is by cross-validation on the training samples. An alternate approach involves learning the posterior distributions in the parameters in a Markov chain Monte Carlo (MCMC) setting, an example of which is described in \cite{carin_spike:tsp09}. In our framework, the inverse-Gamma prior is chosen specifically because it is conjugate to the Gaussian distribution, leading to closed form expressions for the posterior. Similarly, $\kappa$ can be sampled from a Beta distribution, since the Beta-Bernoulli is a well-known conjugate pair of distributions. With these assumptions and good initialization of \emph{hyperparameters}, the MCMC method will lead to posterior estimates for all parameters (see Algorithm \ref{alg:sspic} in Appendix \ref{appendix:b}). In fact, a significant advantage of this approach is that it can be completely training-free \cite{carin_spike:tsp09} if hyperparameters are carefully selected.

%
%
%
%

\begin{table}[t]
  \centering
  \caption[AVIRIS Indian Pines hyperspectral image]{AVIRIS Indian Pines hyperspectral image.}
  \label{tab:hsi_aviris_sspic}
  \begin{tabular}{|l|c|c|c|c|c|}
  \hline
  Class type & Training & Test & SVM & SRC & SSPIC \\
  \hline
  Corn-notill & 144 & 1290 & 88.06 & 89.32 & 91.82 \\
  Grass/trees & 75 & 672 & 96.72 & 97.64 & 97.85 \\
  Soybeans-notill & 97 & 871 & 72.91 & 79.64 & 85.26 \\
  Soybeans-min & 247 & 2221 & 82.14 & 88.32 & 89.73 \\
  \hline
  Overall & 563 & 5054 & 84.12 & 88.34 & 90.57 \\
  \hline
  \end{tabular}
\end{table}

\begin{figure}[t]
  \centering
  \subfigure[Ground truth]{\includegraphics[scale=0.3]{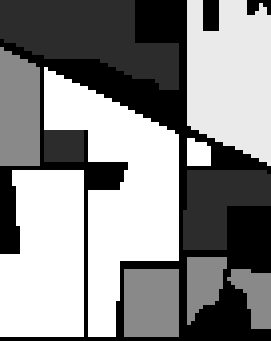}} \hspace{2mm}
  \subfigure[SVM.]{\includegraphics[scale=0.3]{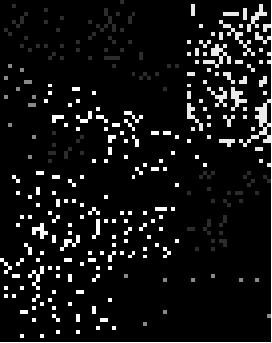}} \hspace{2mm}
  \subfigure[SRC.]{\includegraphics[scale=0.3]{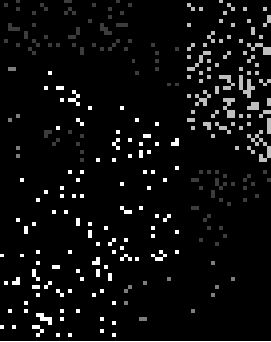}} \hspace{2mm}
  \subfigure[SSPIC.]{\includegraphics[scale=0.3]{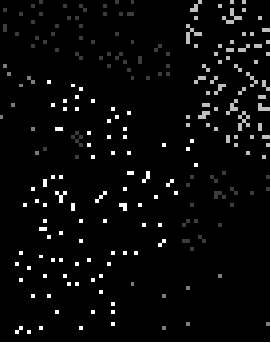}} \vspace{-3mm}
  \caption[Error maps for AVIRIS Indian Pine data set]{Error maps for AVIRIS Indian Pine data set.}
  \label{fig:class_map}
\end{figure}

\begin{table}[t]
  \centering
  \caption[Confusion matrix: Lung]{Confusion matrix: Lung.}
  \label{tab:conf_lung_sspic}
  \begin{tabular}{|c|c|c||c|}
  \hline
  Class & Healthy & Inflammatory & Method \\
  \hline
  Healthy & 0.734 & 0.266 & SVM \\
  & 0.706 & 0.294 & SRC \\
  & \textbf{0.868} & 0.132 & SSPIC \\
  \hline
  Inflammatory& 0.333 & 0.667 &  SVM \\
  & 0.366 & 0.634 &  SRC \\
  & 0.294 & \textbf{0.706} & SSPIC \\
  \hline
  \end{tabular}
\end{table}

\section{Experimental Results}
\label{sec:experiments}

\begin{figure}[t]
  \centering
  \subfigure[Healthy lung.]{\includegraphics[scale=0.3]{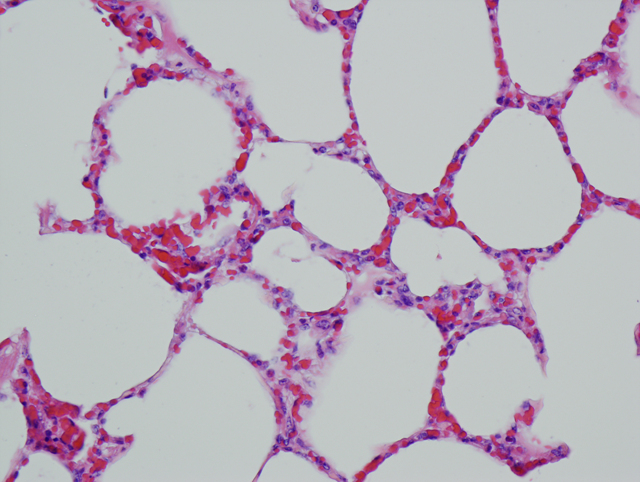}} \hspace{3mm}
  \subfigure[Inflamed lung.]{\includegraphics[scale=0.3]{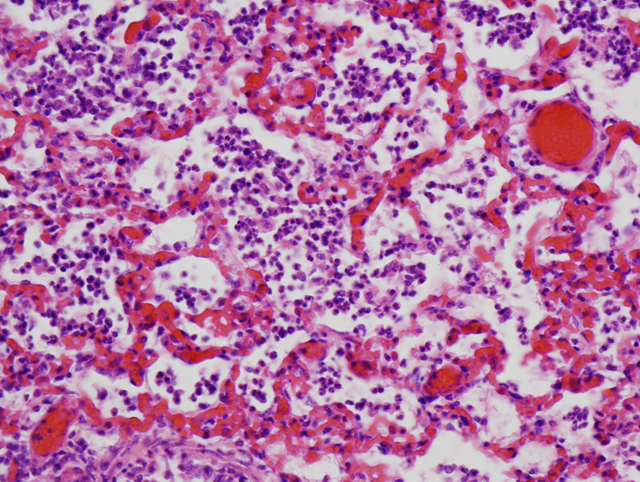}}
  \caption[Sample lung tissue images]{Sample lung tissue images.}
  \label{fig:hist_adl_sspic}
\end{figure}

\begin{figure}[t]
  \centering
  \includegraphics[scale=0.5]{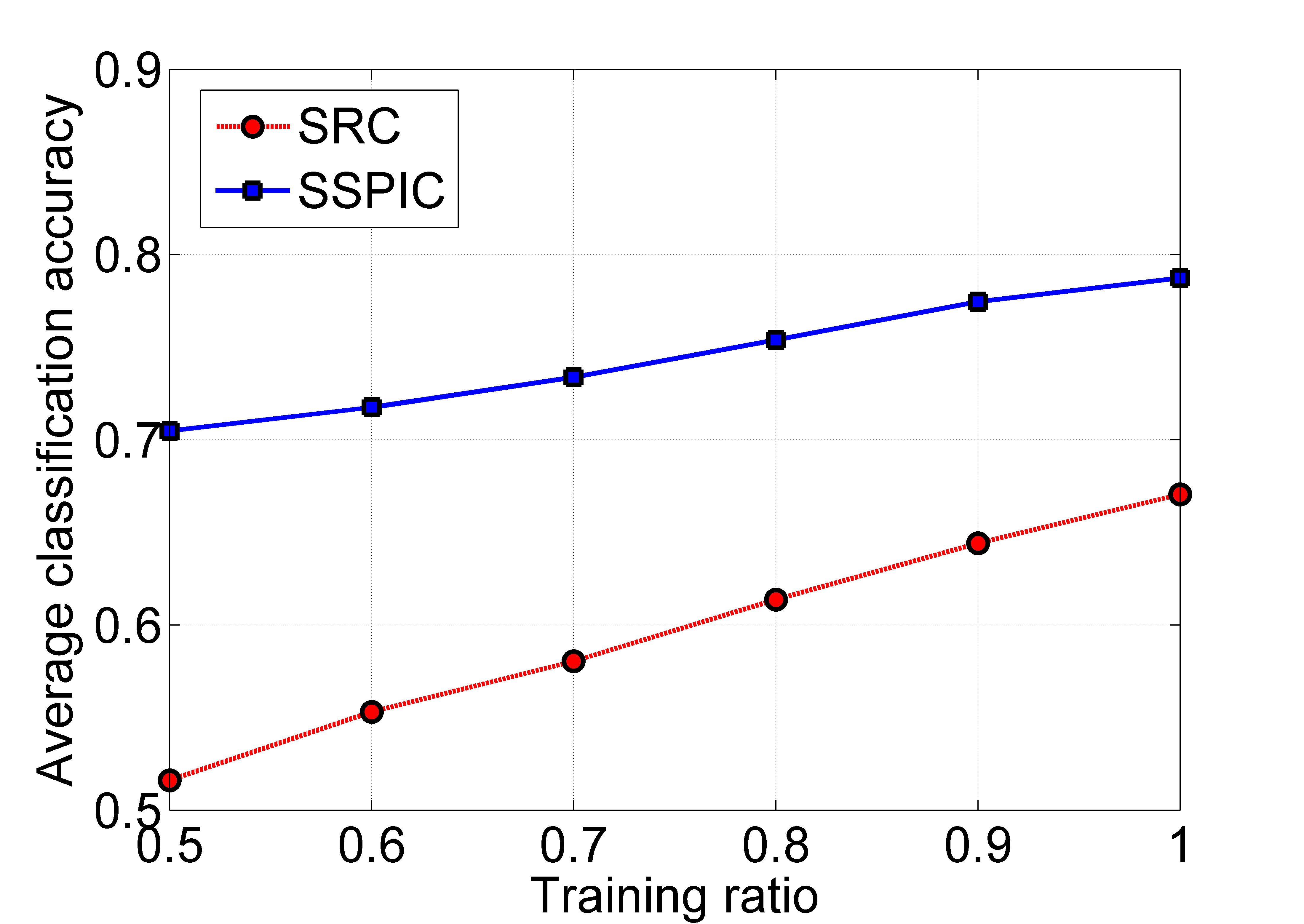}
  \caption[Performance as a function of training ratio]{Performance as a function of training ratio. Training ratio is the actual fraction of 40 training images per class used in the experiment.}
  \label{fig:hist_adl_trng}
\end{figure}

We perform experiments on two different data sets. We solve an $l_1$-relaxation of (P$_5$) per class using software from the SPArse Modeling Software (SPAMS) toolbox \cite{bach_spams:ftml12}.

\noindent \textbf{Hyperspectral image classification:} The first set is the AVIRIS Indian Pines hyperspectral image \cite{aviris}. The AVIRIS sensor generates 220 bands across the spectral range from 0.2 to 2.4 $\mu$m, of which only 200 bands are considered by removing 20 water absorption bands. This image has spatial resolution of 20m per pixel and spatial dimension $145 \times 145$. Following \cite{gualtieri:spie98}, we consider a subset scene of size $86 \times 68$ consisting of pixels $[27-94]\times [31-116]$. Each pixel is classified into one of four classes: corn-notill, grass/trees, soybeans-notill, and soybeans-min. Table \ref{tab:hsi_aviris_sspic} shows the class-wise classification rates for a specific training-test combination, comparing three approaches: state of the art support vector machine (SVM) \cite{gualtieri:spie98}, SRC, and SSPIC. Fig. \ref{fig:class_map} shows the error maps.

\noindent \textbf{Histopathological image classification:} We classify bovine lung tissue samples as either healthy or inflammatory. The images are acquired by pathologists at the Animal Diagnostic Laboratory, Pennsylvania State University \cite{adl}. The H$\&$E-stained tissues are scanned using a whole slide digital scanner at 40x optical magnification. All images are downsampled to $100 \times 75$ pixels (aliasing-free). Sample healthy and inflamed tissue images are shown in Fig. \ref{fig:hist_adl}(a)-(b). $40$ images per condition are used for training, and performance is evaluated over a set of $10$ images from each class. The ground truth labels for healthy and inflammatory tissue are obtained via manual detection and segmentation by ADL pathologists. In order to mitigate selection bias, the experiment is repeated over 1000 trials of randomly selected training. Table \ref{tab:conf_lung_sspic} reports the mean classification accuracy as a confusion matrix, where rows refer to the actual identity of test images and columns refer to the classifier output. We compare the performance of SSPIC with SRC using the luminance channels of the color images and a state of the art SVM classifier \cite{madabhushi:rbe09}.

For each data set, the classification rates for SSPIC are better than those for SVM and SRC. In Fig. \ref{fig:hist_adl_trng}, the average (of healthy and inflamed tissue) classification rate is plotted as a function of training set size. Given adequate training, SSPIC offers significant benefits over SRC. Crucially, as training ratio is reduced, SSPIC suffers a graceful decay in performance. This is enabled by the use of class-specific priors that offer additional discriminability over class-specific dictionaries.

\section{Conclusion}
\label{ch4:conclusion}

In this chapter, we have proposed discriminative models on sparse signal representations in two different ways. First, we develop a simultaneous sparsity model for histopathological image representation and classification. The central idea of our approach is to exploit the correlations among the red, green and blue channels of the color images in a sparse linear model setting with attendant color channel constraints. We formulate and solve a new sparsity-based optimization problem. We also introduce a robust locally adaptive version of the simultaneous sparsity model to address the issue of correspondence of local image objects located at different spatial locations. This modification results in benefits that have significant practical relevance: we demonstrate that the sparsity model for classification can work even under limited training if local blocks are chosen carefully.

In the second half of this chapter, discriminative structure in sparse representations is revealed by an appropriate choice of sparsity-inducing priors. We use class-specific dictionaries in conjunction with discriminative class-specific priors, specifically the spike-and-slab prior widely applied in Bayesian regression. Significantly, the proposed framework takes the burden off the demand for abundant training necessary for the success of sparsity-based classification schemes.

\chapter{Conclusions and Future Directions}
\label{chapter:conclusion}

\section{Summary of Main Contributions}
The overarching theme in this research is the design of \emph{discriminative models} for \emph{robust image classification}. Two different families of discriminative models have been explored, based on: probabilistic graphical models and sparse signal representations. We have primarily considered multi-task classification scenarios where the different image representations exhibit discriminative structure - deterministic or probabilistic - that can be leveraged for robustness benefits.

In Chapters \ref{chapter:gm} and \ref{chapter:sparsity_gm}, we explore the ability of graphical models to fuse multiple feature sets for classification tasks. In particular, we base our contributions on a recent approach to discriminative learning of tree-structured graphs \cite{tan:tsp10}. By learning simple trees iteratively on the larger graphs formed by concatenating all nodes from each hypothesis and accumulating all edges, we obtain a final graphical model with dense edge structure that explicitly captured statistical correlations - which encode discriminative structure - across feature sets. The framework makes minimal demands on the choice of feature sets - it merely requires a collection of features that capture complementary yet correlated class information. In order to verify that the framework is indeed applicable in a variety of scenarios and for different choices of feature sets, we consider three important practical applications:
\begin{enumerate}
  \item Automatic target recognition: Here, wavelet sub-band LL, LH and HL features are chosen as the feature sets, owing to their popularity as feature choices for the ATR problem. On a related note, it is well-known that wavelet coefficients can be modeled using tree-structured graphs \cite{willsky02}.
  \item Hyperspectral target detection and classification: Hyperspectral imaging exhibits some unique properties, such as the presence of joint spatio-spectral information and the observation of spatial homogeneity of pixels. These ideas are exploited for robust classification by extracting multiple sparse features from pixels in local neighborhoods and fusing them via discriminative graphs. Our contribution directly builds upon very recent work \cite{Chen_2011c} which proposed a joint sparsity model to enforce identical sparse representation structure on neighboring pixels.
  \item Face recognition: The robustness of local features, in comparison with global image features, for classification tasks is well-known. We leverage this idea for robust face recognition by extracting sparse features from local informative regions such as the eyes, nose and mouth. Analogous to a single global dictionary in SRC, we design locally adaptive dictionaries borrowing from recent work \cite{chen:icip10} that encode robustness to minor geometric and photometric distortions. Discriminative graphs are then learned on these distinct feature representations akin to the procedure described for the above two applications.
\end{enumerate}
In each problem, we observe that our graphical model framework exhibits robustness to distortion scenarios intrinsic to the task. For example, pixel corruption and registration errors can cause significant degradation in face recognition performance, while the extended operating scenarios in ATR are acknowledged to be particularly severe conditions to test algorithm performance. In addition, our framework exhibits a more graceful decay in classification performance with reduction in the size of the training set. We consider this to be a significant experimental contribution, since the issue of limited training has not been as thoroughly investigated before despite being a concern in many practical problems.

In the second half of this dissertation, we propose discriminative models for sparse signal representations. Our first contribution is a simultaneous sparsity model which exploits correlations among the different color channels of medical images for disease identification. While this problem has similarities with many group/simultaneous models proposed recently in literature, the novelty of our contribution is in designing constraints based on an understanding of the imaging physics, leading to the formulation and solution of new optimization problems. The sparse coefficient matrices in our formulation exhibit a unique block-sparse structure that is not amenable to popular row sparsity-norm techniques such as the SOMP \cite{tropp:sp06a}. As the final contribution of this dissertation, we revisit the SRC framework from a Bayesian perspective. We design class-specific spike-and-slab priors in conjunction with class-specific dictionaries to obtain improved classification performance, even under the regime of limited training.

\section{Suggestions for Future Research}
The contributions in the previous chapters naturally point towards various directions for future research. We mention some of the possible extensions in this section.

\subsection{Discriminative Graph Learning}
Learning arbitrary graph structures is known to be an NP-hard problem \cite{cooper:1990}. While tree graphs are easy to learn, they are also limited in their modeling capacity. However, extensions of trees, such as junction trees and block trees, can still be learned in a tractable manner. Block trees have the interesting qualitative interpretation of clustering groups of graph nodes based on similarity. An immediate extension of our graphical model framework could therefore be to learn such block trees on the multiple feature representations and identify correlations across feature sets, or equivalently, graph node clusters.

\subsection{Joint Sparsity Models}
Sparse representation-based image classification is an area of ongoing research interest, and here we identify some connections to our work in published literature. Our framework can be generalized for any multi-variate/multi-task classification problem \cite{obozinski:sc09} by simply including training from those tasks as new sub-dictionaries. Recent work in multi-task classification has explored the idea of sparse models on image features \cite{yuan:tip12}. Admittedly, the sparse linear model may not be justifiable for all types of image classification problems. However, one way of incorporating non-linear sparse models is to consider the sparse model in a feature space induced by a kernel function \cite{chen:tgrs13}. Recent work has focused attention on solving the costly sparsity optimization problem more effectively \cite{bruckstein:siam09,yang:icip10,shia:asilomar11}. Our solution to the optimization problem in \eqref{eq:sparsity_main} is a modification of the greedy SOMP algorithm. We believe a deeper investigation towards efficient solutions to our modified optimization problem is a worthwhile research pursuit.

\subsection{Design of Class-specific Priors}
This aspect of our research opens the doors to many interesting theoretical extensions for robust image classification. We have demonstrated in Chapter \ref{chapter:struct_sparsity} that sparsity-inducing priors can introduce robustness to limited training. This analytical development has been demonstrated for the single task classification case. In our very recent work \cite{srinivas:icassp13b}, we have shown that the spike-and-slab prior can be extended in a hierarchical manner to account for collaborative representation problems. The formulations have parallels to the hierarchical lasso and the collaborative hierarchical lasso \cite{sprechmann:tsp11}, with the additional improvement that our formulation explicitly captures sparsity via an $l_0$-norm minimization (compared to $l_1$-relaxations in \cite{sprechmann:tsp11}). These extensions readily carry forward to classification tasks by learning the priors in a class-specific manner.

We obtain multiple optimization problems, one for each class. These are distinguished by the different values of regularization parameters, chosen in a class-specific manner. So an important research challenge is the accurate selection of regularization parameters. One of the benefits of Bayesian learning is that a full posterior estimate of parameter distributions is obtained, instead of a point estimate used in the optimization-based approaches. Other approaches to learn these parameters, such as MCMC sampling, can also be explored. We have primarily considered the spike-and-slab priors on account of their wide acceptability in modeling sparse signals. An interesting research direction would be the investigation of other families of priors that can simultaneously capture sparsity and discriminative structure, the resulting optimization problems and their solutions. Finally, this framework can be applied to a variety of multi-modal classification problems where robustness is an important concern.

%
\appendix
\Appendix{A Greedy Pursuit Approach to Multi-task Classification}
\label{sec:appA}

\noindent{\textbf{Notation:}} Let $\vect y_i \in \mathbb{R}^n, i = 1,\ldots,T$ be $T$ different representations of the same physical event, which is to be classified into one of $K$ different classes. Let $\mat Y := [\vect y_1~\ldots ~ \vect y_T] \in \mathbb{R}^{m \times T}$. Assuming $n$ training samples/events in total, we design $T$ dictionaries $\mat D_i \in \mathbb{R}^{m\times n}, i = 1,\ldots,T$, corresponding to the $T$ representations. We define a new composite dictionary $\mat D := [\mat D_1 ~\ldots ~ \mat D_T] \in \mathbb{R}^{m \times nT}$. Further, each dictionary $\mat D_i$ is represented as the concatenation of the sub-dictionaries from all classes corresponding to the $i$-th representation of the event:
\be
\mat D_i := [\mat D_{i}^{1} ~\mat D_{i}^{2}~\ldots ~ \mat D_{i}^{K}],
\ee
where $\mat D_{i}^{j}$ represents the collection of training samples for representation $i$ that belong to the $j$-th class. So, we have:
\be
\mat D := [\mat D_1 ~\ldots ~ \mat D_T] = [\mat D_{1}^{1} ~\mat D_{1}^{2}~\ldots ~ \mat D_{1}^{K} ~~~ \ldots ~~~ \mat D_{T}^{1} ~\mat D_{T}^{2}~\ldots ~ \mat D_{T}^{K}].
\ee

A test event $\mat Y$ can now be represented as a linear combination of training samples as follows:
\beaa
\mat Y & = & [\vect y_1 ~~\ldots ~~ \vect y_T] = \mat D \mat S \\
& = &\left[\mat D_{1}^{1} ~\mat D_{1}^{2}~\ldots ~ \mat D_{1}^{K} ~~~ \ldots ~~~ \mat D_{T}^{1} ~\mat D_{T}^{2}~\ldots ~ \mat D_{T}^{K}\right]\left[\vect \alpha_1 ~~ \ldots ~~\vect \alpha_T\right],
\eeaa
where the coefficient vectors $\vect \alpha_i \in \mathbb{R}^{nT}, i = 1,\ldots,T$, and $\mat S = \left[\vect \alpha_1 ~~ \ldots ~~\vect \alpha_T\right] \in \mathbb{R}^{nT \times T}$.

Since $\mat S$ obeys column correspondence, we introduce a new matrix $\mat S' \in \mathbb{R}^{n\times T}$ as the transformation of $\mat S$ with the zero coefficients removed,
\[
\mat S' = \left[\begin{array}{ccccc}
\vect{\alpha}_{1}^{1} & \ldots & \vect{\alpha}_{i}^{1} & & \vect{\alpha}_{T}^{1}\\
\vdots & \vdots & \vdots & \vdots & \vdots\\
\vect{\alpha}_{1}^{K} & \ldots& \vect{\alpha}_{i}^{K} & \ldots& \vect{\alpha}_{T}^{K}
\end{array}\right],
\]
where $\vect \alpha_i^j$ refers to the sub-vector extracted from $\vect \alpha_i$ that corresponds to coefficients from the $j$-th class. Note that, in the $i$-th column of $\mat S'$, only the coefficients corresponding to $\mat D_i$ are retained (for $i = 1,\ldots,T$).

\begin{algorithm}[t]
\caption{SOMP for multi-task multivariate sparse representation-based classification}
\label{alg1}
\begin{algorithmic}
\REQUIRE Dictionary $\mat D$, signal matrix $\mat Y$, number of iterations $K$\\
\textbf{Initialization}: residual $\mat R_0 = \mat Y$, index set $\Lambda_0 = \phi$, iteration counter $k = 1$
\WHILE{$k\leq K$}
\STATE (1) Find the index of the atom that best approximates all residuals:\\
$\lambda_{i,k} = \arg\max\limits_{j = 1,\ldots,n} \sum_{q=1}^{T}w_q\left\Vert \mat R_{k-1}^{t}\vect d_{q,j}\right\Vert_{p}, p \geq 1$

\STATE(2) Update the index set $\Lambda_{i,k} = \Lambda_{i,k-1} \bigcup \left\{\lambda_{i,k}\right\}, i = 1,\ldots,T$

\STATE(3) Compute the orthogonal projector
$\vect p_{i,k} = \left(\mat D_{\Lambda_{i,k}}^t\mat D_{\Lambda_{i,k}}\right)^{-1}\mat D_{\Lambda_{i,k}}^t\vect y_i$, for $i = 1,\ldots,T$, where $\mat D_{\Lambda_{i,k}}\in \mathbb{R}^{n \times k}$ consists of the $k$ atoms in $\mat D_i$ indexed in $\Lambda_{i,k}$

\STATE(4) Update the residual matrix $\mat R_k = \mat Y - \left[\mat D_{\Lambda_{1,k}}\vect p_{1,k} ~~ \ldots ~~\mat D_{\Lambda_{T,k}}\vect p_{T,k}\right]$

\STATE(5) Increment $k$: $k \leftarrow k+1$
\ENDWHILE

\ENSURE Index set $\Lambda_i = \Lambda_{i,K}, i = 1,\ldots,T$; sparse representation $\hat{\mat S}'$  whose non-zero rows indexed for each representation by $\Lambda_i$, $i, i = 1,\ldots,T$, are the $K$ rows of the matrix $\left(\mat D_{\Lambda_{i,K}}^t\mat D_{\Lambda_{i,K}}\right)^{-1}\mat D_{\Lambda_{i,K}}^t\mat Y$.
\end{algorithmic}
\end{algorithm}

We can now apply row-sparsity constraints similar to the approach in \cite{tropp:sp06a}. Our modified optimization problem becomes:
\begin{eqnarray}
\hat{\mat S}' = \arg\min_{\mat S'} \left\Vert \mat S'\right\Vert_{\mbox{\scriptsize{row}},0} & \mbox{subject to} & \left\Vert \mat Y - \mat D\mat S\right\Vert_{F}\leq\epsilon,
\label{eq:sparsity main_app}
\end{eqnarray}
for some tolerance $\epsilon > 0$. We minimize the number of non-zero rows, while the constraint guarantees a good approximation.

The matrix $\mat S$ can be transformed into $\mat S'$ by introducing matrices $\mat H \in \mathbb{R}^{nT\times T}$ and $\mat J \ \in \mathbb{R}^{n\times nT}$,
\[
\mat H = \mbox{diag}\left[\vect 1 ~ \vect 1 ~ \ldots ~ \vect 1\right],\mat J = \left[\mat I_n ~ \mat I_n ~ \ldots ~ \mat I_n\right],
\]
where $\vect 1 \in \mathbb{R}^n$ is the vector of all ones, and $\mat I_n$ denotes the $n$-dimensional identity matrix. Finally, we obtain $\mat S' = \mat J \left(\mat H\circ\mat S\right)$, where $\circ$ denotes the Hadamard product, $\left(\mat H\circ\mat S\right)_{ij}\triangleq h_{ij}s_{ij}$ for all $i,j$. Eq. (\ref{eq:sparsity main_app}) represents a hard optimization problem due to presence of the non-invertible transformation from $\mat S$ to $\mat S'$. We bypass this difficulty by proposing a modified version of the SOMP algorithm for the multi-task multivariate case.

Recall that the original SOMP algorithm gives $K$ distinct atoms (assuming $K$ iterations) from a dictionary $\bm{D}$ that best represent the data matrix $\bm{Y}$. In every iteration $k$, SOMP measures the residual for each atom in $\bm{D}$ and creates an orthogonal projection with maximal correlation. Extending this to the multi-task setting, for every representation $i, i = 1,\ldots,T$, we can identify the index set that gives the highest correlation with the residual at the $k$-th iteration as follows:
\[
\lambda_{i,k} = \arg\max\limits_{j = 1,\ldots,n} \sum_{q=1}^{T}w_q\left\Vert \mat R_{k-1}^{t}\vect d_{q,j}\right\Vert_{p}, p \geq 1,
\]
where $w_q$ denotes the weight (confidence) assigned to the $q$-th representation, $\vect d_{q,j}$ represents the $j$-th column of $\mat D_q, q = 1,\ldots,T$, and the superscript $(\cdot)^t$ indicates the matrix transcript operator. After finding $\lambda_{i,k}$, we modify the index set to:
\ben
\Lambda_{i,k} = \Lambda_{i,k-1}\bigcup\lambda_{i,k}, i = 1,\ldots,T.
\een
Thus, by finding the index set for the $T$ distinct representations, we can create an orthogonal projection with each of the atoms in their corresponding representations. The algorithm is summarized in Algorithm \ref{alg1}.
\Appendix{Parameter Learning for SSPIC}
\label{appendix:b}

\begin{algorithm}[H]
\caption{Parameter learning}
\label{alg:sspic}
\begin{algorithmic}
\REQUIRE $\mat A$\\
\textbf{Initialization}: iteration counter $k = 1$
\WHILE{$k\leq K$}
\STATE (1) Update spike-and-slab parameters ($i = 1,\ldots,n$):\\

\beaa
\hat{\vect y_i} &\leftarrow &\vect y - \sum_{j \neq i} x_j\vect a_j\\
\Sigma_i & = &\sigma^2\left(\lambda + \vect a_i^T\vect a_i\right)^{-1}\\
\mu_i & = &\frac{1}{\sigma^2}\Sigma_i\vect a_i^T\hat{\vect y_i}\\
\gamma_i &\leftarrow &\left(1 + \frac{1-\gamma_i}{\gamma_i\exp(-1/2\Sigma_i\mu_i^2)}\right)^{-1} \\
(x_i | -) & \sim & (1-\gamma_i)\delta_0 + \gamma_i \mathcal{N}(\mu_i,\Sigma_i)
\eeaa
\STATE(2) Update signal noise variance:

\ben
(\sigma^2 | - ) \sim \Gamma^{-1}\left(e_0 + 0.5m, f_0 + 1/2\|\vect y - \mat A\vect x\|_2^2\right)
\een

\STATE(3) Increment $k$: $k \leftarrow k+1$
\ENDWHILE

\ENSURE $\sigma^2, \{\gamma_i\}$
\end{algorithmic}
\end{algorithm}
} 

  \begin{singlespace}
   \bibliographystyle{IEEEtran}
   \addcontentsline{toc}{chapter}{Bibliography}
   \bibliography{IEEEabrv,bibChapter1,bibChapter2,bibChapter3,bibChapter4,bibChapter5}
   \end{singlespace}

\backmatter


Umamahesh Srinivas received his B.Tech degree in electronics and communication engineering from the National Institute of Technology Karnataka (NITK), Surathkal, India in May 2007 and his M.S. degree in electrical engineering from The Pennsylvania State University in December 2009. During the summer of 2010, he was a summer research intern at Qualcomm MEMS Technologies, San Jose, CA, where he worked on halftoning algorithms for color image displays. In the fall of 2012, he was a research intern at the U.S. Army Research Laboratory, Adelphi, MD.

Mr. Srinivas received the 2012 IEEE Mikio Takagi Prize for the Best Student Paper at the IEEE International Geoscience and Remote Sensing Symposium held in Munich, Germany in July 2012. He also received the Best Automatic Target Recognition Student Paper Award at the SPIE Defense, Security and
Sensing Symposium held in Baltimore, MD in April 2012. He is a student member of IEEE, SPIE and SIAM.

\end{document}